\documentclass[twoside,11pt]{article}

\usepackage{jmlr2e}
\usepackage[latin1]{inputenc}
\usepackage[T1]{fontenc}
\usepackage{amsmath,amsfonts,amssymb}
\usepackage{bm}
\usepackage{subfigure}
\usepackage{url}
\usepackage{longtable}
\usepackage{enumitem}
\usepackage[width=.9\textwidth]{caption}

\DeclareMathOperator{\Kfu}{\mathbf{K}_{f,u}}
\DeclareMathOperator{\Kuf}{\mathbf{K}_{u,f}}
\DeclareMathOperator{\Kff}{\mathbf{K}_{f,f}}
\DeclareMathOperator{\iKff}{\mathbf{K}_{f,f}^{-1}}
\DeclareMathOperator{\Kfa}{\mathbf{K}_{f,\tilde{f}}}
\DeclareMathOperator{\Kaf}{\mathbf{K}_{\tilde{f},f}}
\DeclareMathOperator{\Kaa}{\mathbf{K}_{\tilde{f},\tilde{f}}}
\DeclareMathOperator{\Kuu}{\mathbf{K}_{u,u}}
\DeclareMathOperator{\iKuu}{\mathbf{K}_{u,u}^{-1}}
\DeclareMathOperator{\Kau}{\mathbf{K}_{\tilde{f},u}}
\DeclareMathOperator{\Kua}{\mathbf{K}_{u,\tilde{f}}}
\DeclareMathOperator{\Qff}{\mathbf{Q}_{f,f}}

\DeclareMathOperator{\x}{\mathbf{x}}
\DeclareMathOperator{\f}{\mathbf{f}}
\DeclareMathOperator{\y}{\mathbf{y}}
\DeclareMathOperator{\h}{\mathbf{h}}
\DeclareMathOperator{\uu}{\mathbf{u}}
\DeclareMathOperator{\LL}{\mathbf{\Lambda}}
\DeclareMathOperator{\bb}{\mathbf{b}}
\def\WAIC{\mathrm{WAIC}}

\DeclareMathOperator{\Poisson}{Poisson}

\DeclareMathOperator{\GP}{\mathcal{GP}}
\DeclareMathOperator{\N}{N}

\DeclareMathOperator*{\argmax}{arg\,max}
\DeclareMathOperator*{\argmin}{arg\,min}
\newcommand{\mb}{\mathbf}
\newcommand{\pkg}[1]{{\fontseries{b}\selectfont #1}}
\newcommand{\proglang}{}
\newcommand{\email}[1]{\href{mailto:#1}{\normalfont\texttt{#1}}}

\newcommand{\code}[1]{{\normalfont\texttt{#1}}}

\DeclareMathOperator{\E}{E}
\DeclareMathOperator{\VAR}{Var}
\DeclareMathOperator{\COV}{Cov}

\makeatletter
\def\@oddhead{}%
\def\@oddfoot{}%
\def\@evenhead{}%
\def\@evenfoot{}%
\def\@starteditor{}%
\makeatother
\editor{}
\usepackage{hyperref}
\hypersetup{
    bookmarks=true,         
    pdfstartview={FitH},    
    colorlinks=true,        
    linkcolor=black,        
    citecolor=black,        
    filecolor=black,        
    urlcolor=black          
}

\begin{document}

\title{Bayesian Modeling with Gaussian Processes using the
   \pkg{GPstuff} Toolbox}

\date{2015}

\author{\name Jarno Vanhatalo\thanks{Work done mainly while at the
    Department of Biomedical
    Engineering and Computational Science, Aalto University} \email jarno.vanhatalo@helsinki.fi \\
  \addr Department of Environmental Sciences\\
  University of Helsinki\\
  P.O. Box 65, FI-00014 Helsinki, Finland \AND
  \name Jaakko Riihim\"aki\thanks{Not anymore in Aalto University} \email { } \\
  \name Jouni Hartikainen$^{\dagger}$ \email { } \\
  \name Pasi Jyl\"anki$^{\dagger}$ \email { } \\
  \name Ville Tolvanen$^{\dagger}$ \email { } \\
  \name Aki Vehtari\thanks{Corresponding author} \email aki.vehtari@aalto.fi \\
  \addr Helsinki Institute for Information Technology HIIT, Department of Computuer Science\\
  Aalto University School of Science\\
  P.O. Box 15400, FI-00076 Aalto, Finland}

\maketitle

\begin{abstract}
  Gaussian processes (GP) are powerful tools for probabilistic
  modeling purposes. They can be used to define prior distributions 
  over latent functions in hierarchical Bayesian models. The prior
  over functions is defined implicitly by the mean and covariance
  function, which determine the smoothness and variability of the
  function. The inference can then be conducted directly in the
  function space by evaluating or approximating the posterior process.
  Despite their attractive theoretical properties GPs provide
  practical challenges in their implementation. \pkg{GPstuff} is a
  versatile collection of computational tools for GP models compatible
  with Linux and Windows \proglang{MATLAB} and Octave.  It includes,
  among others, various inference methods, sparse approximations and
  tools for model assessment. In this work, we review these tools and
  demonstrate the use of \pkg{GPstuff} in several models.\\ Last updated 2015-07-15.
\end{abstract}


%

\newpage

\tableofcontents

\section{Introduction}

This work reviews a free open source toolbox \pkg{GPstuff} (version
4.6) which implements a collection of inference methods and tools for
inference for various Gaussian process (GP) models. The toolbox is
compatible with Unix and Windows \proglang{MATLAB} \citep{MATLAB:2010}
(r2009b or later, earlier versions may work, but has not been tested
extensively) and \proglang{Octave} \citep{octave:2012} (3.6.4 or
later, compact support covariance functions are not currently working
in Octave). It is available from
\url{http://becs.aalto.fi/en/research/bayes/gpstuff/}.
If you find GPstuff useful, please use the reference \citep{Vanhatalo+gpstuff:2013}
\begin{itemize}
\item[] Jarno Vanhatalo, Jaakko Riihim{\"a}ki, Jouni Hartikainen, Pasi
  Jyl{\"a}nki, Ville Tolvanen and Aki Vehtari (2013). GPstuff: Bayesian
  Modeling with Gaussian Processes. \emph{Journal of Machine Learning
    Research}, 14:1175-1179
\end{itemize}
and appropriate references mentioned in this text or in the GPstuff
functions you are using.

GP is a stochastic process, which provides a powerful tool for
probabilistic inference directly on distributions over functions
\citep[e.g.][]{OHagan:1978} and which has gained much attention in
recent years \citep[][]{Rasmussen+Williams:2006}. In many practical GP
models, the observations $\y = [y_1,...,y_n]^{\text{T}}$ related to
inputs (covariates) $\mb{X} = \{\x_i = [x_{i,1},...
,x_{i,d}]^{\text{T}} \}_{i=1}^n$ are assumed to be conditionally
independent given a latent function (or predictor) $f(\x)$ so that the
likelihood $p(\y|\f) = \prod_{i=1}^{n} p(y_i|f_i)$, where $\f =
[f(\x_1),...,f(\x_n)]^{\text{T}}$, factorizes over cases. GP prior is
set for the latent function after which the posterior $p(f|\y,\mb{X})$
is solved and used for prediction. GP defines the prior over the
latent function implicitly through the mean and covariance function,
which determine, for example, the smoothness and variability of the
latent function.  Usually, the model hierarchy is extended to the
third level by giving also prior for the parameters of the covariance
function and the observation model.

Most of the models in \pkg{GPstuff} follow the above definition and
can be summarized as:
\begin{align}
  \text{Observation model:}   &&  \y|\f,\phi   &\sim  \prod_{i=1}^{n} p(y_i|f_i, \phi) \label{likelihood}\\
  \text{GP prior:}   && f(\x)|\theta    &\sim  \GP\left(m(\x),  k(\x,\x'|\theta)\right) \label{GP_prior} \\
  \text{hyperprior:}   &&  \theta, \phi  &\sim  p(\theta)p(\phi). \label{hyper_prior}
\end{align}
Here $\theta$ denotes the parameters in the covariance function
$k(\x,\x'|\theta)$, and $\phi$ the parameters in the observation
model. We will call the function value $f(\x)$ at fixed $\x$ a latent
variable and for brevity we sometimes denote all the covariance
function and observation model parameters with $\vartheta = [\theta,
\phi]$. For the simplicity of presentation the mean function is
considered zero, $m(\x) \equiv 0$, throughout this paper. We will also
denote both the observation model and likelihood by $p(\y|\f,\phi)$
and assume the reader is able to distinguish between these two from
the context. The likelihood naming convention is used in the toolbox
for both likelihood and observation model related functionalities
which follows the naming convention used, for example, in \pkg{INLA}
\citep{Rue+Martino+Chopin:2009} and \pkg{GPML}
\citep{Rasmussen+Nickisch:2010} software packages. There are also
models with multiple latent processes and likelihoods where each
factor depens on multiple latent values. These are discussed in
Section \ref{sec:multilatent-models}.

Early examples of GP models can be found, for example, in time series
analysis and filtering \citep{Wiener:1949} and in geostatistics
\citep[e.g.][]{Matheron:1973}.  GPs are still widely used in these
fields and useful overviews are provided in
\citep{Cressie:1993,Grewal+Andrews:2001,Diggle+Ribeiro:2007,Gelfand+Diggle+Fuentes+Guttorp:2010}.
\citet{OHagan:1978} was one of the firsts to consider GPs in a general
probabilistic modeling context and provided a general theory of GP
prediction. This general regression framework was later rediscovered
as an alternative for neural network models
\citep{Williams+Rasmussen:1996,Rasmussen:1996} and extended for other
problems than regression \citep{Neal:1997,Williams+Barber:1998}. This
machine learning perspective is comprehensively summarized by
\citet{Rasmussen+Williams:2006}.

Nowadays, GPs are used, for example, in weather forecasting
\citep{Fuentes+Raftery:2005,Berrocal+Raftery+Gneiting+Steed:2009},
spatial statistics
\citep{Best+Richardson+Thomson:2005,Kaufman+Schervish+Nychka:2008,Banerjee+Gelfand+Finley+Sang:2008,Myllymaki+Sarkka+Vehtari:2014},
computational biology
\citep{Honkela+Gao+Ropponen+Rattray+Lawrence:2011}, reinforcement
learning
\citep{Deisenroth+Rasmussen+Peters:2009,Deisenroth+Rasmussen+Fox:2011},
healthcare applications
\citep{Stegle+Fallert+MacKay+Brage:2008,Vanhatalo+Pietilainen+Vehtari:2010,Rantonen+Vehtari+etal:2012,Rantonen+Vehtari+etal:2014},
survival analysis \citep{Joensuu+etal:2012a,Joensuu+Reichardt+Eriksson+Hall+Vehtari:2014}
industrial applications \citep{Kalliomaki+Vehtari+Lampinen:2005},
computer model calibration and emulation
\citep{Kennedy+OHagan:2001,Conti+Gosling+Oakley+OHagan:2009}, prior
elicitation \citep{Moala+OHagan:2010} and density estimation
\citep{Tokdar+Ghosh:2007,Tokdar+Zhu+Ghosh:2010,Riihimaki+Vehtari:2014} to name a few.
Despite their attractive theoretical properties and wide range of
applications, GPs provide practical challenges in implementation.

\pkg{GPstuff} provides several state-of-the-art inference algorithms
and approximative methods that make the inference easier and faster
for many practically interesting models. \pkg{GPstuff} is a modular
toolbox which combines inference algorithms and model structures in an
easily extensible format. %
It also provides various tools for model checking and comparison.
These are essential in making model assessment and criticism an
integral part of the data analysis. Many algorithms and models in
\pkg{GPstuff} are proposed by others but reimplemented for
\pkg{GPstuff}. In each case, we provide reference to the original work
but the implementation of the algorithm in \pkg{GPstuff} is always
unique. The basic implementation of two important algorithms, the
Laplace approximation and expectation propagation algorithm (discussed
in Section \ref{sec_cond_post_of_latent}), follow the pseudocode from
\citep{Rasmussen+Williams:2006}. However, these are later generalized
and improved as described in
\citep{Vanhatalo+Jylanki+Vehtari:2009,Vanhatalo+Pietilainen+Vehtari:2010,
  Vanhatalo+Vehtari:2010,Jylanki+Vanhatalo+Vehtari:2011,Riihimaki+Jylanki+Vehtari:2013,Riihimaki+Vehtari:2014}.

There are also many other toolboxes for GP modelling than
\pkg{GPstuff} freely available. Perhaps the best known packages are
nowadays the Gaussian processes for Machine Learning \pkg{GPML}
toolbox \citep{Rasmussen+Nickisch:2010} and the flexible Bayesian
modelling (\pkg{FBM}) toolbox by Radford Neal. A good overview of
other packages can be obtained from the Gaussian processes website
\url{http://www.gaussianprocess.org/} and the \proglang{R} Archive
Network \url{http://cran.r-project.org/}.  Other GP softwares have
some overlap with \pkg{GPstuff} and some of them include models that
are not implemented in \pkg{GPstuff}.  The main advantages of
\pkg{GPstuff} over other GP software are its versatile collection of
models and computational tools as well as modularity which allows easy
extensions. A comparison between \pkg{GPstuff}, \pkg{GPML} and
\pkg{FBM} is provided in the Appendix \ref{appendix:comparison}.

Three earlier GP and Bayesian modelling packages have influenced our
work.  Structure of \pkg{GPstuff} is mostly in debt to the
\pkg{Netlab} toolbox \citep{Nabney:2001}, although it is far from
being compatible.  The \pkg{GPstuff} project was started in 2006 based on
the \pkg{MCMCStuff}-toolbox (1998-2006)
(\url{http://becs.aalto.fi/en/research/bayes/mcmcstuff/}). \pkg{MCMCStuff}
for its part was based on \pkg{Netlab} and it was also influenced by
the \pkg{FBM}. The \pkg{INLA} software package by
\citet{Rue+Martino+Chopin:2009} has also motivated some of the
technical details in the toolbox. In addition to these, some technical
implementations of \pkg{GPstuff} rely on the sparse matrix toolbox
\pkg{SuiteSparse} \citep{Davis:2005}
(\url{http://www.cise.ufl.edu/research/sparse/SuiteSparse/}).

This work concentrates on discussing the essential theory and methods
behind the implementation of \pkg{GPstuff}. We explain important parts
of the code, but the full code demonstrating the important features of
the package (including also data creation and such), are included in
the demonstration files \code{demo\_*} to which we refer in the text.

\section{Gaussian process models}

\subsection{Gaussian process prior}

GP prior over function $f(\x)$ implies that any set of function values
$\f$, indexed by the input coordinates $\mb{X}$, have a multivariate
Gaussian distribution
\begin{equation}\label{eq_GP_prior}
p(\f|\mb{X},\theta) = \N(\f|\mb{0}, \Kff),
\end{equation}
where $\Kff$ is the covariance matrix. Notice, that the distribution
over functions will be denoted by $\GP(\cdot,\cdot)$, whereas the
distribution over a finite set of latent variables will be denoted by
$\N(\cdot,\cdot)$. The covariance matrix is constructed by a
covariance function, $[\Kff]_{i,j} = k(\mb{x}_i, \mb{x}_j|\theta)$,
which characterizes the correlation between different points in the
process.  Covariance function can be chosen freely as long as the
covariance matrices produced are symmetric and positive semi-definite
($\mb{v}^{\text{T}}\Kff\mb{v}\geq 0, \forall \mb{v}\in \Re^n$). An
example of a stationary covariance function is the squared exponential
\begin{equation}
k_{\mathrm{se}}(\x_i,\x_j|\theta)=\sigma_{\mathrm{se}}^2\exp(-r^2/2),
\end{equation}
where $r^2 = \sum_{k=1}^d (x_{i,k}-x_{j,k})^2/l_k^2$ and $\theta =
[\sigma_{\mathrm{se}}^2, l_1,...,l_d]$. Here, $\sigma_{\mathrm{se}}^2$
is the scaling parameter, and $l_k$ is the length-scale, which governs
how fast the correlation decreases as the distance increases in the
direction $k$. Other common covariance functions are discussed, for
example, by \citet{Diggle+Ribeiro:2007},
\citet{Finkenstadt+Held+Isham:2007} and
\citet{Rasmussen+Williams:2006} and the covariance functions in
\pkg{GPstuff} are summarized in the appendix
\ref{covariance_functions}.

Assume that we want to predict the values $\tilde{\f}$ at new input
locations $\tilde{\mb{X}}$. The joint prior for latent variables $\f$
and $\tilde{\f}$ is
\begin{equation}
\left[ \begin{matrix} \f \\ \tilde{\f} \end{matrix} \right] | \mb{X},
\tilde{\mb{X}},\theta
\sim \N\left(\mb{0}, \left[ \begin{matrix} \Kff & \Kfa \\ \Kaf & \Kaa
    \end{matrix} \right] \right),
\end{equation}
where $\Kff = k(\mb{X},\mb{X}|\theta)$, $\Kfa =
k(\mb{X},\tilde{\mb{X}}|\theta)$ and $\Kaa =
k(\tilde{\mb{X}},\tilde{\mb{X}}|\theta)$. Here, the covariance
function $k(\cdot,\cdot)$ denotes also vector and matrix valued
functions $k(\x,\mb{X}):\Re^d \times \Re^{d \times n}\rightarrow
\Re^{1\times n}$, and $k(\mb{X},\mb{X}):\Re^{d\times n}\times
\Re^{d\times n}\rightarrow \Re^{n\times n}$. By definition of GP the
marginal distribution of $\tilde{\f}$ is
$p(\tilde{\f}|\tilde{\mb{X}},\theta) = \N(\tilde{\f}|\mb{0}, \Kaa)$
similar to \eqref{eq_GP_prior}.  The conditional distribution of
$\tilde{\f}$ given $\f$ is
\begin{equation}\label{eq_conditional_ftilde_given_f}
\tilde{\f}|\f, \mb{X},
\tilde{\mb{X}}, \theta \sim \N(\Kaf\iKff\f, \Kaa - \Kaf\iKff\Kfa),
\end{equation}
where the mean and covariance of the conditional distribution are
functions of input vector $\tilde{\x}$ and $\mb{X}$ serves as a fixed
parameter.  Thus, the above distribution generalizes to GP with mean
function $m(\tilde{\x}|\theta) = k(\tilde{\x},\mb{X}|\theta)\iKff\f$
and covariance $k(\tilde{\x}, \tilde{\x}'|\theta) = k(\tilde{\x},
\tilde{\x}'|\theta) - k(\tilde{\x},\mb{X}|\theta)\iKff
k(\mb{X},\tilde{\x}'|\theta)$, which define the conditional
distribution of the latent function $f(\tilde{\x})$.

\subsection{Conditioning on the observations}

The cornerstone of the Bayesian inference is Bayes' theorem by
which the conditional probability of the latent function and
parameters after observing the data can be solved. This posterior
distribution contains all information about the latent function and
parameters conveyed from the data $\mathcal{D} = \{\mb{X}, \mb{y}\}$
by the model. Most of the time we cannot solve the posterior but need
to approximate it.  \pkg{GPstuff} is built so that the first inference
step is to form (either analytically or approximately) the conditional
posterior of the latent variables given the parameters 
\begin{equation}\label{eq_conditional_posterior_of_f}
p(\f|\mathcal{D}, \theta, \phi) =
\frac{p(\y|\f,\phi)p(\f|,\mb{X}, \theta)}{\int p(\y|\f,\phi)p(\f|\mb{X},\theta) d\f},
\end{equation}
which is discussed in the section~\ref{sec_cond_post_of_latent}. After
this, we can (approximately) marginalize over the parameters to obtain the
marginal posterior distribution for the latent variables
\begin{equation}
p(\f|\mathcal{D}) = \int p(\f|\mathcal{D}, \theta, \phi) p(\theta, \phi|\mathcal{D}) d\theta d\phi
\end{equation}
treated in the section~\ref{sec_marginalization_over_hyperparam}. The
posterior predictive distributions can be obtained similarly by first
evaluating the conditional posterior predictive distribution, for
example $p(\tilde{f}|\mathcal{D}, \theta, \phi, \tilde{\x})$, and then
marginalizing over the parameters. The joint predictive distribution
$p(\tilde{\y}|\mathcal{D}, \theta, \phi, \tilde{\x})$ would require
integration over possibly high dimensional distribution
$p(\tilde{\f}|\mathcal{D}, \theta, \phi, \tilde{\x})$. However,
usually we are interested only on the marginal predictive distribution
for each $\tilde{y}_i$ separately which requires only one dimensional
integrals
\begin{equation}
  p(\tilde{y}_i | \mathcal{D}, \tilde{\mb{x}}_i, \theta, \phi) = \int p(\tilde{y_i}|
  \tilde{f}_i, \phi) p(\tilde{f}_i | 
  \mathcal{D}, \tilde{\x}_i, \theta, \phi)d \tilde{f}_i.
\end{equation}

If the parameters are considered fixed, GP's marginalization and
conditionalization properties can be exploited in the prediction.
Given the conditional posterior distribution $p(\f|\mathcal{D},\theta,
\phi)$, which in general is not Gaussian, we can evaluate the
posterior predictive mean from the conditional mean
$\E_{\tilde{\f}|\f,\theta,\phi}[f(\tilde{\x})] =
k(\tilde{\x},\mb{X})\iKff\f$ (see equation
\eqref{eq_conditional_ftilde_given_f} and the text below it). Since
this holds for any $\tilde{\f}$, we obtain a parametric posterior mean
function
\begin{eqnarray}\label{eq_posterior_predictive_mean}
m_p(\tilde{\x}|\mathcal{D},\theta,\phi) =
\int \E_{\tilde{\f}|\f,\theta,\phi}[f(\tilde{\x})]
p(\f|\mathcal{D},\theta,\phi) d\f = k(\tilde{\x},\mb{X}|\theta)\iKff
\E_{\f|\mathcal{D},\theta,\phi}[\f]. 
\end{eqnarray}
The posterior predictive covariance between any set of latent
variables, $\tilde{\f}$ is 
\begin{equation}
\COV_{\tilde{\f}|\mathcal{D},\theta,\phi}[\tilde{\f}] = \E_{\f|\mathcal{D},\theta,\phi}
\left[\COV_{\tilde{\f}|\f}[\tilde{\f}] \right] +
\COV_{\f|\mathcal{D},\theta,\phi}\left[\E_{\tilde{\f}|\f}[\tilde{\f} ]\right],
\end{equation}
where the first term simplifies to the conditional covariance in
equation \eqref{eq_conditional_ftilde_given_f} and the second term to
$k(\tilde{\x},\mb{X}) \iKff \COV_{\f|\mathcal{D},\theta,\phi}[\f]\iKff
k(\mb{X},\tilde{\x}')$. The posterior covariance function is then
\begin{equation}\label{eq_posterior_predictive_covariance}
k_p(\tilde{\x},\tilde{\x}'|\mathcal{D},\theta,\phi) = k(\tilde{\x},\tilde{\x}'|\theta) - k(\tilde{\x},\mb{X}|\theta) \left(\iKff -
\iKff \COV_{\f|\mathcal{D},\theta,\phi}[\f]\iKff\right)k(\mb{X},\tilde{\x}'|\theta). 
\end{equation}
From now on the posterior predictive mean and covariance will be
denoted $m_p(\tilde{\x})$ and $k_p(\tilde{\x},\tilde{\x}')$.

Even if the exact posterior $p(\tilde{f}|\mathcal{D},\theta,\phi)$ is
not available in closed form, we can still approximate its posterior
mean and covariance functions if we can approximate
$\E_{\f|\mathcal{D},\theta,\phi}$ and
$\COV_{\f|\mathcal{D},\theta,\phi}[\f]$. A common practice to
approximate the posterior $p(\f|\mathcal{D},\theta,\phi)$ is either
with Markov chain Monte Carlo (MCMC)
\citep[e.g.][]{Neal:1997,Neal:1998,Diggle+Tawn+Moyeed:1998,Kuss+Rasmussen:2005,Christensen+Roberts+Skold:2006}
or by giving an analytic approximation to it
\citep[e.g.][]{Williams+Barber:1998,Gibbs+MacKay:2000,Minka:2001,Csato+Opper:2002,Rue+Martino+Chopin:2009}.
The analytic approximations considered here assume a Gaussian form in
which case it is natural to approximate the predictive distribution
with a Gaussian as well.  In this case, the equations
\eqref{eq_posterior_predictive_mean} and
\eqref{eq_posterior_predictive_covariance} give its mean and
covariance. Detailed considerations on the approximation error and the
asymptotic properties of the Gaussian approximation are presented, for
example, by \citet{Rue+Martino+Chopin:2009} and
\citet{Vanhatalo+Pietilainen+Vehtari:2010}.

\section{Conditional posterior and predictive distributions}\label{sec_cond_post_of_latent}

\subsection{Gaussian observation model: the analytically tractable case}
\label{sec:gauss-observ-model}

With a Gaussian observation model, $y_i \sim \N(f_i, \sigma^2)$, where
$\sigma^2$ is the noise variance, the conditional posterior of the
latent variables can be evaluated analytically. Marginalization over
$\f$ gives the marginal likelihood
\begin{equation}\label{eq_marginal_likelihood_gaussian_case}
p(\y|\mb{X},\theta, \sigma^2) = \N(\y|\mb{0}, \Kff + \sigma^2\mb{I}).
\end{equation}
Setting this in the denominator of the equation
\eqref{eq_conditional_posterior_of_f}, gives a Gaussian distribution
also for the conditional posterior of the latent variables
\begin{equation}\label{eq_posterior_in_gaussian_case}
\f|\mathcal{D},\theta, \sigma^2 \sim \N(\Kff(\Kff + \sigma^2\mb{I})^{-1}\y, \Kff - \Kff(\Kff + \sigma^2\mb{I})^{-1}\Kff).
\end{equation}

Since the conditional posterior of $\f$ is Gaussian, the posterior
process, or distribution $p(\tilde{f}|\mathcal{D})$, is also Gaussian.
By placing the mean and covariance from
\eqref{eq_posterior_in_gaussian_case} in the equations
\eqref{eq_posterior_predictive_mean} and
\eqref{eq_posterior_predictive_covariance} we obtain the predictive
distribution
\begin{equation}\label{eq_posterior_predictive_in_gaussian_case}
\tilde{f}|\mathcal{D},\theta, \sigma^2 \sim \GP\left(m_{\text{p}}(\tilde{\x}), k_{\text{p}}(\tilde{\x},\tilde{\x}')\right),
\end{equation}
where the mean is $m_{\text{p}}(\tilde{\x}) = k(\tilde{\x},\mb{X})(\Kff +
\sigma^2\mb{I})^{-1}\y$ and covariance is $k_{\text{p}}(\tilde{\x},\tilde{\x}') =k(\tilde{\x},\tilde{\x}') -
k(\tilde{\x},\mb{X})(\Kff +
\sigma^2\mb{I})^{-1}k(\mb{X},\tilde{\x}')$. The predictive
distribution for new observations $\tilde{\y}$ can be obtained by
integrating $p(\tilde{\y}|\mathcal{D},\theta, \sigma^2) = \int
p(\tilde{\y}|\tilde{\f},\sigma^2)p(\tilde{\f}|\mathcal{D},\theta,
\sigma^2)d\tilde{\f}$. The result is, again, Gaussian with mean
$\E_{\tilde{\f}|\mathcal{D},\theta}[\tilde{\f}]$ and covariance
$\COV_{\tilde{\f}|\mathcal{D},\theta}[\tilde{\f}] + \sigma^2\mb{I}$. 

\subsection{Laplace approximation}

With a non-Gaussian likelihood the conditional posterior needs to be
approximated. The Laplace approximation is constructed from the second
order Taylor expansion of $\log p(\f|\y,\theta, \phi)$ around the mode
$\hat{\f}$, which gives a Gaussian approximation to the conditional
posterior
\begin{equation}\label{eq_latent_posterior_in_Laplace_case}
p(\f|\mathcal{D},\theta,\phi) \approx q(\f|\mathcal{D},\theta,\phi) =
\N(\f|\hat{\f}, \mb{\Sigma}), 
\end{equation}
where $\hat{\f}=\argmax_{\mb{f}} p(\f|\mathcal{D},\theta,\phi)$ and
$\mb{\Sigma}^{-1}$ is the Hessian of the negative log conditional posterior
at the mode \citep{Gelman+etal+BDA3:2013,Rasmussen+Williams:2006}: 
\begin{equation}
\mb{\Sigma}^{-1} = -\nabla\nabla \log
p(\f|\mathcal{D},\theta,\phi)|_{\f=\hat{\f}} = \iKff + \mb{W},
\label{Hessian} 
\end{equation}
where $\mb{W}$ is a diagonal matrix with entries $\mb{W}_{ii} =
\nabla_{f_i}\nabla_{f_i} \log p(y|f_i,\phi)|_{f_i=\hat{f}_i}$.
We call the approximation scheme Laplace method following
\citet{Williams+Barber:1998}, but essentially the same approximation
is named Gaussian approximation by \citet{Rue+Martino+Chopin:2009}.

Setting $\E_{\f|\mathcal{D},\theta}[\f] = \hat{\f}$ and
$\COV_{\f|\mathcal{D},\theta}[\f] = (\iKff + \mb{W})^{-1}$ into
\eqref{eq_posterior_predictive_mean} and
\eqref{eq_posterior_predictive_covariance} respectively, gives after
rearrangements and using $\iKff\hat{\f} = \nabla\log
p(\y|\f)|_{\f=\hat{\f}}$, the approximate posterior predictive
distribution
\begin{equation}\label{eq_posterior_in_Laplace_case}
\tilde{f}|\mathcal{D},\theta, \phi \sim \GP\left(m_{\text{p}}(\tilde{\x}), k_{\text{p}}(\tilde{\x},\tilde{\x}')\right).
\end{equation}
Here the mean and covariance are $m_{\text{p}}(\tilde{\x}) = k(\tilde{\x},\mb{X})\nabla\log
p(\y|\f)|_{\f=\hat{\f}}$ and 
$k_{\text{p}}(\tilde{\x},\tilde{\x}')=k(\tilde{\x},\tilde{\x}') - k(\tilde{\x},\mb{X})(\Kff +
\mb{W})^{-1}k(\mb{X},\tilde{\x}')$.
The approximate conditional predictive density of $\tilde{y}_i$ can
now be evaluated, for example, with quadrature integration over each
$\tilde{f}_i$ separately
\begin{equation}
p(\tilde{y}_i|\mathcal{D}, \theta, \phi) \approx \int
p(\tilde{y}_i|\tilde{f}_i,\phi)q(\tilde{f}_i|\mathcal{D}, \theta,
\phi) d \tilde{f}_i. 
\end{equation}
\subsection{Expectation propagation algorithm}

The Laplace method constructs a Gaussian approximation at the
posterior mode and approximates the posterior covariance via the
curvature of the log density at that point. The expectation
propagation (EP) algorithm \citep{Minka:2001}, for its part, tries to
minimize the Kullback-Leibler divergence from the true posterior to
its approximation
\begin{equation}\label{eq_EP_post}
 q(\f|\mathcal{D},
\theta,\phi) = \frac{1}{Z_{\text{EP}}}p(\f|\theta)\prod_{i=1}^n
t_i(f_i|\tilde{Z}_i, \tilde{\mu}_i,\tilde{\sigma}_i^2),
\end{equation}
where the likelihood terms have been replaced by site functions
$t_i(f_i|\tilde{Z}_i, \tilde{\mu}_i,\tilde{\sigma}_i^2) = \tilde{Z}_i
\N(f_i|\tilde{\mu}_i,\tilde{\sigma}_i^2)$ and the normalizing constant
by $Z_{\text{EP}}$. Detailed description of the algorithm is
provided, for example by \citet{Rasmussen+Williams:2006} and
\citet{Jylanki+Vanhatalo+Vehtari:2011}. EP's conditional posterior
approximation is
\begin{equation}\label{eq_posterior_in_EP_case}
q(\f|\mathcal{D},\theta,\phi) = \N(\f|\Kff(\Kff +
  \tilde{\Sigma})^{-1} \tilde{\mu}, \Kff -  \Kff(\Kff +
  \tilde{\Sigma})^{-1}\Kff),
\end{equation}
where $\tilde{\Sigma} =
\text{diag}[\tilde{\sigma}_1^{2},...,\tilde{\sigma}_n^{2}]$ and
$\tilde{\mu} = [\tilde{\mu}_1,...,\tilde{\mu}_n]^{\text{T}}$. The
predictive mean and covariance are again obtained from equations
\eqref{eq_posterior_predictive_mean} and
\eqref{eq_posterior_predictive_covariance} analogically to the Laplace
approximation.

From equations \eqref{eq_posterior_in_gaussian_case},
\eqref{eq_latent_posterior_in_Laplace_case}, and
\eqref{eq_posterior_in_EP_case} it can be seen that the Laplace and EP
approximations are similar to the exact solution with the Gaussian
likelihood. The diagonal matrices $\mb{W}^{-1}$ and
$\mb{\tilde{\Sigma}}$ correspond to the noise variance
$\sigma^2\mb{I}$ and, thus, these approximations can be interpreted as
Gaussian approximations to the likelihood
\citep{Nickisch+Rasmussen:2008}.

\subsection{Markov chain Monte
  Carlo}\label{sec_MCMC_for_conditional_of_latents}

The accuracy of the approximations considered so far is limited by
the Gaussian form of the approximating function. An approach, which
gives exact solution in the limit of an infinite computational time,
is the Monte Carlo integration \citep{Robert+Casella:2004}. This is
based on sampling from $p(\f|\mathcal{D}, \theta, \phi)$ and using the
samples to represent the posterior distribution. 

\begin{figure}
  \begin{center}
    \subfigure[Disease mapping]{
      \includegraphics[width=4cm]{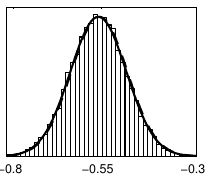}
    }
    ~
    \subfigure[Classification]{ 
      \includegraphics[width=4cm]{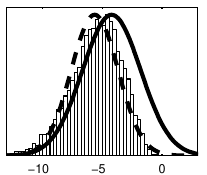}
    }
  \end{center}\caption{Illustration of the Laplace approximation
    (solid line), EP (dashed line) and MCMC (histogram) for the
    conditional posterior of a latent variable
    $p(f_i|\mathcal{D},\theta)$ in two applications. On the left, a
    disease mapping problem with Poisson likelihood \citep[used
    in][]{Vanhatalo+Pietilainen+Vehtari:2010} where the Gaussian
    approximation works well. On the right, a classification problem
    with probit likelihood \citep[used in][]{Vanhatalo+Vehtari:2010}
    where the posterior is skewed and the Gaussian approximation is
    not so good.}
  \label{latent_intergation} 
\end{figure}

In MCMC methods \citep{Gilks+Richardson+Spiegelhalter:1996}, one
constructs a Markov chain whose stationary distribution is the
posterior distribution and uses the Markov chain samples to obtain
Monte Carlo estimates. \pkg{GPstuff} provides, for example, a scaled
Metropolis Hastings algorithm \citep{Neal:1998} and Hamiltonian Monte
Carlo (HMC) \citep{Duane+Kennedy+Pendleton+Roweth:1987,Neal:1996a}
with variable transformation discussed in
\citep{Christensen+Roberts+Skold:2006,Vanhatalo+Vehtari:2007} to
sample from $p(\f|\mathcal{D}, \theta, \phi)$. The approximations to
the conditional posterior of $\f$ are illustrated in
Figure~\ref{latent_intergation}.

After having the posterior sample of latent variables, we can sample
from the posterior predictive distribution of any set of $\tilde{\f}$
simply by sampling with each $\f^{(i)}$ one $\tilde{\f}^{(i)}$ from
$p(\tilde{\f}|\f^{(i)},\theta,\phi)$, which is given in the equation
\eqref{eq_conditional_ftilde_given_f}. Similarly, we can obtain a
sample of $\tilde{\y}$ by drawing one $\tilde{\y}^{(i)}$ for each
$\tilde{\f}^{(i)}$ from $p(\tilde{\y}|\tilde{\f},\theta, \phi)$.

\section{Marginal likelihood given parameters}\label{sec_marginal_likelihood}

The marginal likelihood given the parameters,
$p(\mathcal{D}|\theta,\phi) = \int
p(\mb{y}|\f,\phi)p(\f|\mb{X},\theta) d\f$, is an important quantity
when inferring the parameters as discussed in the next section. With a
Gaussian likelihood it has an analytic solution
\eqref{eq_marginal_likelihood_gaussian_case} which gives the log
marginal likelihood
\begin{equation}\label{eq_log_marginal_likelihood}
\log p(\mathcal{D}|\theta,\sigma) =  -\frac{n}{2}\log(2\pi) -\frac{1}{2}
\log |\Kff + \sigma^2\mb{I}| -\frac{1}{2} \y^{\text{T}} (\Kff +
\sigma^2\mb{I})^{-1} \y. 
\end{equation}

If the likelihood is not Gaussian, the marginal likelihood needs to be
approximated. The Laplace approximation to the marginal likelihood is
constructed, for example, by making a second order Taylor expansion
for the integrand $p(\mb{y}|\f,\phi)p(\f|\mb{X},\theta)$ around
$\hat{\f}$. This gives a Gaussian integral over $\f$ multiplied by a
constant, and results in the log marginal likelihood approximation
\begin{equation}\label{eq_Laplace_marginal_likelihood}
\log p(\mathcal{D}|\theta,\phi) \approx \log
q(\mathcal{D}|\theta,\phi) \propto
-\frac{1}{2}\hat{\f}^{\text{T}}\iKff\hat{\f} + \log p(\mb{y}|\hat{\f},\phi)
- \frac{1}{2} \log |\mb{B}|,  
\end{equation}
where $|\mb{B}| = |I+\mb{W}^{1/2}\Kff \mb{W}^{1/2}|$. See also
\citep[][]{Tierney+Kadane:1986,Rue+Martino+Chopin:2009,Vanhatalo+Jylanki+Vehtari:2009}
for more discussion.

EP's marginal likelihood approximation is the normalization constant
\begin{equation}
Z_{\text{EP}} = \int p(\f|\mb{X}, \theta)\prod_{i=1}^n \tilde{Z}_i
\N(f_i|\tilde{\mu}_i,\tilde{\sigma}_i^2) d f_i 
\end{equation}
in equation \eqref{eq_EP_post}. This is a Gaussian integral multiplied
by a constant $\prod_{i=1}^n \tilde{Z}_i$, giving
\begin{align}\label{EP_marg_likelih}
\log p(\mathcal{D}|\theta,\phi) \approx \log Z_{\text{EP}} =&  -\frac{1}{2} \log|K+\tilde{\Sigma}|
-\frac{1}{2}\tilde{\mu}^{\text{T}} \left(K+\tilde{\Sigma}
\right)^{-1}\tilde{\mu} + C_{\text{EP}},
\end{align}
where $C_{\text{EP}}$ collects the terms that are not explicit
functions of $\theta$ or $\phi$ (there is an implicit dependence
through the iterative algorithm, though). For more discussion on EP's
marginal likelihood approximation see
\citep{Seeger:2005,Nickisch+Rasmussen:2008}.

\section{Marginalization over parameters}\label{sec_marginalization_over_hyperparam}

The previous section treated methods to evaluate exactly (the Gaussian
case) or approximately (Laplace and EP approximations) the log
marginal likelihood given parameters. Now, we describe approaches for
estimating parameters or integrating numerically over them.

\subsection{Maximum a posterior estimate of parameters}

In a full Bayesian approach we should integrate over all unknowns.
Given we have integrated over the latent variables, it often happens
that the posterior of the parameters is peaked or predictions are
unsensitive to small changes in parameter values.  In such case, we can
approximate the integral over $p(\theta,\phi|\mathcal{D})$ with the
maximum a posterior (MAP) estimate
\begin{equation}
\{\hat{\theta}, \hat{\phi}\} = \argmax_{\theta,\phi}
p(\theta,\phi|\mathcal{D}) = \argmin_{\theta,\phi}
\left[ -\log p(\mathcal{D}|\theta,\phi) - \log p(\theta,\phi) \right].
\end{equation}
In this approximation, the parameter values are given a point mass one
at the posterior mode, and the marginal of the latent function is
approximated as $p(\f|\mathcal{D}) \approx p(\f|\mathcal{D},
\hat{\theta}, \hat{\phi})$.

The log marginal likelihood, and its approximations, are
differentiable with respect to the parameters
\citep{Seeger:2005,Rasmussen+Williams:2006}. Thus, also the log
posterior is differentiable, which allows gradient based optimization.
The advantage of MAP estimate is that it is relatively easy and fast
to evaluate. According to our experience good optimization algorithms
need usually at maximum tens of optimization steps to find the
mode. However, it underestimates the uncertainty in parameters.

\subsection{Grid integration}\label{sec_grid_integration}

Grid integration is based on weighted sum of points evaluated on
grid
\begin{equation}\label{eq_grid_integration}
p(\f|\mathcal{D}) \approx \sum_{i=1}^M p(\f|\mathcal{D}, \vartheta_i)
p(\vartheta_i|\mathcal{D}) \Delta_i.
\end{equation}
Here $\vartheta = [\theta^{\text{T}}, \phi^{\text{T}}]^{\text{T}}$ and
$\Delta_i$ denotes the area weight appointed to an evaluation point
$\vartheta_i$. The implementation follows INLA
\citep{Rue+Martino+Chopin:2009} and is discussed in detail by
\citet{Vanhatalo+Pietilainen+Vehtari:2010}. The basic idea is to first
locate the posterior mode and then to explore the log posterior
surface so that the bulk of the posterior mass is included in the
integration (see Figure \ref{grid_based_integration}).
The grid search is feasible only for a small number of parameters
since the number of grid points grows exponentially with the dimension
of the parameter space $d$. 

\subsection{Monte Carlo integration}

Monte Carlo integration scales better than the grid integration in
large parameter spaces since its error decreases with a rate that is
independent of the dimension \citep{Robert+Casella:2004}.  There are
two options to find a Monte Carlo estimate for the marginal posterior
$p(\f|\mathcal{D})$. The first option is to sample only the parameters
from their marginal posterior $p(\vartheta|\mathcal{D})$ or from its
approximation (see Figure \ref{is_integration}). In this case, the
posterior marginal of the latent variable is approximated with mixture
distribution as in the grid integration. The alternative is to sample
both the parameters and the latent variables.

The full MCMC is performed by alternate sampling from the conditional
posteriors $p(\f|\mathcal{D},\vartheta)$ and
$p(\vartheta|\mathcal{D},\f)$. Possible choices to sample from the
conditional posterior of the parameters are, e.g., HMC, no-U-turn sampler \citep{Hoffman+Gelman:2014}, slice
sampling (SLS) \citep{Neal:2003,Thompson+Neal:2010}. Sampling both the parameters and
latent variables is usually slow since due to the strong correlation
between them
\citep{Vanhatalo+Vehtari:2007,Vanhatalo+Pietilainen+Vehtari:2010}.
Surrogate slice sampling by \citet{Murray+Adams:2010} can be used to
sample both the parameters and latent variables at the same time.

Sampling from the (approximate) marginal, $p(\vartheta|\mathcal{D})$,
is an easier task since the parameter space is smaller.
The parameters can be sampled from their marginal posterior (or its
approximation) with HMC, SLS \citep{Neal:2003,Thompson+Neal:2010} or via importance
sampling \citep{Geweke:1989}. In importance sampling, we use a
Gaussian or Student-$t$ proposal distribution $g(\vartheta)$ with mean
$\hat{\vartheta}$ and covariance approximated with the negative
Hessian of the log posterior, and approximate the integral with
\begin{equation}
p(\f|\mathcal{D}) \approx \frac{1}{\sum_{i=1}^M w_i} \sum_{i=1}^M
q(\f|\mathcal{D}, \vartheta_i) w_i, 
\end{equation}
where $w_i = q(\vartheta^{(i)})/g(\vartheta^{(i)})$ are the importance
weights. In some situations the naive Gaussian or Student-$t$ proposal
distribution is not adequate since the posterior distribution
$q(\vartheta|\mathcal{D})$ may be non-symmetric or the covariance
estimate is poor. An alternative for these situations is the scaled
Student-$t$ proposal distribution \citep{Geweke:1989} which is
adjusted along each main direction of the approximate covariance. The
implementation of the importance sampling is discussed in detail by
\citet{Vanhatalo+Pietilainen+Vehtari:2010}.

The problem with MCMC is that we are not able to draw independent
samples from the posterior. Even with a careful tuning of Markov chain
samplers the autocorrelation is usually so large that the required
sample size is in thousands, which is a clear disadvantage compared
with, for example, the MAP estimate.

\subsection{Central composite design}\label{sec_CCD_integration}

\begin{figure}
  \begin{center}
    \subfigure[Grid based ]{
      \includegraphics[width=4cm]{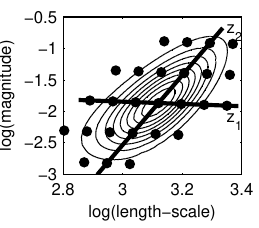}\label{grid_based_integration}
    }
    ~
    \subfigure[Monte Carlo ]{      
      \includegraphics[width=4cm]{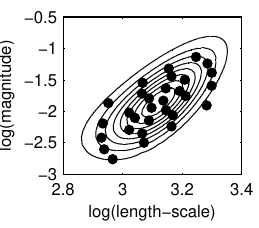}\label{is_integration}
    }
    \subfigure[Central composite design]{      
      \includegraphics[width=4cm]{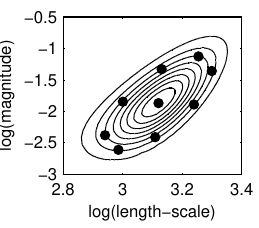}\label{ccd_integration}
    }
  \end{center}\caption{The grid based, Monte Carlo and central
    composite design integration. Contours show the posterior density
    $q(\log(\vartheta)|\mathcal{D})$ and the integration points are
    marked with dots. The left figure shows also the vectors $\mb{z}$
    along which the points are searched in the grid integration and
    central composite design. The integration is conducted over
    $q(\log(\vartheta)|\mathcal{D})$ rather than
    $q(\vartheta|\mathcal{D})$ since the former is closer to Gaussian.
    Reproduced from \citep{Vanhatalo+Pietilainen+Vehtari:2010}.}
  \label{hyperparameter_intergation}
\end{figure}

\citet{Rue+Martino+Chopin:2009} suggest a central composite design
(CCD) for choosing the representative points from the posterior of the
parameters when the dimensionality of the parameters, $d$, is moderate
or high. In this setting, the integration is considered as a quadratic
design problem in a $d$ dimensional space with the aim at finding
points that allow to estimate the curvature of the posterior
distribution around the mode. The design used in \pkg{GPstuff} is the
fractional factorial design \citep{Sanchez+Sanchez:2005} augmented
with a center point and a group of $2d$ star points.  The design
points are all on the surface of a $d$-dimensional sphere and the star
points consist of $2d$ points along each axis, which is illustrated in
Figure \ref{ccd_integration}.  The integration is then a finite sum
\eqref{eq_grid_integration} with special weights
\citep{Vanhatalo+Pietilainen+Vehtari:2010}.

CCD integration speeds up the computations considerably compared to
the grid search or Monte Carlo integration since the number of the
design points grows very moderately. The accuracy of the CCD is
between the MAP estimate and the full integration with the grid
search or Monte Carlo. \citet{Rue+Martino+Chopin:2009} report good
results with this integration scheme, and it has worked well in
moderate dimensions in our experiments as well. 
Since CCD is based on the assumption that the posterior of the
parameter is (close to) Gaussian, the densities
$p(\vartheta_i|\mathcal{D})$ at the points on the circumference should
be monitored in order to detect serious discrepancies from this
assumption. These densities are identical if the posterior is Gaussian
and we have located the mode correctly, and thereby great variability
on their values indicates that CCD has failed. The posterior of the
parameters may be far from a Gaussian distribution but for a suitable
transformation, which is made automatically in the toolbox, the
approximation may work well.

\section[Getting started with GPstuff: regression and classification]{Getting started with \pkg{GPstuff}: regression and classification}\label{chap:intro_to_GPstuff}

\subsection{Gaussian process regression}\label{sec:demo_regression1}

The demonstration program \code{demo\_regression1} considers a simple
regression problem $y_i = f(\x_i) + \epsilon_i$, where $\epsilon_i
\sim N(0,\sigma^2)$.  We will show how to construct the model with a
squared exponential covariance function and how to conduct the
inference.

\subsubsection{Constructing the model}

The model construction requires three steps: 1) Create structures that
define likelihood and covariance function, 2) define priors for the
parameters, and 3) create a GP structure where all the above are
stored. These steps are done as follows:
\begin{verbatim}
lik = lik_gaussian('sigma2', 0.2^2);
gpcf = gpcf_sexp('lengthScale', [1.1 1.2], 'magnSigma2', 0.2^2)

pn=prior_logunif();
lik = lik_gaussian(lik, 'sigma2_prior', pn);

pl = prior_unif();
pm = prior_sqrtunif();
gpcf = gpcf_sexp(gpcf, 'lengthScale_prior', pl, 'magnSigma2_prior', pm);

gp = gp_set('lik', lik, 'cf', gpcf);
\end{verbatim}
Here \code{lik\_gaussian} initializes Gaussian likelihood function and
its parameter values and \code{gpcf\_sexp} initializes the squared
exponential covariance function and its parameter values.
\code{lik\_gaussian} returns structure \code{lik} and
\code{gpcf\_sexp} returns \code{gpcf} that contain all the information
needed in the evaluations (function handles, parameter values etc.).
The next five lines create the prior structures for the parameters of
the observation model and the covariance function, which are set in the
likelihood and covariance function structures. 
The last line creates the GP structure by giving it the likelihood
and covariance function.

Using the constructed GP structure, we can evaluate basic summaries
such as covariance matrices, make predictions with the present
parameter values etc. For example, the covariance matrices $\Kff$
and $\mb{C} = \Kff+\sigma^2\mb{I}$ for three two-dimensional
input vectors are:
\begin{verbatim}
example_x = [-1 -1 ; 0 0 ; 1 1];
[K, C] = gp_trcov(gp, example_x)
K =
    0.0400    0.0187    0.0019
    0.0187    0.0400    0.0187
    0.0019    0.0187    0.0400
C =
    0.0800    0.0187    0.0019
    0.0187    0.0800    0.0187
    0.0019    0.0187    0.0800
\end{verbatim} 

\subsubsection{MAP estimate for the parameters}

\code{gp\_optim} works as a wrapper for usual gradient based
optimization functions. It is used as follows:
\begin{verbatim}
opt=optimset('TolFun',1e-3,'TolX',1e-3,'Display','iter');
gp=gp_optim(gp,x,y,'opt',opt);
\end{verbatim}
\code{gp\_optim} takes a GP structure, training input $\x$,
training target $\y$ (which are defined in
\code{demo\_regression1}) and options, and returns a GP structure
with parameter values optimized to their MAP estimate. By default
\code{gp\_optim} uses \code{fminscg} function, but \code{gp\_optim}
can use also, for example, \code{fminlbfgs} or \code{fminunc}.
Optimization options are set with \code{optimset} function. It is
also possible to set optimisation options as
\begin{verbatim}
opt=struct('TolFun',1e-3,'TolX',1e-3,'Display','iter');
\end{verbatim}
which is useful when using an optimiser not supported by \code{optimset}.
All the estimated parameter values can be easily checked using the
function \code{gp\_pak}, which packs all the parameter values from
all the covariance function structures in a vector, usually using
log-transformation (other transformations are also possible). The
second output argument of \code{gp\_pak} lists the labels for the
parameters:
\begin{verbatim}
[w,s] = gp_pak(gp);
disp(s), disp(exp(w))

    'log(sexp.magnSigma2)'
    'log(sexp.lengthScale x 2)'
    'log(gaussian.sigma2)'

    2.5981    0.8331    0.7878    0.0427
\end{verbatim}
It is also possible to set the parameter vector of the model to
desired values using \code{gp\_unpak}. \code{gp\_pak} and
\code{gp\_unpak} are used internally to allow use of generic
optimisation and sampling functions, which take the parameter
vector as an input argument.

\begin{figure}[]
    \begin{center}
      \subfigure[The predictive mean and training data.]{
        \label{}
        \includegraphics[width=6cm]{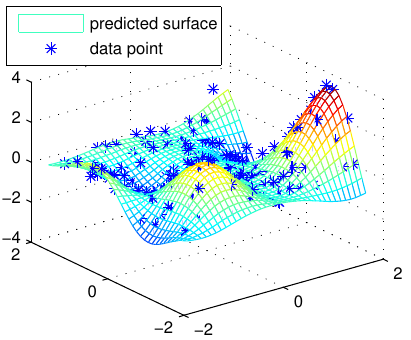}
      }
      ~
      \subfigure[The marginal posterior predictive distributions
      $p(f_i|\mathcal{D})$. ]{ 
        \label{}
        \includegraphics[width=5cm]{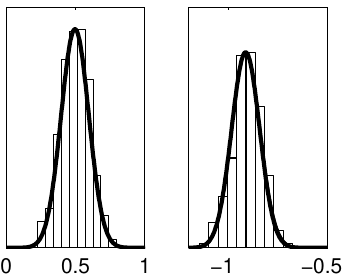}
       }
       \caption[]{The predictive mean surface, training data, and the
         marginal posterior for two latent variables in
         \code{demo\_regression1}. Histograms show the MCMC solution
         and the grid integration solution is drawn with a
         line.}\label{demo_regression1_fig1}
    \end{center}
\end{figure}

Predictions for new locations $\tilde{\x}$, given the training data
$(\x,\y)$, are done by \code{gp\_pred} function, which returns the
posterior predictive mean and variance for each $f(\tilde{\x})$ (see
equation \eqref{eq_posterior_predictive_in_gaussian_case}). This is
illustrated below where we create a regular grid where the posterior
mean and variance are computed. The posterior mean $m_{p}(\tilde{\x})$
and the training data points are shown in
Figure~\ref{demo_regression1_fig1}.

\begin{verbatim}
[xt1,xt2]=meshgrid(-1.8:0.1:1.8,-1.8:0.1:1.8);
xt=[xt1(:) xt2(:)];
[Eft_map, Varft_map] = gp_pred(gp, x, y, xt);
\end{verbatim}

\subsubsection{Marginalization over parameters with grid integration}

To integrate over the parameters we can use any method
described in the section \ref{sec_marginalization_over_hyperparam}.
The grid integration is performed with the following line:
\begin{verbatim}
[gp_array, P_TH, th, Eft_ia, Varft_ia, fx_ia, x_ia] = ...
                              gp_ia(gp, x, y, xt, 'int_method', 'grid');
\end{verbatim}
\code{gp\_ia} returns an array of GPs (\code{gp\_array}) for parameter
values \code{th} ($[\vartheta_i]_{i=1}^M$) with weights \code{P\_TH}
($[p(\vartheta_i|\mathcal{D}) \Delta_i]_{i=1}^M)$. Since we use the
grid method the weights are proportional to the marginal posterior and
$\Delta_i \equiv 1 \forall i$ (see section
\ref{sec_grid_integration}).  \code{Ef\_ia} and \code{Varf\_ia}
contain the predictive mean and variance at the prediction locations.
The last two output arguments can be used to plot the predictive
distribution $p(\tilde{f}_i|\mathcal{D})$ as demonstrated in Figure
\ref{demo_regression1_fig1}.  \code{x\_ia(i,:)} contains a regular
grid of values $\tilde{f}_i$ and \code{fx\_ia(i,:)} contains
$p(\tilde{f}_i|\mathcal{D})$ at those values.

\subsubsection{Marginalization over parameters with MCMC}

The main function for conducting Markov chain sampling is \code{gp\_mc},
which loops through all the specified samplers in turn and saves the
sampled parameters in a record structure. In later sections, we will
discuss models where also latent variables are sampled, but now we
concentrate on the covariance function parameters, which are sampled
as follows:
\begin{verbatim}
[gp_rec,g,opt] = gp_mc(gp, x, y, 'nsamples', 220);
gp_rec = thin(gp_rec, 21, 2);
[Eft_s, Varft_s] = gpmc_preds(rfull, x, y, xt);
[Eft_mc, Varft_mc] = gp_pred(gp_rec, x, y, xt);
\end{verbatim}
The \code{gp\_mc} function generates \code{nsamples} (here 220) Markov
chain samples. At each iteration \code{gp\_mc} runs the actual
samplers.  The function \code{thin} removes the burn-in from the
sample chain (here 21) and thins the chain more (here by 2). This way
we can decrease the autocorrelation between the remaining samples.
\pkg{GPstuff} provides also diagnostic tools for Markov chains. See,
for example,
\citep{Gelman+etal+BDA3:2013,Robert+Casella:2004} for
discussion on convergence and other Markov chain diagnostics. The
function \code{gpmc\_preds} returns the conditional predictive mean
and variance for each sampled parameter value. These are
$\E_{p(\mb{f}|\mb{X}, \mathcal{D}, \mb{\vartheta}^{(s)})}[\tilde{\f}],
s=1,...,M$ and $\VAR_{p(\mb{f}|\mb{X}, \mathcal{D},
  \mb{\vartheta}^{(s)})}[\tilde{\f}], s=1,...,M$, where $M$ is the
number of samples.
Marginal predictive mean and variance are computed directly with
\code{gp\_pred}.

\subsection{Gaussian process classification}\label{GP_classification}


We will now consider a binary GP classification (see
\code{demo\_classific}) with observations, $y_i \in \{-1,1\},
i=1,...,n$, associated with inputs $\mb{X} = \{ \mb{x} \}_ {i=1}^n$.
The observations are considered to be drawn from a Bernoulli
distribution with a success probability $p(y_i=1|\mb{x}_i)$. The
probability is related to the latent function via a sigmoid function
that transforms it to a unit interval. \pkg{GPstuff} provides a probit
and logit transformation, which give
\begin{align}\label{eq_probit_likelihood}
p_{\text{probit}}(y_i|f(\x_i)) &= \Phi(y_if(\mb{x}_i)) =
\int_{-\infty}^{y_if(\mb{x}_i)} N(z|0,1) d z \\
p_{\text{logit}}(y_i|f(\x_i)) &= \frac{1}{1 + \exp(-y_if(\x_i))}.\label{eq_logit_likelihood}
\end{align}
Since the likelihood is not Gaussian we need to use approximate
inference methods, discussed in the section
\ref{sec_cond_post_of_latent}.

\subsubsection{Constructing the model}

The model construction for the classification follows closely the
steps presented in the previous section. The model is constructed
as follows:
\begin{verbatim}
lik = lik_probit();
gpcf = gpcf_sexp('lengthScale', [0.9 0.9], 'magnSigma2', 10);
gp = gp_set('lik', lik, 'cf', gpcf, 'jitterSigma2', 1e-9);
\end{verbatim}
The above lines first initialize the likelihood function, the
covariance function and the GP structure. Since we do not specify
parameter priors, the default priors are used. The model construction
and the inference with logit likelihood (\code{lik\_logit}) would be
similar with probit likelihood. A small jitter value is added to the
diagonal of the training covariance to make certain matrix operations
more stable (this is not usually necessary but is shown here for
illustration).

\subsubsection{Inference with Laplace approximation}\label{sec_classific_Laplace}

The MAP estimate for the parameters can be found using
\code{gp\_optim} as
\begin{verbatim}
gp = gp_set(gp, 'latent_method', 'Laplace');
gp = gp_optim(gp,x,y,'opt',opt);
\end{verbatim}
The first line defines which inference method is used for the latent
variables. It initializes the Laplace algorithm and sets needed fields
in the GP structure. The default method for latent variables is
Laplace, so this line could be omitted.
\code{gp\_optim} uses the default optimization function,
\code{fminscg}, with the same options as above in the regression
example.

\code{gp\_pred} provides the mean and variance for the latent
variables (first two outputs), the log predictive probability for a
test observation (third output), and mean and variance for the
observations (fourth and fifth output) at test locations \code{xt}.
\begin{verbatim}
[Eft_la, Varft_la, lpyt_la, Eyt_la, Varyt_la] = ... 
                           gp_pred(gp, x, y, xt, 'yt', ones(size(xt,1),1) );
\end{verbatim}
The first four input arguments are the same as in the section
\ref{sec:demo_regression1}. The fifth and sixth arguments are a
parameter-value pair where \code{yt} tells that we give test
observations \code{ones(size(xt,1),1)} related to \code{xt} as an
optional input, in which case \code{gp\_pred} evaluates their marginal
posterior log predictive probabilities \code{lpyt\_la}. Here we evaluate
the probability to observe class $1$ and thus we give a vector of ones
as test observations. 

\subsubsection{Inference with expectation propagation}

EP works as the Laplace approximation. We only need to set the latent
method to \code{'EP'} but otherwise the commands are the same as above:
\begin{verbatim}
gp = gp_set(gp, 'latent_method', 'EP');
gp = gp_optim(gp,x,y,'opt',opt);
[Eft_ep, Varft_ep, lpyt_ep, Eyt_ep, Varyt_ep] = ...
                    gp_pred(gp, x, y, xt, 'yt', ones(size(xt,1),1) );
\end{verbatim}

\subsubsection{Inference with MCMC}

With MCMC we sample both the latent variables and the parameters:
\begin{verbatim}
gp = gp_set(gp, 'latent_method', 'MCMC', 'jitterSigma2', 1e-6);
[gp_rec,g,opt]=gp_mc(gp, x, y, 'nsamples', 220);
gp_rec=thin(gp_rec,21,2);
[Ef_mc, Varf_mc, lpy_mc, Ey_mc, Vary_mc] = ...
    gp_pred(gp_rec, x, y, xt, 'yt', ones(size(xt,1),1) );
\end{verbatim}
For MCMC we need to add a larger jitter on the diagonal to prevent
numerical problems.
By default sampling from the latent value distribution $p(\f|\theta,
\mathcal{D})$ is done with the elliptical slice sampling
\citep{Murray+Adams+MacKay:2010}. Other samplers provided by the
\pkg{GPstuff} are a scaled Metropolis Hastings algorithm
\citep{Neal:1998} and a scaled HMC \code{scaled\_hmc}
\citep{Vanhatalo+Vehtari:2007}. From the user point of view, the
actual sampling and prediction are performed similarly as with the
Gaussian likelihood.  The \code{gp\_mc} function handles the sampling
so that it first samples the latent variables from
$p(\f|\theta,\mathcal{D})$ after which it samples the parameters from
$p(\theta|\f,\mathcal{D})$.  This is repeated until \code{nsamples}
samples are drawn.

In classification model MCMC is the most accurate inference method,
then comes EP and Laplace approximation is the worst.  However, the
inference times line up in the opposite order.  The difference between
the approximations is not always this large.  For example, with
Poisson likelihood, discussed in the section \ref{sec_spatial_demo1},
Laplace and EP approximations work, in our experience, practically as
well as MCMC.

\subsubsection{Marginal posterior corrections}

\pkg{GPstuff} also provides methods for marginal posterior corrections using either Laplace or EP as a latent method \citep{Cseke+Heskes:2011}. Provided correction methods are either \texttt{cm2} or \texttt{fact} for Laplace and \texttt{fact} for EP. These methods are related to methods proposed by \citet{Tierney+Kadane:1986} and \citet{Rue+Martino+Chopin:2009}.

Univariate marginal posterior corrections can be computed with function \texttt{gp\_predcm} which returns the corrected marginal posterior distributtion for the given indices of input/outputs:
\begin{verbatim}
  [pc, fvec, p] = gp_predcm(gp,x,y,'ind', 1, 'fcorr', 'fact');
\end{verbatim}
The returned value \texttt{pc} corresponds to corrected marginal posterior for the latent value $f_1$, \texttt{fvec} corresponds to vector of grid points where \texttt{pc} is evaluated and \texttt{p} is the original Gaussian posterior distribution evaluated at \texttt{fvec}. Marginal posterior corrections can also be used for predictions in \texttt{gp\_pred} and sampling of latent values in \texttt{gp\_rnd} (see \texttt{demo\_improvedmarginals1} and \texttt{demo\_improvedmarginals2}). \texttt{gp\_rnd} uses univariate marginal corrections with Gaussian copula to sample from the joint marginal posterior of several latent values.

\section{Other single latent models}

In this section, we desribe other likelihood functions, which
factorize so that each factor depend only on single latent value.
 
\subsection{Robust regression}

\subsubsection{Student-$t$ observation model}

\begin{figure}
  \begin{center}
    \subfigure[Gaussian model.]{
      \includegraphics[width=5cm]{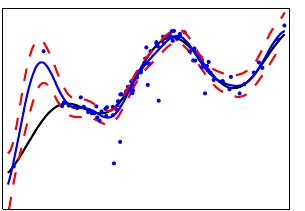}\label{neal_data_pic1}
    }    
    ~
    \subfigure[Student-$t$ model.]{ 
      \includegraphics[width=5cm]{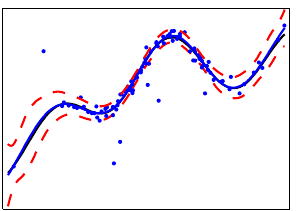}\label{neal_data_pic2}
    }
  \end{center}\caption{An example of regression with outliers from
    \citep{Neal:1997}. On the left Gaussian and on the right the
    Student-$t$ model.  The real function is plotted with
    black line.}
  \label{single_obs_figure} 
\end{figure}

A commonly used observation model in the GP regression is the Gaussian 
distribution. This is convenient since the marginal likelihood is
analytically tractable. However, a known limitation with the Gaussian
observation model is its non-robustness, due which outlying
observations may significantly reduce the accuracy of the inference
(see Figure~\ref{single_obs_figure}).  A well-known robust observation
model is the Student-$t$ distribution \citep{OHagan:1979}
\begin{equation}
 \y | \f, \nu, \sigma_t  \sim \prod_{i=1}^n \frac{\Gamma((\nu+1)/2)}{\Gamma(\nu/2)\sqrt{\nu\pi}\sigma_t}\left(1 +
  \frac{(y_i-f_i)^2}{\nu\sigma_t^2} \right)^{-(\nu+1)/2},
\end{equation}
where $\nu$ is the degrees of freedom and $\sigma_t$ the scale
parameter.  The Student-$t$ distribution can be utilized as such or it
can be written via the scale mixture representation
\begin{align}
y_i | f_i,\alpha, U_i & \sim N(f_i, \alpha U_i) \label{eq_scale_mixture_1}\\
U_i & \sim \text{Inv-}\chi^2(\nu, \tau^2), \label{eq_scale_mixture_2}
\end{align}
where each observation has its own noise variance $\alpha U_i$ that is
$\text{Inv-}\chi^2$ distributed
\citep{Neal:1997,Gelman+etal+BDA3:2013}. The degrees of freedom $\nu$
corresponds to the degrees of freedom in the Student-$t$ distribution 
and $\alpha\tau$ corresponds to $\sigma_t$.

In \pkg{GPstuff} both of the representations are implemented. The
scale mixture representation is implemented in
\code{lik\_gaussiansmt} and can be inferred only with MCMC \citep[as
described by][]{Neal:1998}. The Student-$t$ model is
implemented in \code{lik\_t} and can be inferred with Laplace and EP
approximation and MCMC \citep[as described
by][]{Vanhatalo+Jylanki+Vehtari:2009,Jylanki+Vanhatalo+Vehtari:2011}.
These are demonstrated in \code{demo\_regression\_robust}.

\subsubsection{Laplace observation model}

Besides the Student-T observation model, Laplace distribution is another robust alternative to the normal 
Gaussian observation model
\begin{equation}
 \y | \f, \sigma  \sim \prod_{i=1}^n \left( 2\sigma \right)^{-1} \exp \left( -\frac{|y_i - f_i|}{\sigma} \right),
\end{equation}
where $\sigma$ is the scale parameter of the Laplace distribution. 

In \pkg{GPstuff}, the Laplace observation model implementation is in \code{lik\_laplace}. It should be noted that since the Laplace distribution is 
not differentiable with respect to $\f$ in the mode, only EP approximation or MCMC can be used for inference. 
\subsection{Count data}\label{sec_spatial_demo1}

Count data are observed in applications. One such common application
is spatial epidemiology, which concerns both describing and
understanding the spatial variation in the disease risk in
geographically referenced health data. One of the most common tasks in
spatial epidemiology is disease mapping, where the aim is to describe
the overall disease distribution on a map and, for example, highlight
areas of elevated or lowered mortality or morbidity risk
\citep[e.g.][]{Lawson:2001,Richardson:2003,Elliot+Wakefield+Best+Briggs:2001}.
Here we build a disease mapping model following the general approach
discussed, for example, by \citet{Best+Richardson+Thomson:2005}. The
data are aggregated into areas with coordinates $\x_i$. The
mortality/morbidity in an area is modeled with a Poisson, negative
binomial or binomial distribution with mean $e_i\mu_i$, where $e_i$ is
the standardized expected number of cases
\citep[e.g.][]{Ahmad+all:2000}, and $\mu_i$ is the relative risk,
whose logarithm is given a GP prior. The aim is to infer the relative
risk. The implementation of these models in \pkg{GPstuff} is discussed
in \citep{Vanhatalo+Vehtari:2007,Vanhatalo+Pietilainen+Vehtari:2010}.

\subsubsection{Poisson}

\begin{figure}[]
  \begin{center}
    \subfigure[The posterior mean.]{
      \label{}
      \includegraphics[width=4.2cm]{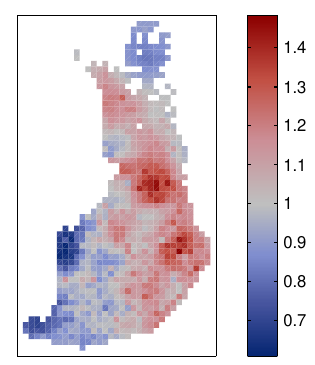}
    }
    ~
    \subfigure[The posterior variance.]{ 
      \label{}
      \includegraphics[width=4.4cm]{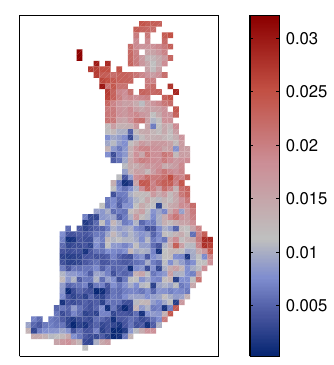}
    }
    \caption[]{The posterior predictive mean and variance of the
      relative risk in the \code{demo\_spatial1} data set obtained
      with FIC.}\label{demo_spatial1_fig1}
  \end{center}
\end{figure}

The Poisson model is implemented in \code{likelih\_poisson} and
it is
\begin{equation}
 \y|\f,\mb{e}  \sim \prod_{i=1}^{n} \Poisson(y_i|\exp(f_i)e_i) \label{Poisson_likelihood}
\end{equation}
where the vector $\mb{y}$ collects the numbers of deaths for each
area. Here $\mu=\exp(f)$ and its posterior predictive mean and
variance solved in the demo \code{demo\_spatial1} are shown in
Figure~\ref{demo_spatial1_fig1}.

\subsubsection{Negative binomial}\label{sec_spatial_demo2}

The negative binomial distribution is a robust version of the Poisson
distribution similarly as Student-$t$ distribution can be considered
as a robustified Gaussian distribution
\citep{Gelman+etal+BDA3:2013}. In \pkg{GPstuff} it is
parametrized as
\begin{equation}
 \y |\f,\mb{e}, r  \sim \prod_{i=1}^n
 \frac{\Gamma(r+y_i)}{y_i!\Gamma(r)}
\left(\frac{r}{r+\mu_i}\right)^r \left(\frac{\mu_i}{r+\mu_i}\right)^{y_i},
\end{equation}
where $\mu_i = e_i\exp(f(\x_i))$ and $r$ is the dispersion parameter
governing the variance. The model is demonstrated in
\code{demo\_spatial2}.

\subsubsection{Binomial}

In disease mapping, the Poisson distribution is used to approximate
the true binomial distribution. This approximation works well if the
background population, corresponding to the number of trials in the
binomial distribution, is large. Sometimes this assumption is not
adequate and we need to use the exact binomial observation model
\begin{equation}
\y|\f,\mathbf{z} \sim \prod_{i=1}^n \frac{z_i!}{y_i!(z_i-y_i)!} p_i^{y_i}(1-p_i)^{(z_i-y_i)},
\end{equation}
where $p_i = \exp(f(\x_i))/ (1+\exp(f(\x_i)))$ is the probability of
success, and the vector $\mathbf{z}$ denotes the number of trials. The
binomial observation model is not limited to spatial modeling but is
an important model for other problems as well. The observation model
is demonstrated in \code{demo\_binomial1} with a one dimensional
simulated data and \code{demo\_binomial\_apc} demonstrates the model in an
incidence risk estimation.

\subsubsection{Hurdle model}
\label{sec_hurdle}

Hurdle models can be used to model excess number of zeros compared
to usual Poisson and negative binomial count models. Hurdle models
assume a two-stage process, where the first process determines
whether the count is larger than zero, and the second process
determines the non-zero count \citep{Mullahy:1986}. These processes
factorize, and thus hurdle model can be implemented using two
independent GPs in GPstuff. \code{lik\_probit} or \code{lik\_logit}
can be used for the zero process and \code{lik\_negbinztr} can be
used for the count part. \code{lik\_negbinztr} provides zero
truncated negative binomial model, which can be used also to
approximate zero-truncated Poisson model by using high dispersion
parameter value. Construction of a hurdle model is demonstrated in
\code{demo\_hurdle}. Gaussian process model with logit negative
binomial hurdle model implemented using \pkg{GPstuff} was used in
reference \citep{Rantonen+etal:2011} to model sick absence days due
to low back symptoms.

Alternative way to model excess number of zeros is to couple the zero
and count processes as in zero-inflated negative binomial model
described in section~\ref{sec_zinegbin}.

\subsection{Log-Gaussian Cox process}

Log-Gaussian Cox-process is an inhomogeneous Poisson process model
used for point data, with unknown intensity function $\lambda(\x)$,
modeled with log-Gaussian process so that $f(\x)=\log \lambda(\x)$
\citep[see][]{Rathbun+Cressie:1994,Moller+Syversveen+Waagepetersen:1998}.
If the data are points $\mb{X}=\x_i;$ $i=1,2,\ldots,n$ on a finite region
$\mathcal{V}$ in $\Re^d$, then the likelihood of the unknown
function $f$ is
\begin{align}
  p(\mb{X}|f)=\exp\left\{-\left(\int_{\mathcal{V}} \exp(f(\x)) d\x
\right)+\sum_{i=1}^{n} f(\x_i) \right\}.
\end{align}
Evaluation of the likelihood would require nontrivial integration over
the exponential of GP.  \citet{Moller+Syversveen+Waagepetersen:1998}
propose to discretise the region $\mathcal{V}$ and assume locally
constant intensity in subregions. This transforms the problem to a
form equivalent to having Poisson model for each subregion. Likelihood
after the discretisation is
\begin{align}
  p(\mb{X}|f)\approx \prod_{k=1}^K \Poisson(y_k|\exp(f(\dot{\x}_k))),
\end{align}
where $\dot{\x}$ is the coordinate of the $k$th sub-region and $y_k$
is the number of data points in it. \citet{Tokdar+Ghosh:2007} proved
the posterior consistency in limit when sizes of subregions go to
zero.

\begin{figure}[]
  \begin{center}
    \subfigure[Coal mine disasters.]{
      \label{}
      \includegraphics[width=5.5cm]{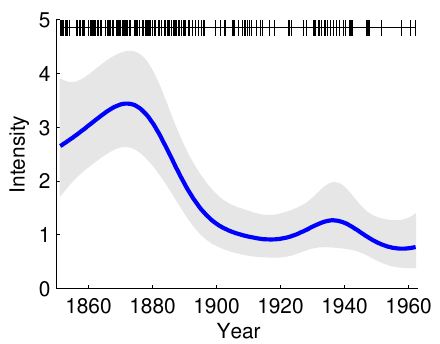}
    }
    ~
    \subfigure[Redwood data.]{ 
      \label{}
      \includegraphics[width=5.5cm]{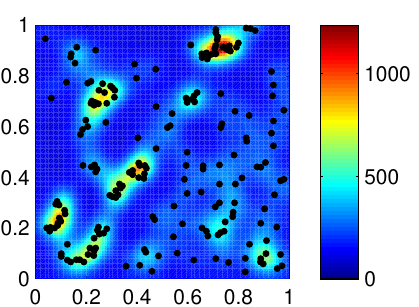}
    }
    \caption[]{Two intensity surfaces estimated with log-Gaussian Cox
      process. The figures are from the \code{demo\_lgcp}, where the
      aim is to study an underlying intensity surface of a point
      process. On the left a temporal and on the right a spatial point
      process.}\label{fig_demo_lgcp}
  \end{center}
\end{figure}

The log-Gaussian Cox process with Laplace and EP approximation is
implemented in the function \code{lgcp} for one or two dimensional
input data. The usage of the function is demonstrated in
\code{demo\_lgcp}. This demo analyzes two data sets.  The first one is
one dimensional case data with coal mine disasters (from R
distribution). The data contain the dates of 191 coal mine explosions
that killed ten or more men in Britain between 15 March 1851 and 22
March 1962. The analysis is conducted using expectation propagation
and CCD integration over the parameters and the results are shown
in Figure \ref{fig_demo_lgcp}. The second data are the redwood data
(from R distribution). This data contain 195 locations of redwood
trees in two dimensional lattice. The smoothed intensity surface is
shown in Figure \ref{fig_demo_lgcp}.

\subsection{Accelerated failure time survival models}

The accelerated failure time survival models are demonstrated in
\code{demo\_survival\_aft}. These models can be used to model also
other positive continuous data with or without censoring.

\subsubsection{Weibull}
\label{sec_weibull_demo}

The Weibull distribution is a widely used parametric baseline hazard
function in survival analysis \citep{Ibrahim+Chen+Sinha:2001}. The
hazard rate for the observation $i$ is
\begin{equation}
  h_i(y)=h_0(y)\exp(f(\x_i)),
\end{equation}
where $y>0$ and the baseline hazard $h_0(y)$ is assumed to follow the
Weibull distribution parametrized in \pkg{GPstuff} as
\begin{equation}
  h_0(y) = r y^{r-1},
\end{equation}
where $r>0$ is the shape parameter. This follows the parametrization
used in \citet{Martino:2011}. The likelihood defined as
\begin{equation}
  L = \prod_{i=1}^n r^{1-z_i} \exp \left( (1-z_i)f(\x_i)+(1-z_i)(r-1)\log(y_i)-\exp(f(\x_i))y_i^r \right),
\end{equation}
where $\mathbf{z}$ is a vector of censoring indicators with $z_i = 0$ for
uncensored event and $z_i = 1$ for right censored event for
observation $i$. Here we present only the likelihood function because we don't have observation model for the censoring. 

\subsubsection{Log-Gaussian}
With Log-Gaussian survival model the logarithms of survival times are assumed to be normally distributed.

The likelihood is defined as
\begin{eqnarray}
  L = && \prod_{i=1}^n (2\pi \sigma^2)^{-(1-z_i)/2}y_i^{1-z_i} \exp \left(-\frac{1}{2\sigma^2}(1-z_i)(\log (y_i) - f(\x_i))^2\right) \\ \nonumber
  &\times & \left(1 - \Phi \left(\frac{\log(y_i) - f(\x_i)}{\sigma}\right)\right)^{z_i} 
\end{eqnarray}
where $\sigma$ is the scale parameter.

\subsubsection{Log-logistic}

The log-logistic likelihood is defined as

\begin{equation}
  L = \prod_{i=1}^n \left( \frac{ry^{r-1}}{\exp(f(\x_i))} \right)^{1-z_i} \left( 1 + \left(\frac{y}{\exp(f(\x_i))}\right)^r \right)^{z_i-2},
\end{equation}
where $r$ is the shape parameter and $z_i$ the censoring indicators.

\subsection{Derivative observations in GP regression}

Incorporating derivative observations in GP regression is fairly
straightforward, because a derivative of Gaussian process is a
Gaussian process. In short, derivative observation are taken into
account by extending covariance matrices to include derivative
observations. This is done by forming joint covariance matrices of
function values and derivatives. Following equations
\citep{Rasmussen+Williams:2006} state how the covariances between
function values and derivatives, and between derivatives are
calculated
\begin{equation}
  \COV(f_i,\frac{\partial f_j}{\partial x_{dj}}) = \frac{\partial k(\textbf{x}_i,\textbf{x}_j}{\partial x_{dj}}), \qquad
  \COV(\frac{\partial f_i}{\partial x_{di}},\frac{\partial f_j}{\partial x_{ej}}) = \frac{\partial^2 k(\textbf{x}_i,\textbf{x}_j}{\partial x_{di}\partial x_{ej}}.\nonumber
\end{equation}
The joint covariance matrix for function values and derivatives is of
the following form
\begin{eqnarray}
  \textbf{K} &=&
  \left[
    \begin{array}{cc}
      \textbf{K}_{ff} & \textbf{K}_{fD}\\
      \textbf{K}_{Df}& \textbf{K}_{DD}
    \end{array}\nonumber
  \right]\\
  \nonumber\\
  \textbf{K}_{ff}^{ij} &=& k(\textbf{x}_i,\textbf{x}_j),\nonumber\\
  \textbf{K}_{Df}^{ij} &=& \frac{\partial k(\textbf{x}_i,\textbf{x}_j)}{\partial x_{di}},\nonumber\\
  \textbf{K}_{fD} &=& (\textbf{K}_{Df})^\top,\label{DerKder3} \nonumber\\
  \textbf{K}_{DD}^{ij} &=& \frac{\partial^2 k(\textbf{x}_i,\textbf{x}_j)}{\partial x_{di} \partial x_{ej}},\nonumber
\end{eqnarray}
Prediction is done as usual but with derivative observations joint
covariance matrices are to be used instead of the normal ones.

Using derivative observations in GPstuff requires two steps: when
initializing the GP structure one must set option
\code{'derivobs'} to \code{'on'}. The second step is to form
right sized observation vector. With input size $n \times m$ the
observation vector with derivatives should be of size $n + m \cdot
n$. The observation vector is constructed by adding partial
derivative observations after function value observations
\begin{equation}
  \textbf{y}_{obs} =
  \left[
    \begin{array}{c}
      y(\textbf{x})\\
      \frac{\partial y(\textbf{x})}{\partial x_1} \\
      \vdots\\
      \frac{\partial y(\textbf{x})}{\partial x_m}
    \end{array}
  \right].
\end{equation}
Different noise level could be assumed for function values and
derivative observations but at the moment the implementation allows
only same noise for all the observations. The use of derivative
observations is demonstrated in \code{demo\_derivativeobs}.

Derivative process can be also used to construct a monotonicity constraint as described in section~\ref{sec_monotonic}.

\subsection{Quantile Regression}

Quantile regression is used for estimating quantiles of the response
variable as a function of input variables \citep{Boukouvalas:2012}.

The likelihood is defines as
\begin{align}
p(y | f, \sigma, \tau) = \frac{\tau(1-\tau)}{\sigma}\exp\left[-\frac{y-f}{\sigma}(\tau - I(y \le f))\right],
\end{align}
where $\tau$ is the quantile of interest and $\sigma$ is the standard
deviation. $I(y \le f)$ is 1 if the condition inside brackets is true
and 0 otherwise. Because the logarithm of the likelihood is not twice
differentiable at the mode, Laplace approximation cannot be used for
inference with Quantile Regression. Quantile Regression with EP and
MCMC is demonstrated in \code{demo\_qgp} with toy data.

\section{Multilatent models}
\label{sec:multilatent-models}

The multilatent models consist of models where individual likelihood
factors depend on multiple latent variables. In this section we
shortly summarize such models in GPstuff.

\subsection{Multiclass classification}

In multiclass classification problems the target variables have more
than two possible class labels, $y_i\in\{1,\ldots,c\}$, where $c>2$ is
the number of classes. In \pkg{GPstuff}, multi-class classification
can be made either using the softmax likelihood (\code{lik\_softmax})
\begin{eqnarray}
p(y_i|\f_i)=\frac{\exp(f_i^{y_i})}{\sum_{j=1}^c\exp(f_i^{j})},
\end{eqnarray}
where $\f_i=\left[f_i^1,\ldots,f_i^c\right]^T$, or the multinomial
probit likelihood  (\code{lik\_multinomialprobit})
\begin{eqnarray}
p(y_i|\f_i)=\mathrm{E}_{p(u_i)} \left\{\prod_{j=1,j\neq y_i}^c \Phi (u_i+f^{y_i}_i-f^j_i)
\right\},
\end{eqnarray}
where the auxiliary variable $u_i$ is distributed as
$p(u_i)=\mathcal{N}(u_i|0,1)$, and $\Phi(x)$ denotes the cumulative
density function of the standard normal distribution.
In Gaussian process literature for multiclass classification, a common
assumption is to introduce $c$ independent prior processes that are
associated with $c$ classes \citep[see, e.g.,][]{Rasmussen+Williams:2006}.
By assuming zero-mean Gaussian processes for latent functions
associated with different classes, we obtain a zero-mean Gaussian
prior
\begin{eqnarray}
  p(\f|X)=\mathcal{N}(\f|\mathbf{0}, K),
\end{eqnarray}
where
$\f=\left[f_1^1,\ldots,f_n^1,f_1^2,\ldots,f_n^2,\ldots,f_1^c,\ldots,f_n^c\right]^T$
and $K$ is a $cn \times cn$ block-diagonal covariance matrix with
matrices $K^1, K^2,\ldots,K^c$ (each of size $n \times n$) on its
diagonal.

Inference for softmax can be made with MCMC or Laplace approximation.
As described by \citet{Williams+Barber:1998} and
\citet{Rasmussen+Williams:2006}, using the Laplace approximation for
the softmax likelihood with the uncorrelated prior processes, the
posterior computations can be done efficiently in a way that scales
linearly in $c$, which is also implemented in
\pkg{GPstuff}. Multiclass classification with softmax is demonstrated
in \code{demo\_multiclass}.

Inference for multinomial probit can be made with expectation
propagation by using a nested EP approach that does not require
numerical quadratures or sampling for estimation of the tilted moments
and predictive probabilities, as proposed by \citet{Riihimaki+Jylanki+Vehtari:2013}.
Similarly to softmax with Laplace's method, the nested EP approach
leads to low-rank site approximations which retain all posterior
couplings but results in linear computational scaling with respect to
$c$.
Multiclass classification with the multinomial probit likelihood is
demonstrated in \code{demo\_multiclass\_nested\_ep}.

\subsection{Multinomial}

The multinomial model (\code{lik\_multinomial}) is an extension of the
multiclass classification to situation where each observation consist
of counts of class observations. Let $y_i = [y_{i,1},\ldots,y_{i,c}]$,
where $c>2$, be a vector of counts of class observations related to
inputs $x_i$ so that
\begin{equation}
y_i \sim \text{Multinomial}([p_{i,1},\ldots,p_{i,c}],n_i)
\end{equation}
where $n_i=\sum_{j=1}^c y_{i,j}$. The propability to observe class $j$ is
\begin{eqnarray}
p_{i,j}=\frac{\exp(f_i^j)}{\sum_{j=1}^c\exp(f_i^j)},
\end{eqnarray}
where $\f_i=\left[f_i^1,\ldots,f_i^c\right]^T$. As in multiclass
classification we can introduce $c$ independent prior processes that
are associated with $c$ classes.
By assuming zero-mean Gaussian processes for latent functions
associated with different classes, we obtain a zero-mean Gaussian
prior
\begin{eqnarray}
  p(\f|X)=\mathcal{N}(\f|\mathbf{0}, K),
\end{eqnarray}
where
$\f=\left[f_1^1,\ldots,f_n^1,f_1^2,\ldots,f_n^2,\ldots,f_1^c,\ldots,f_n^c\right]^T$
and $K$ is a $cn \times cn$ block-diagonal covariance matrix with
matrices $K^1, K^2,\ldots,K^c$ (each of size $n \times n$) on its
diagonal. This model is demonstrated in \code{demo\_multinomial} and
it has been used, for example, in
\citep{Juntunen+Vanhatalo+Peltonen+Mantyniemi:2012}.

\subsection{Cox proportional hazard model}

For the individual $i$, where $i=1,\ldots,n$, we have observed
survival time $y_i$ (possibly right censored) with censoring indicator
$\delta_i$, where $\delta_i=0$ if the $i$th observation is uncensored
and $\delta_i=1$ if the observation is right censored. The traditional
approach to analyze continuous time-to-event data is to assume the Cox
proportional hazard model \citep{Cox:1972}.
\begin{equation}
h_i(t)=h_0(t)\exp(\bm{x}^T_i\bm{\beta}),
\end{equation}
where $h_0$ is the unspecified baseline hazard rate, $\bm{x}_i$ is the
$d\times1$ vector of covariates for the $i$th patient and $\bm{\beta}$
is the vector of regression coefficients. The matrix
$X=[\bm{x}_1,\ldots,\bm{x}_n]^T$ of size $n\times d$ includes all
covariate observations.

The Cox model with linear predictor can be extended to more general
form to enable, for example, additive and non-linear effects of
covariates \citep{Kneib:2006,Martino:2011}. We extend the proportional
hazard model by
\begin{equation}
h_i(t)=\exp(\log(h_0(t))+\eta_i(\bm{x}_i)),
\end{equation}
where the linear predictor is replaced with the latent predictor
$\eta_i$ depending on the covariates $\bm{x}_i$. By assuming a
Gaussian process prior over
$\bm{\eta}=(\eta_1,\ldots,\eta_n)^T$, smooth nonlinear effects of
continuous covariates are possible, and if there are dependencies
between covariates, GP can model these interactions implicitly. 

A piecewise log-constant baseline hazard \citep[see,
e.g.][]{Ibrahim+Chen+Sinha:2001,Martino:2011} is assumed by
partitioning the time axis into $K$ intervals with equal lengths:
$0=s_0<s_1<s_2<\ldots<s_K$, where $s_K>y_i$ for all $i=1,\ldots,n$. In
the interval $k$ (where $k=1,\ldots,K$), hazard is assumed to be
constant:
\begin{eqnarray}
h_0(t)=\lambda_k&\mathrm{for}&t\in(s_{k-1},s_k].
\end{eqnarray}
For the $i$th individual the hazard rate in the $k$th time interval is
then
\begin{eqnarray}
h_i(t)=\exp(f_k+\eta_i(\bm{x}_i)), & t\in(s_{k-1},s_k],
\end{eqnarray}
where $f_k=\log(\lambda_k)$. To assume smooth hazard rate functions,
we place another Gaussian process prior for
$\bm{f}=(f_1,\ldots,f_K)^T$. We define a vector containing the mean
locations of $K$ time intervals as
$\bm{\tau}=(\tau_1,\ldots,\tau_K)^T$. 

The likelihood contribution for the possibly right censored $i$th
observation $(y_i,\delta_i)$ is assumed to be
\begin{equation}
l_i=h_i(y_i)^{(1-\delta_i)} \exp \left(
  -\int_0^{y_i}h_i(t)dt \right).
\end{equation}
Using the piecewise log-constant assumption for the hazard rate
function, the contribution of the observation $i$ for the likelihood
results in
\begin{equation}
l_i=[\lambda_k \exp(\eta_i)]^{(1-\delta_i)}\exp \left( -[(y_i-s_{k-1})\lambda_k
  + \sum_{g=1}^{k-1}(s_g-s_{g-1})\lambda_g ]\exp(\eta_i) \right),
\end{equation}
where $y_i\in(s_{k-1},s_k]$
\citep{Ibrahim+Chen+Sinha:2001,Martino:2011}. By applying the Bayes
theorem, the prior information and likelihood contributions are
combined, and the posterior distribution of the latent variables can
be computed. Due to the form of the likelihood function, the resulting
posterior becomes non-Gaussian and analytically exact inference is
intractable. \pkg{GPstuff} supports MCMC and Laplace approximation to
integrate over the latent variables. The use of Gaussian process Cox
proportional hazard model is demonstrated in
\code{demo\_survival\_coxph}. The Gaussian process Cox proportional
hazard model implemented with GPstuff was used in reference
\citep{Joensuu+etal:2012a} to model risk of gastrointestinal stromal
tumour recurrence after surgery.

\subsection{Zero-inflated negative binomial}
\label{sec_zinegbin}

Zero-inflated negative binomial model is suitable for modelling count
variables with excessive number of zero observations compared to usual
Poisson and negative binomial count models. Zero-inflated negative
binomial models assume a two-stage process, where the first process
determines whether the count is larger than zero, and the second
process determines the non-zero count \citep{Mullahy:1986}.
In \pkg{GPstuff}, these processes are assumed independent a priori, but they
become a posteriori dependent through the likelihood function
\begin{align}
  p + & (1-p)\mathrm{NegBin}(y|y=0),  \quad\mathrm{when}\quad y=0\\
      & (1-p)\mathrm{NegBin}(y|y>0),  \quad\mathrm{when}\quad y>0,
\end{align}
where the probability p is given by a binary classifier with logit
likelihood. NegBin is the Negative-binomial distribution parametrized
for the $i$'th observation as
\begin{equation}
 \y |\f,\mb{e}, r  \sim \prod_{i=1}^n
 \frac{\Gamma(r+y_i)}{y_i!\Gamma(r)}
\left(\frac{r}{r+\mu_i}\right)^r \left(\frac{\mu_i}{r+\mu_i}\right)^{y_i},
\end{equation}
where $\mu_i = e_i\exp(f(\x_i))$ and $r$ is the dispersion parameter
governing the variance. In \pkg{GPstuff}, the latent value vector
$\f=[\f_1^T \f_2^T]^T$ has length $2N$, where $N$ is the number of
observations. The latents $\f_1$ are associated with the
classification process and the latents $\f_2$ with the
negative-binomial count process. 

\subsection{Density estimation and regression}

Logistic Gaussian process can be used for flexible density estimation
and density regression. \pkg{GPstuff} includes implementation based on
Laplace (and MCMC) approximation as described in
\citep{Riihimaki+Vehtari:2012}.

The likelihood is defined as
\begin{align}
  p(\x|\f)=\prod_{i=1}^n \frac{\exp(f_i)}{\sum_{j=1}^n \exp(f_j)}.
\end{align}
For the latent function $f$, we assume
the model $f(\x) = g(\x)+\h(\x)^T \bm{\beta}$, where the GP prior
$g(\x)$ is combined with the explicit basis functions $\h(\x)$.
Regression coefficients are denoted with $\bm{\beta}$, and by placing
a Gaussian prior $\bm{\beta}\sim \mathcal{N}(\bb,B)$ with mean $\bb$
and covariance $B$, the parameters $\bm{\beta}$ can be integrated out
from the model, which results in the following GP prior for $f$:
\begin{equation}\label{gp_process}
  f(\x)\sim \mathcal{GP}\left(\h(\x)^T\bb,\kappa(\x,\x')+\h(\x)^TB\h(\x')\right).
\end{equation}
For the explicit basis functions, we use the second-order polynomials
which leads to a GP prior that can favour density estimates where the
tails of the distribution go eventually to zero.

We use finite-dimensional approximation and evaluate the integral and
the Gaussian process in a grid.
We approximate inference for logistic Gaussian process density
estimation in a grid using Laplace's method or MCMC to integrate over
the non-Gaussian posterior distribution of latent values \citep[see
details in][]{Riihimaki+Vehtari:2012}.

Logistic Gaussian processes are also suitable for estimating
conditional densities $p(t|\x)$, where $t$ is a response variable. We
discretize both input and target space in a finite region to model the
conditional densities with the logistic GP.
To approximate the resulting non-Gaussian posterior distribution, we
use again Laplace's method.

1D and 2D density estimation and density regression are implemented in
the function \code{lgpdens}. The usage of the function is demonstrated
in \code{demo\_lgpdens}.

\begin{figure}[]
  \begin{center}
    \subfigure[Galaxy data.]{
      \label{}
      \includegraphics[width=7cm]{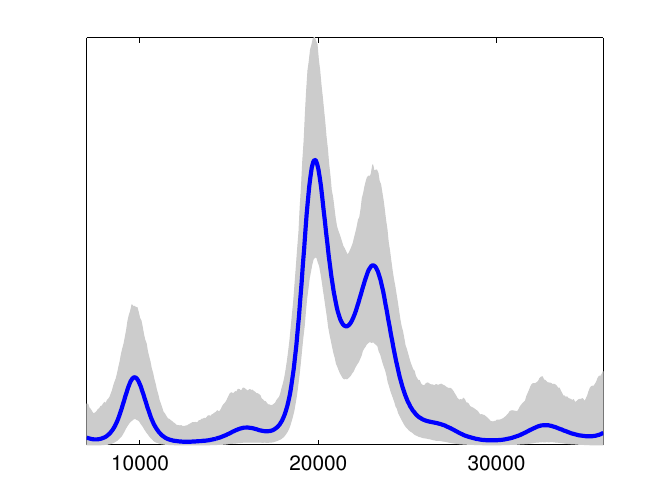}
    }
    ~
    \subfigure[Old faithful data.]{ 
      \label{}
      \includegraphics[width=7cm]{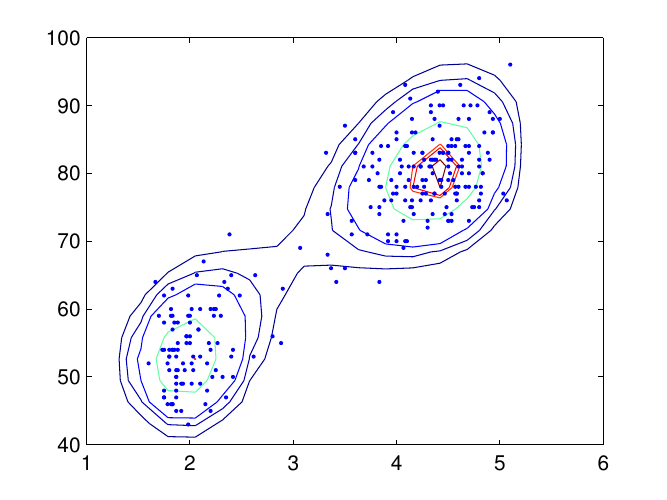}
    }
    \caption[]{Density estimates for two different distributions. Figures are from \code{demo\_lgpdens}. On the left blue line is density estimation of Galaxy data while gray area is 95\% mean region and on the right is 2D density estimation with countours based on Old faithful data.}\label{fig_demo_lgpdens}
  \end{center}
\end{figure}

\subsection{Input dependent models}

In input-dependent models more than one latent variables are used for modeling different parameters of the distribution. Additional latent variables in turn enable modeling dependencies between inputs and parameters. 

\subsubsection{Input-dependent Noise}
Input-dependent noise model \citep{Goldberg:1997} can be used to model
regression data. The model assumes Gaussian distributed data with
latent variables defining both mean and variance of the Gaussian
distribution instead of just the mean like in the standard Gaussian
likelihood. Setting independent GP priors for the two latent
variables, it is possible to infer non-constant noise,
e.g. input-dependent, in the data.

The likelihood is defined as
\begin{align}
  p(\y|\f^{(1)}, \f^{(2)}, \sigma^2)=\prod_{i=1}^n N(y_i | f_i^{(1)}, \sigma^2 \exp(f_i^{(2)})),
\end{align}
with latent function $f^{(1)}$ defining the mean and $f^{(2)}$
defining the variance. Input-dependent noise is demonstrated in
\code{demo\_inputdependentnoise} with both heteroscedastic and
constant noise.

\subsubsection{Input-dependent overdispersed Weibull}
In input-dependent Weibull model additional latent variable is used for modeling shape parameter $r$ of the standard Weibull model. 

The likelihood is defined as
\begin{eqnarray}
 p( \y |\f^{(1)}, \f_i^{(2)},\mathbf{z}) =&& \prod_{i=1}^n \exp(f_i^{(2)})^{1-z_i} \exp \Big{(} (1-z_i)f^{(1)}_i \\ \nonumber
 &+&(1-z_i)(\exp(f_i^{(2)})-1)\log(y_i)\exp(f^{(1)}_i)y_i^{\exp(f_i^{(2)})} \Big{)}, \nonumber
\end{eqnarray}
where $\mathbf{z}$ is a vector of censoring indicators with $z_i = 0$ for
uncensored event and $z_i = 1$ for right censored event for
observation $i$. Latent variable $f_2$ is used for modeling the shape parameter here. Because the dispersion parameter is constrained to be greater than 0, it must be computed through $\exp(f_2)$ to ensure positiveness. Input-dependent Weibull is demonstrated in \code{demo\_inputdependentweibull}.

\subsection{Monotonicity constraint}
\label{sec_monotonic}.

\citet{Riihimaki+Vehtari:2010} presented how a monotonicity constraint
for GP can be introduced with virtual derivative observations, and
the resulting posterior can be approximated with expectation
propagation.

Currently GPstuff supports monotonicity constraint for models which
have Gaussian likelihood or fully factoring likelihoods with EP
inference and which use covariance functions \code{gpcf\_constant},
\code{gpcf\_linear} and
\code{gpcf\_sexp}. \citet{Riihimaki+Vehtari:2010} used type II MAP
estimate for the hyperparameters, but GPstuff allows easy integration
over the hyperparameters with any method described in
Section~\ref{sec_marginalization_over_hyperparam}.

GP model with monotonicity constraints on one or many covariates can
be easily constructed with the function
\code{gp\_monotonic}. Monotonicity constraint is demonstrated in
\code{demo\_monotonic} and \code{demo\_monotonic2}.

\begin{figure}
  \begin{center}       
      \includegraphics[]{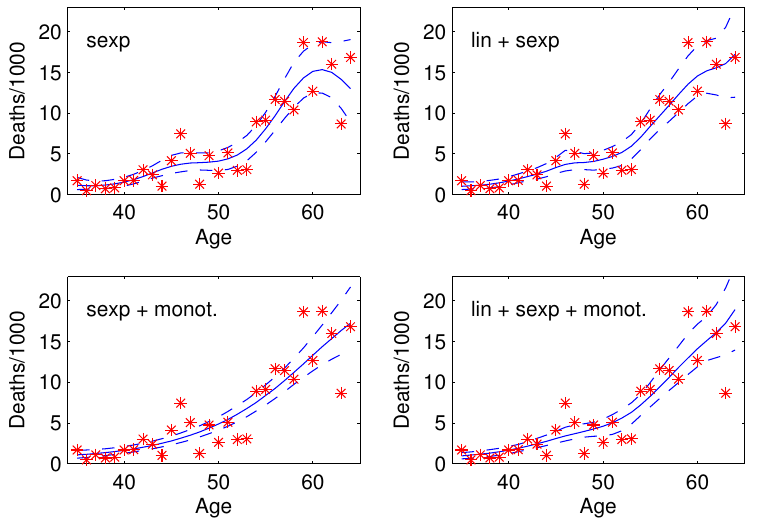}  
    \end{center}\caption{Illustration how the monotonicity constraint
      of the latent function affects the predictions. Mortality rate
      data from \citet{Broffitt:1988}.}
  \label{monotonicity} 
\end{figure}

Figure~\ref{monotonicity} displays the predictions with models with
and without the monotonicity constraint. The monotonicity constraint
allows inclusion of sensible prior information that death rate
increases as age increases. The effect is shown most clearly for
larger ages where there is less observations, but also overall reduction in
uncertainty related to the latent function can be seen.

\section{Mean functions}
\label{cha:mean-functions}

In the standard GP regression a zero mean function is assumed for
the prior process. This is convenient but there are nonetheless
some advantages in using a specified mean function. The basic
principle in doing GP regression with a mean function is to apply a
zero mean GP for the difference between observations and the mean
function.

\subsection{Explicit basis functions}

Here we follow closely the presentation of
\citep{Rasmussen+Williams:2006} about the subject and briefly
present the main results. A mean function can be specified as a
weighted sum of some basis functions $h$
\begin{equation}
  \textbf{m} = \textbf{h}(\textbf{x})^\top \boldsymbol{\beta},   \nonumber
\end{equation}
with weights $\boldsymbol{\beta}$. The target of modeling, the underlying process \textbf{g}, is assumed to be a sum of mean function and a zero mean GP. 
\begin{equation}
  \textbf{g} = \textbf{h}(\textbf{x})^\top \boldsymbol{\beta} + \mathcal{GP}(0,\textbf{K}).	\nonumber
\end{equation}
By assuming Gaussian prior for the weights $\boldsymbol{\beta} \sim \mathcal{N}(\textbf{b},\textbf{B}),\nonumber$ the weight parameters can be integrated out and the prior for \textbf{g} is another GP
\begin{equation}
  \label{GP_mean_prior}
  \emph{prior}\; \textbf{g} \sim \mathcal{GP}\left(\,\textbf{h}(\textbf{x})^\top \textbf{b},\,\textbf{K} + 
    \textbf{h}(\textbf{x})^\top \textbf{B}\,\textbf{h}(\textbf{x})\,\right). \nonumber
\end{equation}
The predictive equations are obtained by using the mean and covariance of the prior \eqref{GP_mean_prior} in the zero mean GP predictive equations \eqref{eq_posterior_in_gaussian_case}
\begin{eqnarray}
  \label{gp_mean_predE}
  \mathbb{E(\textbf{g}_*)} &=& \mathbb{E(\textbf{f}_*)} + \textbf{R}^\top \boldsymbol{\overline{\beta}},	\\
  \label{gp_mean_predV}\COV(\textbf{g}_*) &=& \COV(\textbf{f}_*) + \textbf{R}^\top\left( \textbf{B}^{-1} + 
    \textbf{H}\textbf{K}_{y}^{-1}\textbf{H}^\top \right)\textbf{R},	\\
  \nonumber\\
  &\boldsymbol{\overline{\beta}}& = 
  \left( \textbf{B}^{-1} + \textbf{H}\textbf{K}_y^{-1}\textbf{H}^\top \right)^{-1} 
  \left( \textbf{B}^{-1}\textbf{b} + \textbf{H}\textbf{K}_y^{-1}\textbf{y}\right), \nonumber\\
  &\textbf{R}& = \textbf{H}_* - \textbf{H}\textbf{K}_y^{-1}\textbf{K}_*,\nonumber\\
  &\textbf{H}& = 	\left[\begin{array}{c}
      h_1(\textbf{x})\\
      h_2(\textbf{x}) \\
      \vdots\\
      h_k(\textbf{x})
    \end{array}\right],\,  \textit{\textbf{x} is row vector}.\nonumber
\end{eqnarray}
If the prior assumptions about the weights are vague then $\textbf{B}^{-1} \rightarrow O$, ($O$ is a zero matrix) and the predictive 
equations \eqref{gp_mean_predE} and \eqref{gp_mean_predV} don't depend on $\textbf{b}$ or $\textbf{B}$
\begin{eqnarray}
  \label{gp_mean_predEvague}
  \mathbb{E(\textbf{g}_*)} &=& \mathbb{E(\textbf{f}_*)} + \textbf{R}^\top \boldsymbol{\overline{\beta}_v},\\
  \label{gp_mean_predVvague}\COV(\textbf{g}_*) &=& \COV(\textbf{f}_*) + \textbf{R}^\top\left( 
    \textbf{H}\textbf{K}_{y}^{-1}\textbf{H}^\top \right)\textbf{R},	\\
  \nonumber\\
  &\boldsymbol{\overline{\beta}_v}& = 
  \left(\textbf{H}\textbf{K}_y^{-1}\textbf{H}^\top \right)^{-1} 
  \textbf{H}\textbf{K}_y^{-1}\textbf{y}, \nonumber
\end{eqnarray}
Corresponding the exact and vague prior for the basis functions' weights there are two versions of the marginal likelihood.
With exact prior the marginal likelihood is
\begin{eqnarray}
  \log p(\textbf{y} \mid \textbf{X}, \textbf{b}, \textbf{B}) &=& -\frac{1}{2}\textbf{M}^\top \textbf{N}^{-1}\textbf{M} 
  -\frac{1}{2}\log|\textbf{K}_y|  -\frac{1}{2}\log|\textbf{B}|-\frac{1}{2}\log|\textbf{A}| -\frac{n}{2}\log 2\pi,\nonumber\\
  \textbf{M} &=& \textbf{H}^\top\textbf{b}-\textbf{y} ,\nonumber\\
  \textbf{N} &=& \textbf{K}_y  + \textbf{H}^\top \textbf{B}\textbf{H},\nonumber\\
  \textbf{A} &=& \textbf{B}^{-1} + \textbf{H}\textbf{K}_y^{-1}\textbf{H}^\top ,\nonumber
\end{eqnarray}
where $n$ is the amount of observations. Its derivative with respect to hyperparameters are
\begin{eqnarray}
  \frac{\partial }{\partial \theta_i} \log p(\textbf{y} \mid \textbf{X}, \textbf{b}, \textbf{B}) = &+&\frac{1}{2} 
  \textbf{M}^\top\textbf{N}^{-1}\frac{\partial \textbf{K}_y}{\partial \theta_i}\textbf{N}^{-1}\textbf{M}^\top \nonumber\\
  &-&\frac{1}{2}\text{tr}\left(\textbf{K}_y^{-1} \frac{\partial \textbf{K}_y}{\partial \theta_i}\right) - \frac{1}{2}\text{tr}\left(\textbf{A}^{-1} \frac{\partial \textbf{A}}{\partial \theta_i} \right), \nonumber\\
  \frac{\partial \textbf{A}}{\partial \theta_i} =&-&\textbf{H}\textbf{K}_y^{-1}\frac{\partial \textbf{K}_y}{\partial \theta_i}\textbf{K}_y^{-1}\textbf{H}^\top. 		\nonumber
\end{eqnarray}
With a vague prior the marginal likelihood is 
\begin{eqnarray}
  \log p(\textbf{y} \mid \textbf{X}) = &-&\frac{1}{2}\textbf{y}^\top \textbf{K}_y^{-1}\textbf{y} 
  + \frac{1}{2}\textbf{y}^\top \textbf{C}\textbf{y}			\nonumber\\
  &-&\frac{1}{2}\log |\textbf{K}_y|-\frac{1}{2}\log |\textbf{A}_v|
  -\frac{n-m}{2}\log 2\pi, 						\nonumber\\
  \textbf{A}_v = && \textbf{H}\textbf{K}_y^{-1}\textbf{H}^\top			\nonumber\\
  \textbf{C} =&& \textbf{K}_y^{-1}\textbf{H}^\top\textbf{A}_v^{-1}\textbf{H}\textbf{K}_y^{-1}, \nonumber
\end{eqnarray}
where $m$ is the rank of $H^\top$. Its derivative is
\begin{eqnarray}
  \label{gp_mean_marglikeli_pvag_der}
  \frac{\partial }{\partial \theta_i} \log p(\textbf{y} \mid \textbf{X})
  = &&\frac{1}{2}\textbf{y}^\top \textbf{P}\textbf{y}
  +\frac{1}{2}\left(-\textbf{y}^\top\textbf{P}\textbf{G}
    -\textbf{G}^\top\textbf{P}\textbf{y}
    +\textbf{G}^\top\textbf{P}\textbf{G}\right)\\
  &-&\frac{1}{2}\text{tr}\left(\textbf{K}_y^{-1} \frac{\partial
      \textbf{K}_y}{\partial \theta_i} \right) -
  \frac{1}{2}\text{tr}\left(\textbf{A}_v^{-1} \frac{\partial
      \textbf{A}_v}{\partial \theta_i} \right),        \nonumber\\
  \textbf{P} &=&    \textbf{K}_y^{-1} \frac{\partial \textbf{K}_y}{\partial
    \theta_i}\textbf{K}_y^{-1},\nonumber\\
  \textbf{G}
  &=&    \textbf{H}^\top\textbf{A}^{-1}\textbf{H}\textbf{K}_y^{-1}\textbf{y},    \nonumber
\end{eqnarray} 
where has been used the fact that matrices $\textbf{K}_y^{-1}$,
$\frac{\partial \textbf{K}_y}{\partial \theta_i}$ and $\textbf{A}_v$
are symmetric. The above expression
\eqref{gp_mean_marglikeli_pvag_der} could be simplified a little
further because $\textbf{y}^\top\textbf{P}\textbf{G}
+\textbf{G}^\top\textbf{P}\textbf{y} = 2\cdot
\textbf{y}^\top\textbf{P}\textbf{G}$. The use of mean functions is
demonstrated in \code{demo\_regression\_meanf}.

\section{Sparse Gaussian processes}
\label{cha:sparse-GP}

The evaluation of the inverse and determinant of the covariance matrix
in the log marginal likelihood (or its approximation) and its gradient
scale as $O(n^3)$ in time. This restricts the implementation of GP
models to moderate size data sets. However, the unfavorable scaling in
computational time can be alleviated with sparse approximations to GP
and compactly supported (CS) covariance functions. The sparse
approximations scale as $O(nm^2)$, where $m<n$ and the CS covariance
functions lead to computations which scale, in general, as $O(n^3)$
but with a smaller constant than with traditional globally supported
covariance functions. Many sparse GPs are proposed in the literature
and some of them are build into \pkg{GPstuff}.

\subsection{Compactly supported covariance functions}

A CS covariance function gives zero correlation between data points
whose distance exceeds a certain threshold. This leads to a sparse
covariance matrix. The challenge with constructing CS covariance
functions is to guarantee their positive definiteness and much
literature has been devoted on the subject \citep[see
e.g.][]{Sanso+Schuh:1987,Wu:1995,Wendland:1995,Gaspari+Cohn:1999,Gneiting:1999,Gneiting:2002,Buhmann:2000}
The CS functions implemented in \pkg{GPstuff} are Wendland's piecewise
polynomials $k_{\text{pp},q}$ \citep{Wendland:2005}, such as
\begin{equation}\label{ppcs2}
k_{\text{pp},2} = 
\frac{\sigma_{\text{pp}}^2}{3} (1 - r)_+^{j+2} \left( (j^2 + 4j
   +3) r^2  +   (3j + 6)r + 3\right) , 
\end{equation}
where $j = \lfloor d/2 \rfloor + 3$. These functions correspond to
processes that are $q$ times mean square differentiable and are
positive definite up to an input dimension $d$. Thus, the degree of
the polynomial has to be increased alongside the input dimension. The
dependence of CS covariance functions to the input dimension is very
fundamental. There are no radial CS functions that are positive
definite on every $\Re^d$ \citep[see e.g.][theorem
9.2]{Wendland:1995}.

The key idea with CS covariance functions is that, roughly speaking,
only the nonzero elements of the covariance matrix are used in the
calculations. This may speed up the calculations substantially since
in some situations only a fraction of the elements of the covariance
matrix are non-zero \citep[see
e.g.][]{Vanhatalo+Vehtari:2008,Rue+Martino+Chopin:2009}. In practice,
efficient sparse matrix routines are needed \citep{Davis:2006}. These
are nowadays a standard utility in many statistical computing packages
or available as an additional package for them.  \pkg{GPstuff}
utilizes the sparse matrix routines from \pkg{SuiteSparse} written by
Tim Davis (\url{http://www.cise.ufl.edu/research/sparse/SuiteSparse/})
and this package should be installed before using CS covariance
functions.

The CS covariance functions have been rather widely studied in the
geostatistics applications \citep[see
e.g.][]{Gneiting:2002,Furrer+Genton+Nychka:2006,Moreaux:2008}. There
the computational speed-up relies on efficient linear solvers for the
predictive mean $[\tilde{\f}] = \Kaf (\Kff +\sigma^2)^{-1}\y$. The
parameters are either fitted to the empirical covariance, optimized
using a line search in one dimension
\citep{Kaufman+Schervish+Nychka:2008} or sampled with Metropolis
Hastings. The benefits from a sparse covariance matrix have been
immediate since the problems collapse to solving sparse linear
systems. However, utilizing the gradient of the log posterior of the
parameters, as is done in \pkg{GPstuff} needs extra sparse matrix
tools. These are introduced and discussed by
\citet{Vanhatalo+Vehtari:2008}.  EP algorithm requires also special
considerations with CS covariance functions. The posterior covariance
in EP \eqref{eq_posterior_in_EP_case} does not remain sparse, and
thereby it has to be expressed implicitly. This issue is discussed by
\citet{Vanhatalo+Vehtari:2010} and
\citet{Vanhatalo+Pietilainen+Vehtari:2010}.

\subsubsection{Compactly supported covariance functions in GPstuff}

\begin{figure}[]
    \begin{center}
      \subfigure[The nonzero elements of $\Kff$.]{
        \label{}
        \includegraphics[width=3.2cm]{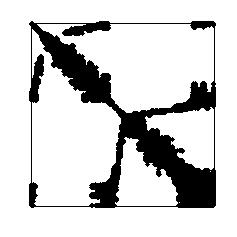}
      }
      \hspace{-8mm}
      ~
      \subfigure[The posterior predictive mean surface.]{ 
        \label{}
        \includegraphics[width=9cm]{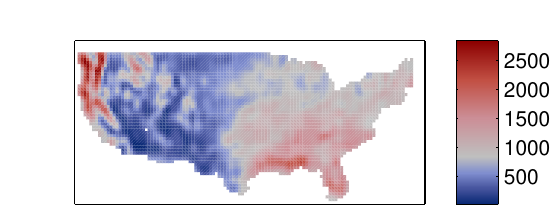}
       }
       \caption[]{The nonzero elements of $\Kff$ with
         $k_{\text{pp,2}}$ function, and the posterior predictive mean
         of the latent function in the US precipitation data
         set.}\label{demo_ppcsCov}
    \end{center}
\end{figure}

The demo \code{demo\_regression\_ppcs} contains a regression example
with CS covariance function $k_{\text{pp},2}$ (\code{gpcf\_ppcs2}).
The data contain US annual precipitation summaries from year 1995 for
5776 observation stations. The user interface of \pkg{GPstuff} makes
no difference between globally and compactly supported covariance
functions but the code is optimized to use sparse matrix routines
whenever the covariance matrix is sparse.  Thus, we can construct the
model, find the MAP estimate for the parameters and predict to new
input locations in a familiar way:
\begin{verbatim}
pn = prior_t('nu', 4, 's2', 0.3);
lik = lik_gaussian('sigma2', 1, 'sigma2_prior', pn);
pl2 = prior_gamma('sh', 5, 'is', 1);
pm2 = prior_sqrtt('nu', 1, 's2', 150);
gpcf2 = gpcf_ppcs2('nin', nin, 'lengthScale', [1 2], 'magnSigma2', 3, ...
                   'lengthScale_prior', pl2, 'magnSigma2_prior', pm2);
gp = gp_set('lik', lik, 'cf', gpcf2, 'jitterSigma2', 1e-6);
gp = gp_optim(gp,x,y,'opt',opt);
Eft = gp_pred(gp, x, y, xx);
\end{verbatim}
With this data the covariance matrix is rather sparse since only about
5\% of its elements are non-zero. The structure of the covariance
matrix is plotted after the approximate minimum degree (AMD)
permutation \citep{Davis:2006} in Figure~\ref{demo_ppcsCov}. The demo
\code{demo\_spatial2} illustrates the use of compactly supported
covariance functions with a non-Gaussian likelihood.

\subsection{FIC and PIC sparse approximations}\label{FIC_PICsparse_approximations}

\citet{Snelson+Ghahramani:2006} proposed a sparse pseudo-input GP
(SPGP), which \citet{Quinonero-Candela+Rasmussen:2005} named later
fully independent conditional (FIC).
The partially independent conditional (PIC) sparse approximation is an
extension of FIC
\citep{Quinonero-Candela+Rasmussen:2005,Snelson+Ghahramani:2007a}, and
they are both treated here following
\citet{Quinonero-Candela+Rasmussen:2005}. See also
\citep{Vanhatalo+Vehtari:2008,Vanhatalo+Pietilainen+Vehtari:2010} for
further discussion. The approximations are based on introducing an
additional set of latent variables $\uu = \{u_i \}_{i=1}^{m}$, called
\emph{inducing variables}. These correspond to a set of input
locations $\mb{X}_{u}$, called \emph{inducing inputs}. The latent
function prior is approximated as
\begin{equation}\label{cond_independ_joint_prior}   
  p(\f|\mb{X}, \theta) \approx q(\f | \mb{X}, \mb{X}_{u}, \theta) = \int q(\f |
  \mb{X}, \mb{X}_{u}, \uu, \theta)p(\uu | \mb{X}_{u}, \theta)\mathrm{d}\mathbf{u},
\end{equation}
where $q(\f|\mb{X}, \mb{X}_{u}, \uu, \theta)$ is an inducing
conditional. The above decomposition leads to the exact prior if the
true conditional $\f | \mb{X}, \mb{X}_{u}, \uu, \theta \sim \N(\Kfu \iKuu\uu,
\Kff-\mb{K}_{\text{f,u}} \iKuu \mb{K}_{\text{u,f}})$ is used. However,
in FIC framework the latent variables are assumed to be conditionally
independent given $\uu$, in which case the inducing conditional
factorizes $q(\f | \mb{X}, \mb{X}_{u}, \uu, \theta)\!= \prod q_i(f_i|\mb{X}, \mb{X}_{u}, \uu, \theta)$. In PIC latent
variables are set in blocks which are conditionally independent of
each others, given $\uu$, but the latent variables within a block have
a multivariate normal distribution with the original covariance. The
approximate conditionals of FIC and PIC can be summarized as
\begin{equation}
q(\f | \mb{X}, \mb{X}_{u}, \uu, \theta, \mb{M}) = \N(\f|\Kfu \iKuu\uu,
\mathrm{mask}\left(\Kff-\mb{K}_{\text{f,u}} \iKuu
\mb{K}_{\text{u,f}}|\mb{M}\right)), 
\end{equation}
where the function $\LL=\mathrm{mask}\left(\cdot|\mb{M}\right)$, with
matrix $\mb{M}$ of ones and zeros, returns a matrix $\LL$ of size
$\mb{M}$ and elements $\LL_{ij} = [\cdot]_{ij}$ if $\mb{M}_{ij} = 1$
and $\LL_{ij} =0$ otherwise.  An approximation with $\mb{M} = \mb{I}$
corresponds to FIC and an approximation where $\mb{M}$ is block
diagonal corresponds to PIC.  The inducing variables are given a
zero-mean Gaussian prior $\uu|\theta,\mb{X}_u\sim \N(\mb{0},\Kuu)$ so
that the approximate prior over latent variables is
\begin{equation}
  q(\f | \mb{X}, \mb{X}_{u},\theta,\mb{M}) = \N(\f | \mb{0}, \mb{K}_{\text{f,u}} \iKuu\mb{K}_{\text{u,f}}+\LL).
\label{GP_FITC_prior}
\end{equation}
The matrix $\mb{K}_{\text{f,u}} \iKuu\mb{K}_{\text{u,f}}$ is of
rank $m$ and $\LL$ is a rank $n$ (block) diagonal matrix. The prior
covariance above can be seen as a non-stationary covariance function of its
own 
where the inducing inputs $\mb{X}_u$ and the matrix $\mb{M}$ are free
parameters similar to parameters, which can be optimized
alongside $\theta$ \citep{Snelson+Ghahramani:2006,Lawrence:2007}.

The computational savings are obtained by using the
Woodbury-Sherman-Morrison lemma \citep[e.g.][]{Harville:1997} to
invert the covariance matrix in \eqref{GP_FITC_prior} as
\begin{equation}\label{Woodbury_eq}
(\mb{K}_{\text{f,u}} \iKuu\mb{K}_{\text{u,f}}+\LL)^{-1} = \LL^{-1} -
\mb{V}\mb{V}^{\text{T}},
\end{equation}
where $\mb{V} =\LL^{-1}\Kfu \text{chol}[(\Kuu
+\Kuf\LL^{-1}\Kfu)^{-1}]$. There is a
similar result also for the determinant. With FIC the computational
time is dominated by the matrix multiplications, which need time
$O(m^2n)$. With PIC the cost depends also on the sizes of the blocks
in $\LL$. If the blocks were of equal size $b\times b$, the time for
inversion of $\LL$ would be $O(n/b\times b^3)=O(nb^2)$. With blocks at
most the size of the number of inducing inputs, that is $b=m$, the
computational cost in PIC and FIC are similar. PIC approaches FIC in
the limit of a block size one and the exact GP in the limit of a block
size $n$ \citep{Snelson:2007b}.

\subsubsection{FIC sparse approximation in GPstuff}

The same data that were discussed in the section
\ref{sec:demo_regression1} is analyzed with sparse approximations in
the demo \code{demo\_regression\_sparse1}. The sparse approximation is
a property of the GP structure and we can construct the model
similarly to the full GP models:
\begin{verbatim}
gp_fic = gp_set('type', 'FIC', 'lik', lik, 'cf', gpcf, 'X_u', X_u)
\end{verbatim}
The difference is that we have to define the type of the sparse
approximation, here \verb|'FIC'|, and set the inducing inputs
\code{X\_u} in the GP structure. Since the inducing inputs are
considered as extra parameters common to all of the covariance
functions (there may be more than one covariance function in additive
models) they are set in the GP structure instead of the covariance
function structure. If we want to optimize the inducing inputs
alongside the parameters they need to have a prior as well.  GP
structure initialization gives them a uniform prior by default.

We can optimize all the parameters, including the inducing inputs, as
with a full GP. Sometimes, for example in spatial problems, it is
better to fix the inducing inputs
\citep{Vanhatalo+Pietilainen+Vehtari:2010} or it may be more efficient
to optimize the parameters and inducing inputs separately, so that we
iterate the separate optimization steps until convergence
(demonstrated in \code{demo\_regression\_sparse1}).  The parameters to
be optimized are defined by the field \code{infer\_params} in the GP
structure. This field regulates which parameters are considered fixed
and which are inferred in the group level (covariance function,
inducing inputs, likelihood).  We may also want to fix one of the
parameters inside these groups. For example, covariance function
magnitude.  If this is the case, then the parameter to be fixed should
be given an empty prior (\code{prior\_fixed}). If the parameter has a
prior structure it is an indicator that we want to infer that
parameter.
 
\subsubsection{PIC sparse approximation in GPstuff}\label{sec:PIC_regression}

In PIC, in addition to defining the inducing inputs, we need to appoint
every data point in a block. The block structure is common to all
covariance functions, similarly to the inducing inputs, for which reason
the block information is stored in the GP structure. After the
blocking and initialization of the inducing inputs the GP structure is
constructed as follows:
\begin{verbatim}
gp_pic = gp_set('type','PIC','lik',lik,'cf',gpcf,'X_u',X_u,'tr_index',trindex);
\end{verbatim}
Above, the cell array \code{trindex} contains the block index vectors
for training data. It means that, for example, the inputs and outputs
\code{x(trindex\{i\},:)} and \code{y(trindex\{i\},:)} belong to the
i'th block. The optimization of parameters and inducing inputs is
done the same way as with FIC or a full GP model. In prediction,
however, we have to give one extra input, \code{tstindex}, for
\code{gp\_pred}. This defines how the prediction inputs are appointed
in the blocks in a same manner as \code{trindex} appoints the training
inputs. 
\begin{verbatim}
Eft_pic = gp_pred(gp_pic, x, y, xt, 'tstind', tstindex);
\end{verbatim}
\subsection{Deterministic training conditional, subset of regressors and
  variational sparse approximation}

The deterministic training conditional is based on the works by
\citet{Csato+Opper:2002} and \citet{Seeger+Williams+Lawrence:2003} and
was earlier called Projected Latent Variables \citep[see][for
more details]{Quinonero-Candela+Rasmussen:2005}. The approximation 
can be constructed similarly as FIC and PIC by defining the inducing
conditional, which in the case of DTC is 
\begin{equation}
q(\f | \mb{X}, \mb{X}_{u}, \uu, \theta) = \N(\f|\Kfu \iKuu\uu,0).
\end{equation}
This implies that the approximate prior over latent variables is
\begin{equation}
  q(\f | \mb{X}, \mb{X}_{u}, \theta) = \N(\f | \mb{0}, \mb{K}_{\text{f,u}} \iKuu\mb{K}_{\text{u,f}}).
\label{GP_DTC_prior}
\end{equation}
The deterministic training conditional is not strictly speaking a
proper GP since it uses different covariance function for the latent
variables appointed to the training inputs and for the latent
variables at the prediction sites, $\tilde{f}$. The prior covariance
for $\tilde{f}$ is the true covariance $\Kaa$ instead of
$\Kau\iKuu\Kua$. This does not affect the predictive mean since the
cross covariance $\COV[\f,\tilde{\f}] = \Kfu\iKuu\Kua$, but it gives a
larger predictive variance. An older version of DTC is the subset of
regressors (SOR) sparse approximation which utilizes $\Kau\iKuu\Kua$.
However, this resembles a singular Gaussian distribution and thus the
predictive variance may be negative. DTC tries to fix this problem by
using $\Kaa$ \citep[see][]{Quinonero-Candela+Rasmussen:2005}. DTC and
SOR are identical in other respects than in the predictive variance
evaluation. In spatial statistics, SOR has been used by
\citet{Banerjee+Gelfand+Finley+Sang:2008} with a name Gaussian
predictive process model.

The approximate prior of the variational approximation by
\citet{Titsias:2009} is exactly the same as that of DTC. The
difference between the two approximations is that in the variational
setting the inducing inputs and covariance function parameters are
optimized differently. The inducing inputs and parameters can be seen
as variational parameters that should be chosen to maximize the
variational lower bound between the true GP posterior and the sparse
approximation. This leads to optimization of modified log marginal
likelihood
\begin{equation}
V(\theta, \mb{X}_u) = \log [N(\y|0, \sigma^2\mb{I} + \Qff) ] -
\frac{1}{2\sigma^2} \text{tr}(\Kff - \Kfu\iKuu\Kuf)
\end{equation}
with Gaussian likelihood. With non-Gaussian likelihood, the
variational lower bound is similar but $\sigma^2\mb{I}$ is replaced by
$\mb{W}^{-1}$ (Laplace approximation) or $\tilde{\mb{\Sigma}}$ (EP).

\subsubsection{Variational, DTC and SOR sparse approximation in GPstuff}


The variational, DTC and SOR sparse approximations are constructed
similarly to FIC. Only the type of the GP changes:
\begin{verbatim}
gp_var = gp_set('type', 'VAR', 'lik', lik, 'cf', gpcf, 'X_u', X_u);
gp_dtc = gp_set('type', 'DTC', 'lik', lik, 'cf', gpcf, 'X_u', X_u);
gp_sor = gp_set('type', 'SOR', 'lik', lik, 'cf', gpcf, 'X_u', X_u);
\end{verbatim}
The sparse GP approximations are compared in the demo
\code{demo\_regression\_sparse2}, and, for example,
\citet{Quinonero-Candela+Rasmussen:2005}, \citet{Snelson:2007b},
\citet{Titsias:2009}, and \citet{Alvarez+Luengo+Titsias+Lawrence:2010}
treat them in more detail.

\subsection{Sparse GP models with non-Gaussian likelihoods}

The extension of sparse GP models to non-Gaussian likelihoods is
straightforward in \pkg{GPstuff}. User can define the sparse GP just
as described in the previous two sections and then continue with the
construction of likelihood exactly the same way as with a full GP.
The Laplace approximation, EP and integration methods can be used with
the same commands as with full GP. This is demonstrated in
\code{demo\_spatial1}.

\subsection{Predictive active set selection for classification}

Predictive active set selection for Gaussian processes \citep[PASS-GP,
][]{Henao+Winther:2012} is a sparse approximation for classification
models. PASS-GP uses predictive distributions to estimate which data
points to include in the active set, so that it will include points
with potentially high impact to the classifier decision process while
removing those that are less relevant.
Function \code{passgp} accepts classification GP with latent method
set to EP or Laplace, and selects the active set and optimises the
hyperparameters. Its use is demonstrated in \code{demo\_passgp}.

\section{Modifying the covariance functions}\label{sec:modifying_covariance}

\subsection{Additive models}\label{sec_additive_models}

In many practical situations, a GP prior with only one covariance
function may be too restrictive since such a construction can model
effectively only one phenomenon. For example, the latent function may
vary rather smoothly across the whole area of interest, but at the
same time it can have fast local variations. In this case, a more
reasonable model would be $f(\mb{x}) = g(\mb{x}) + h(\mb{x})$, where
the latent value function is a sum of two functions, slow and fast
varying. By placing a separate GP prior for both of the functions $g$
and $h$ we obtain an additive prior
\begin{equation}\label{two_length_scale_model}
f(\x)|\theta \sim \GP(0,k_g(\x,\x') + k_h(\x,\x')).
\end{equation}
The marginal likelihood and posterior distribution of the latent
variables are as before with $\Kff = \mb{K}_{\text{g,g}} +
\mb{K}_{\text{h,h}}$. However, if we are interested on only, say,
phenomenon $g$, we can consider the $h$ part of the latent function as
correlated noise and evaluate the predictive distribution for $g$,
which with the Gaussian likelihood would be
\begin{equation}
  \tilde{g}(\tilde{\x})|\mathcal{D},
  \theta \sim \GP\left(k_g(\tilde{\x},\mb{X})(\Kff + \sigma^2\mb{I})^{-1}\mb{y}, 
    k_g(\tilde{\x},\tilde{\x}') -
    k_g(\tilde{\x},\mb{X})(\Kff + \sigma^2\mb{I})^{-1}  k_g(\mb{X},\tilde{\x}')\right).
\end{equation}
With non-Gaussian likelihood, the Laplace and EP approximations for
this are similar since only $\sigma^2\mb{I}$ and $(\Kff +
\sigma^2\mb{I})^{-1}\mb{y}$ change in the approximations.

The multiple length-scale model can be formed also using specific
covariance functions. For example, a rational quadratic covariance
function (\code{gpcf\_rq}) can be seen as a scale mixture of squared
exponential covariance functions \citep{Rasmussen+Williams:2006}, and
could be useful for data that contain both local and global phenomena.
However, using sparse approximations with the rational quadratic would
prevent it from modeling local phenomena. The additive model
\eqref{two_length_scale_model} suits better for sparse GP formalism
since it enables to combine FIC with CS covariance functions.

As discussed in section \ref{FIC_PICsparse_approximations}, FIC can be
interpreted as a realization of a special kind of covariance function.
\citet{Vanhatalo+Vehtari:2008} proposed to add FIC with CS covariance
function which leads to a latent variable prior
\begin{equation}\label{GP_CS+FIC_prior}
\f \mid \mb{X}, \mb{X}_{u}, \theta \sim \N(\mb{0},
\Kfu\iKuu\Kuf+\hat{\LL}),
\end{equation}
referred as CS+FIC. Here, the matrix
$\hat{\LL}=\LL+k_{\text{pp,q}}(\mb{X},\mb{X})$ is sparse with the same
sparsity structure as in $k_{\text{pp,q}}(\mb{X},\mb{X})$ and it is
fast to use in computations and cheap to store. 

\subsubsection{Additive models in GPstuff}

The additive models are demonstrated in \code{demo\_periodic} and
\code{demo\_regression\_additive1}. Their construction follows closely the
steps introduced in the previous sections. Consider that we want to
construct a GP with a covariance function that is a sum of squared
exponential and piecewise polynomial $k_{\text{se}}(\x,\x') +
k_{\text{pp},2}(\x,\x')$. In \pkg{GPstuff} this is done as follows:
\begin{verbatim}
gpcf1 = gpcf_sexp();
gpcf2 = gpcf_ppcs2('nin', 2);
lik = lik_gaussian('sigma2', 0.1);
gp = gp_set('lik', lik, 'cf', {gpcf1, gpcf2}); 
\end{verbatim}
The difference to previous models is that we give more than one
covariance function structure in cell array for the \code{gp\_set}.
The inference with additive model is conducted the same way as
earlier. If we want to make predictions for the two components
$k_{\text{se}}$ and $k_{\text{pp},2}$ independently we can do it by
giving an extra parameter-value pair for the \code{gp\_pred} as
\begin{verbatim}
[Ef_sexp, Varf_sexp] = gp_pred(gp, x, y, x, 'predcf', 1);
[Ef_ppcs2, Varf_ppcs2] = gp_pred(gp, x, y, x, 'predcf', 2);
\end{verbatim}

Additive models are constructed analogously with sparse approximations
by changing the type of the GP model. CS+FIC is a special kind of GP
structure and has its own type definition:
\begin{verbatim}
gp = gp_set('type', 'CS+FIC', 'lik', lik, 'cf', {gpcf1, gpcf2}, 'X_u', Xu);
\end{verbatim}

It is worth mentioning few things that should be noticed in relation
to sparse approximations. In general, FIC is able to model only
phenomena whose length-scale is long enough compared to the distance
between adjacent inducing inputs.  PIC on the other hand is able to
model also fast varying phenomena inside the blocks. Its drawback,
however, is that the correlation structure is discontinuous which may
result in discontinuous predictions. The CS+FIC model corrects these
deficiencies. In FIC and PIC the inducing inputs are parameters of
every covariance function, which means that all the correlations are
induced through the inducing inputs and the shortest length-scale the
GP is able to model is defined by the locations of the inducing
inputs. In CS+FIC, the CS covariance functions do not utilise inducing
inputs but evaluate the covariance exactly for which reason both the
long and short length-scale phenomena can be captured.  If there are
more than two covariance functions in CS+FIC all the globally
supported functions utilize inducing inputs and all the CS functions
are added to $\hat{\Lambda}$. A detailed treatment on the subject is
given in
\citep[][]{Vanhatalo+Vehtari:2008,Vanhatalo+Vehtari:2010,Vanhatalo+Pietilainen+Vehtari:2010}

\subsection{Additive covariance functions with selected variables}
\label{sec:addit-covar-funct}

In the demo (\code{demo\_regression\_additive2}), we demonstrate how
covariance functions can be modified so that they are functions of
only a subset of inputs. We model an artificial 2D regression data
with additive covariance functions that use only either the first or
second input variable. That is, the covariance is
$k_1(x_1,x_1'|\theta_1) + k_2(x_2,x_2'|\theta_2): \Re^2 \times
\Re^2\mapsto \Re$, where the covariance functions are of type
$k_1(x_1,x_1'|\theta_1): \Re \times \Re \mapsto \Re$. In the example
the covariance is a sum of a squared exponential, which is a function
of the first input dimension $x_1$, and a linear covariance function,
which is a function of the second input dimension $x_2$. The
covariance functions are constructed as follows:
\begin{verbatim}
gpcf_s1 = gpcf_sexp('selectedVariables', 1, 'lengthScale',0.5, ...
                    'magnSigma2', 0.15);
gpcf_l2 = gpcf_linear('selectedVariables', 2);
\end{verbatim}
The smaller set of inputs is chosen with the field input variable pair
\code{'selectedVariables',1}.

\subsection{Product of covariance functions}

A product of two or more covariance functions $k_1(\x,\x') \cdot
k_2(\x,\x')...$ is a valid covariance function as well. Such
constructions may be useful in situations where the phenomenon is
assumed to be separable. Combining covariance functions into product
form is done with a special covariance function \code{gpcf\_prod}. For
example, multiplying exponential and Mátern covariance functions to
produce separable spatio-temporal model $k(\x,\x') = k_1(x_1,x_1')
\cdot k_2([x_2,x_3]^{\text{T}},[x'_2,x'_3]^{\text{T}})$ where the
temporal component has covariance function $k_1$ and the spatial
components $k_2$ is done as follows:
\begin{verbatim}
gpcf1 = gpcf_exp('selectedVariables', 1);
gpcf2 = gpcf_matern32('selectedVariables', [2 3];
gpcf = gpcf_prod('cf', {gpcf1, gpcf2});
\end{verbatim}

The product covariance \code{gpcf\_prod} can also be used to combine
categorical covariance \code{gpcf\_cat} with other covariance
functions to build hierarchical linear and non-linear models, as
illustrated in \code{demo\_regression\_hier}.


\section{State space inference}\label{sec:kalman}
{\footnotesize By {Arno Solin}, {Jukka Koskenranta}, and {Simo S{\"a}rkk{\"a}}.\\}

\noindent
For one-dimensional Gaussian processes, instead of directly working with the
usual kernel formalism,$f \sim \mathcal{GP}(0,k(t,t'))$, of the Gaussian
process, certain classes of covariance functions allow to work with the
mathematical dual \citep{Sarkka+Solin+Hartikainen:2013}, where the Gaussian
process is though of as a solution to a $m$th order linear stochastic
differential equation (SDE). The corresponding inference problem can then solved
with Kalman filtering type of methods \citep{Grewal+Andrews:2001,Sarkka:2013},
where the computational complexity is $\mathcal{O}(m^3n)$ making this
formulation beneficial for long (often temporal) data sets.

The state space model corresponding to the GP regression problem can be given as
a linear time-invariant stochastic differential equation of the following form:
\begin{align} \label{eq:SDE}
  \begin{split}
  {\mathrm{d} \mathbf{f}(t) \over \mathrm{d} t} 
    &= \mathbf{F} \mathbf{f}(t) + \mathbf{L} \mathbf{w}(t) \\
  y_k 
    &= \mathbf{H} \mathbf{f}(t_k) + \varepsilon_k, 
    \quad \varepsilon_k \sim \mathrm{N}(0,\sigma_\mathrm{n}^2),
  \end{split}
\end{align}
where $\mathbf{f}(t) = \begin{pmatrix} f_1(t), f_2(t), \ldots, f_m(t)
\end{pmatrix}^\mathrm{T}$ holds the $m$ stochastic processes, and
$\mathbf{w}(t)$ is a multi-dimensional white noise process with spectral density
$\mathbf{Q}_\mathrm{c}$. The model is defined by the feedback matrix
$\mathbf{F}$ and the noise effect matrix $\mathbf{L}$.

The continuous-time linear time-invariant model \eqref{eq:SDE} can be solved for
discrete points. This is the closed-form solution to the SDE at the specified
time points, and it is given as
\begin{equation} \label{eq:state-space}
  \mathbf{f}_{k+1} = \mathbf{A}_k \mathbf{f}_k + \mathbf{q}_k, 
  \quad \mathbf{q}_k \sim \mathrm{N}(\mathbf{0}, \mathbf{Q}_k),
\end{equation}
where $\mathbf{f}(t_k) = \mathbf{f}_k$, and the state transition and process
noise covariance matrices can be solved analytically \citep[see,
e.g.,][]{Sarkka+Solin+Hartikainen:2013}. They are given as:
\begin{align}
  \mathbf{A}_k &= \mathbf{\Phi}(\Delta t_k), \\
  \mathbf{Q}_k &= \int_{0}^{\Delta t_k} 
    \mathbf{\Phi}(\Delta t_k-\tau)
    \mathbf{L} \mathbf{Q}_\mathrm{c} \mathbf{L}^\mathrm{T} 
    \mathbf{\Phi}(\Delta t_k-\tau)^\mathrm{T} \, \mathrm{d} \tau,
\end{align}
where $\Delta t_k = t_{k+1}-t_k$. The iteration is started form the stationary
state $\mathbf{f}_0 \sim \mathrm{N}(\mathbf{0},\mathbf{P}_\infty)$, where
$\mathbf{P}_\infty$ is a solution to the Lyapunov equation: $\mathbf{F}
\mathbf{P}_\infty + \mathbf{P}_\infty \mathbf{F}^\mathrm{T} + \mathbf{L}
\mathbf{Q}_\mathrm{c} \mathbf{L}^\mathrm{T} = \mathbf{0}$. For more details on
the connection between state space models and Gaussian processes, see, e.g.,
\citet{Sarkka+Solin+Hartikainen:2013}.

The inference problem is now directly solvable using Kalman filtering and
Rauch-Tung-Striebel smoothing methods \citep{Grewal+Andrews:2001, Sarkka:2013}.
The marginal likelihood and analytic gradients for conjugate gradient
optimization can be calculated in closed form. The sequential nature of the
state space solution is not well handled by Matlab/Octave, and thus this
formulation becomes appealing when the number of data points becomes very large
(in thousands).

In \pkg{GPstuff}, the following covariance functions are currently supported
(details on how the conversion is done are given in the accompanying
references):
\begin{description}[labelindent=\parindent]
  \item[Constant] 
    covariance function.
  \item[Noise] 
    covariance function.
  \item[Linear] 
    covariance function.
  \item[Exponential] 
    (Ornstein-Uhlenbeck) covariance function.
  \item[Mat\'ern] 
    covariance function (for $\nu=3/2$ and $\nu=5/2$).
  \item[Squared exponential] 
    covariance function \citep[as presented in][]{Hartikainen+Sarkka:2010}.
  \item[Periodic] 
    covariance function both with and without decay (quasi-periodicity) 
    \citep[as presented in][]{Solin+Sarkka:2014}.
  \item[Rational quadratic] 
    covariance function \citep[as presented in][]{Solin+Sarkka:2014-MLSP}.
  \item[Sums and products] 
    of covariance functions \citep[as given in][]{Solin+Sarkka:2014}.
\end{description}
These methods can be accessed by specifying the model type to 
`\textsc{kalman}', that is
\begin{verbatim}
gp = gp_set(gp,'type','KALMAN');
\end{verbatim}
Currently the `\textsc{kalman}' option is only supported for one-dimensional GP
regression, which can be performed by the usual workflow in \pkg{GPstuff}
after specifying the model type. For a more practical introduction, see the
demos `\code{demo\_kalman1}' (a simple one-dimensional regression problem) and
`\code{demo\_kalman2}' \citep[periodic modeling of the Mauna Loa $\text{CO}_2$
data, following][]{Solin+Sarkka:2014}.

\section{Model assessment and comparison}\label{sec:model_assessment}

There are various means to assess the goodness of the model and its
predictive performance. For an extensive review of the predictive
performance assessment methods see \citet{Vehtari+Ojanen:2012} and for
a shorter review comparing cross-validation, DIC and WAIC see
\citet{Gelman+Hwang+Vehtari:2014}. \pkg{GPstuff} provides four basic
approaches: marginal likelihood, cross-validation, deviance
information criterion (DIC) and widely applicable information
criterion (WAIC).

\subsection{Marginal likelihood}

Marginal likelihood is often used for model selection \citep[see,
e.g.][]{Kass+Raftery:1995}. It corresponds to ML II or with model
priors to MAP II estimate in the model space, selecting the model with
the highest marginal likelihood or highest marginal posterior
probability. 
In \pkg{GPstuff} the marginal posterior and marginal likelihood (or
its approximation in case of non-Gaussian likelihood) given the
parameters are computed by function \code{gp\_e}.

\subsection{Cross-validation}
\label{sec:cross-validation}

Cross-validation (CV) is an approach to estimate the predictive
performance for the future observations while avoiding the double use
of the data. The $i$th observation $(\x_{i},y_{i})$ in the training
data is left out, and then the predictive distribution for $y_i$ is
computed with a model that is fitted to all of the observations except
$(\x_i,y_i)$.
By repeating this for every point in the training data, we get a
collection of leave-one-out cross-validation (LOO-CV) predictive
densities.
For discussion of Bayesian cross-validation see \citet{Vehtari+Lampinen:2002,Vehtari+Ojanen:2012,Gelman+Hwang+Vehtari:2014} and for comparison of fast approximations for latent Gaussian variable models see \citet{Vehtari+etal:2014}.

For GP with given hyperparameters the LOO-CV predictions can be
computed in case of Gaussian likelihood using an analytical
solution \citep{Sundararajan+Keerthi:2001a} and in case of
non-Gaussian likelihood using EP approximation
\citep{Opper+Winther:2000,Rasmussen+Williams:2006} or Laplace
approximation using linear response approach \citep{Vehtari+etal:2014},
which have been implemented in \code{gp\_loopred}.

If set of hyperparameters have been obtained using \code{gp\_mc} or
\code{gp\_ia} LOO-posterior can be approximated using importance
sampling or weighting
\citep{Gelfand+Dey+Chang:1992,Vehtari+Lampinen:2002}, which is also
implemented in \code{gp\_loopred}. We use an integrated approach following \citet[][p. 40]{Vehtari:2001} where leave-one-out
predictive densities given hyperparameters
\begin{equation}
p(y_i|x_i,D_{-i},\vartheta)=\int
p(y_i|f_i,\vartheta)p(f_i|x_i,D_{-i},\vartheta)df_i 
\end{equation}
are obtained using the analytic solution, or the Laplace or EP
approximation as mentioned above \citep[see
also][]{Held+Schrodle+Rue:2010}. Thus the importance weighting is made
only for the hyperparameters. This reduces the variance of importance
sampling estimate significantly making it useful in practice. With
these approximations LOO-CV is very fast to compute in GPstuff. We
also use Pareto smoothed importance sampling
\citep{Vehtari+Gelman:2015,Vehtari+Gelman+Gabry:2015b} to further
reduce the variance of the estimate.

LOO-CV in GPstuff does not currently work for some multilatent
models. In such case or if there is reason to doubt the reliability of
the leave-one-out approximations, it is best to compute the non-approximated
cross-validation predictive densities.  To reduce computation time, in
$k$-fold-CV, $1/k$ part of the training data is left out, and then the
predictive distribution is computed with a model that is fitted to all
of the observations except that part.  Since the $k$-fold-CV
predictive densities are based on smaller training data sets than the
full data set, the estimate is slightly biased. In \pkg{GPstuff} first
order bias correction proposed by \citet{Burman:1989} is used.
\pkg{GPstuff} provides \code{gp\_kfcv}, which computes $k$-fold-CV and
bias-corrected $k$-fold-CV with log-score and root mean squared error
(RMSE). The function \code{gp\_kfcv} provides also basic variance
estimates for the predictive performance estimates.  The variance is
computed from the estimates for each fold ~\citep[see,
e.g.,][]{Dietterich:1998}.  See
\citet{Vehtari+Lampinen:2002,Vehtari+Ojanen:2012} for more details on
estimating the uncertainty in performance estimates.

\subsection{DIC}

Deviance information criterion (DIC) is another very popular model
selection criterion
\citep[][]{Spiegelhalter+Best+Carlin+Linde:2002}. DIC is not fully
Bayesian approach as it estimates the predictive performance if the
predictions were made using point-estimate (plug-in estimate) for
the unknowns, instead of using posterior predictive distribution.
Additionally DIC is defined only for regular models and thus fails
for singular models. DIC is included in \pkg{GPstuff} to allow
comparisons and better alternative WAIC is described in the next
section.

With parametric models without any hierarchy DIC is usually written as
\begin{align}
p_{\text{eff}} &= \E_{\theta|\mathcal{D}}[D(\y, \theta)] - D(\y,
\E_{\theta|\mathcal{D}}[\theta]) \\
\text{DIC} & = \E_{\theta|\mathcal{D}}[D(\y, \theta)] + p_{\text{eff}},
\end{align}
where $p_{\text{eff}}$ is the effective number of parameters and $D =
-2\log(p(\y|\theta))$ is the deviance. Since our models are
hierarchical we need to decide the parameters on focus \citep[see][for
discussion on this]{Spiegelhalter+Best+Carlin+Linde:2002}. The
parameters on the focus are those over which the expectations are
taken when evaluating the effective number of parameters and DIC. In
the above equations, the focus is in the parameters and in the
case of a hierarchical GP model of \pkg{GPstuff} the latent variables
would be integrated out before evaluating DIC. If we have a MAP
estimate for the parameters, we may be interested to evaluate DIC
statistics with the focus on the latent variables. In this case the
above formulation would be
\begin{align}
p_{D}(\theta) &= \E_{\f|\mathcal{D}, \theta}[D(\y, \f)] - D(\y,
\E_{\f|\mathcal{D}, \theta}[\f]) \label{DIC_focus_on_latent_variables1}\\
\text{DIC} & = \E_{\f|\mathcal{D},\theta}[D(\y, \f)] +
p_{D}(\theta).\label{DIC_focus_on_latent_variables2}
\end{align}
Here the effective number of parameters is denoted differently with
$p_{D}(\theta)$ since now we are approximating the effective number of
parameters in $\f$ conditionally on $\theta$, which is different from
the $p_{\text{eff}}$. $p_{D}(\theta)$ is a function of the
parameters and it measures to what extent the prior correlations
are preserved in the posterior of the latent variables given $\theta$.
For non-informative data $p_{D}(\theta)=0$ and the posterior is the
same as the prior. The greater $p_{D}(\theta)$ is the more the model
is fitted to the data and large values compared to $n$ indicate
potential overfit. Also, large $p_{D}(\theta)$ indicates that we
cannot assume that the conditional posterior approaches normality by
central limit theorem. Thus, $p_{D}(\theta)$ can be used for assessing
the goodness of the Laplace or EP approximation for the conditional
posterior of the latent variables as discussed by
\citet{Rue+Martino+Chopin:2009} and
\citet{Vanhatalo+Pietilainen+Vehtari:2010}. The third option is to
evaluate DIC with focus on all the variables, $[\f, \theta]$. In this
case the expectations are over $p(\f,\theta|\mathcal{D})$.

\subsection{WAIC}

\citet{Watanabe:2009,Watanabe:2010a,Watanabe:2010d} presented
widely applicable information criterion (WAIC) and gave a formal
proof of its properties as an estimate for the predictive
performance of posterior predictive distributions for both regular
and singular models. A criterion of similar form was independently
proposed by \citet{Richardson:2002} as a version of DIC, but
without formal justification.

Other information criteria are based on Fisher's asymptotic theory
assuming a regular model for which the likelihood or the posterior
converges to a single point and MLE, MAP, and plug-in estimates are
asymptotically equivalent. With singular models the set of true
parameters consists of more than one point, the Fisher information 
matrix is not positive definite, plug-in estimates are not
representative of the posterior and the distribution of the deviance
does not converge to a $\chi^2_\nu$ distribution. 

Watanabe shows that the Bayesian generalization utility can
be estimated by a criterion 
\begin{align}
  \WAIC_G = BU_{t} - 2(BU_{t} - GU_{t}),
\end{align}
where $BU_t$ is Bayes training utility
\begin{align}
  BU_t = \frac{1}{n} \sum_{i=1}^n \log p(y_i|D,M_k) 
\end{align}
and $GU_t$ is Gibbs training utility
\begin{align}
  GU_t = \frac{1}{n} \sum_{i=1}^n \int \log
  p(y_i|\theta,M_k)  p(\theta|D,M_k) d\theta
\end{align}
WAIC can also be given as a functional variance form 
\begin{align}
  \WAIC_V &= BU_{t} - V/n,
\end{align}
where the functional variance
\begin{align}
  V = 
  \sum_{i=1}^n \Bigg\{ 
    & \E_{\theta|D,M_k} \left[ \left( 
        \log p({y}_i|{x}_i,\theta,M_k)
      \right)^2 \right] \nonumber \\
    & \quad 
    - \left( 
      \E_{\theta|D,M_k} \left[
        \log p({y}_i|{x}_i,\theta,M_k)
      \right]
    \right)^2
  \Bigg\},
\end{align}
describes the fluctuation of the posterior distribution.

WAIC is asymptotically equal to the true logarithmic utility in
both regular and singular statistical models and the error in a
finite case is $o(1/n)$. \citet{Watanabe:2010d} shows also that the
WAIC estimate is asymptotically equal to the Bayesian
cross-validation estimate (section \ref{sec:cross-validation}).
$\WAIC_{G}$ and $\WAIC_V$ are asymptotically equal, but the series
expansion of $WAIC_V$ has closer resemblance to the series
expansion of the logarithmic leave-one-out utility.

\subsection{Model assessment demos}

The model assessment methods are demonstrated with the functions
\code{demo\_modelassesment1} and \code{demo\_modelassesment2}. The
former compares the sparse GP approximations to the full GP with
regression data and the latter compares the logit and probit
likelihoods in GP classification.

Assume that we have built our regression model with a Gaussian noise
and used optimization method to find the MAP estimate for the
parameters. We evaluate the effective number of latent
variables, DIC and WAIC
\begin{verbatim}
p_eff_latent = gp_peff(gp, x, y);
[DIC_latent, p_eff_latent2] = gp_dic(gp, x, y, 'focus', 'latent');
WAIC = gp_waic(gp,x,y);
\end{verbatim}
where $p_{\text{eff}}$ is evaluated with two different approximations.
Since we have the MAP estimate for the parameters the focus is on
the latent variables. In this case we can also use \code{gp\_peff}
which returns the effective number of parameters approximated as
\begin{equation}\label{eq_peff_fast_approx}
p_{D}(\theta) \approx n - \text{tr}(\iKff (\iKff + \sigma^{-2}\mb{I})^{-1})
\end{equation}
\citep{Spiegelhalter+Best+Carlin+Linde:2002}. When the focus is on the
latent variables, the function \code{gp\_dic} evaluates the DIC
statistics and the effective number of parameters as described by the
equations \eqref{DIC_focus_on_latent_variables1} and
\eqref{DIC_focus_on_latent_variables2}. 

The $k$-fold-CV expected utility estimate can be evaluated as follows:
\begin{verbatim}
cvres =  gp_kfcv(gp, x, y);
\end{verbatim}
The \code{gp\_kfcv} takes the ready made model structure \code{gp} and
the training data \code{x} and \code{y}. The function divides the data
into $k$ groups, conducts inference separately for each of the
training groups and evaluates the expected utilities with the test
groups. Since no optional parameters are given the inference is
conducted using MAP estimate for the parameters. The default division
of the data is into 10 groups.  The expected utilities and their
variance estimates are stored in the structure \code{cvres}.
\code{gp\_kfcv} returns also other statistics if more information is
needed and the function can be used to save the results automatically.

Assume now that we have a record structure from \code{gp\_mc}
function with Markov chain samples of the parameters stored in it.
In this case, we have two options how to evaluate the DIC
statistics. We can set the focus on the parameters or all the
parameters (that is parameters and latent variables). The two
versions of DIC and the effective number of parameters and WAIC are
evaluated as follows:
\begin{verbatim}
rgp = gp_mc(gp, x, y, opt);
[DIC, p_eff] =  gp_dic(rgp, x, y, 'focus', 'param');
[DIC2, p_eff2] =  gp_dic(rgp, x, y, 'focus', 'all');
WAIC = gp_waic(gp,x,y);
\end{verbatim}
Here the first line performs the MCMC sampling with options
\code{opt}. The next two lines evaluate the DIC statistics and the
last line evaluates WAIC. With Markov chain samples, we cannot use
the \code{gp\_peff} function to evaluate $p_{D}(\theta)$ since that
is a special function for models with fixed parameters.

The functions \code{gp\_peff}, \code{gp\_dic} and \code{gp\_kfcv} work
similarly for non-Gaussian likelihoods as for a Gaussian one. The only
difference is that the integration over the latent variables is done
approximately.  The way the latent variables are treated is defined in
the field \code{latent\_method} of the GP structure and this is
initialized when constructing the model as discussed in the section
\ref{GP_classification}. The effective number of parameters returned
by \code{gp\_peff} is evaluated as in the equation
\eqref{eq_peff_fast_approx} with the modification that
$\sigma^{-2}\mb{I}$ is replaced by $\mb{W}$ in the case of Laplace
approximation and $\tilde{\mb{\Sigma}}^{-1}$ in the case of EP.

\section{Adding new features in the toolbox}\label{sec_adding_new_features}

As described in the previous sections, \pkg{GPstuff} is build
modularly so that the full model is constructed by combining separate
building blocks. Each building block, such as the covariance function
or likelihood, is written in a m-file which contains
all the functions and parameters specific to it. This construction
makes it easier to add new model blocks in the package. For example, new
covariance functions can be written by copying one of the existing
functions to a new file and modifying all the subfunctions and
parameters according to the new covariance function. With likelihoods
it should be remembered that \pkg{GPstuff} assumes currently that they
factorize as described in the equation~\eqref{likelihood}. All the
inference methods should work for log-concave likelihoods. However,
likelihood functions that are not log-concave may lead to problems
with Laplace approximation and EP since the conditional posterior of
the latent variables may be multimodal and the diagonal matrices
$\mb{W}$ and $\tilde{\mb{\Sigma}}$ may contain negative elements
\citep{Vanhatalo+Jylanki+Vehtari:2009}. For example, Student-$t$
observation model leads to a likelihood, which is not log-concave, and
requires some specific modifications to the Laplace approximation and
EP algorithm \citep[see][]{Jylanki+Vanhatalo+Vehtari:2011}. Using MCMC
should, however, be straightforward with non-log-concave likelihoods
as well. 
See Appendix~\ref{developer_appendix} for technical details of
\code{lik}, \code{gpcf} and \code{prior} functions, and argument
interface for optimisation and sampling functions.

The package is not as modular with respect to the computational
techniques as to the model blocks. Usually, each inference method
requires specific summaries from the model blocks. For example, the EP
algorithm requires that each likelihood has a function to evaluate
integrals such as $\int f_i p(y_i|f_i,\phi) N(f_i|\mu,\sigma^2) df_i$
whereas the Laplace approximation requires the Hessian of the log
likelihood, $\nabla^2_{\f} \log p(\y|\f,\phi)$.  For this reason
adding, for example, variational approximation would require
modifications also in the likelihood functions.

\section{Discussion}

Broadly thinking, GPs have been researched intensively for decades and
many disciplines have contributed to their study. They have not,
necessarily, been called GP but, for example, spatial statistics,
signal processing and filtering literature have their own naming
conventions. The point of view taken in \pkg{GPstuff} comes mainly
from the machine learning literature where the subject has flourished
to a hot topic since late 1990's. Our aim in the \pkg{GPstuff} package
has been to collect existing and create new practical tools for
analyzing GP models. We acknowledge that many of the basic models and
algorithms have been proposed by others. However, the implementation
in \pkg{GPstuff} is unique and contains several practical solutions
for computational problems that are not published elsewhere. Our own
contribution on GP theory that is included in \pkg{GPstuff} is
described mainly in
\citep{Vehtari:2001,Vehtari+Lampinen:2002,Vanhatalo+Vehtari:2007,Vanhatalo+Vehtari:2008,Vanhatalo+Vehtari:2010,Vanhatalo+Jylanki+Vehtari:2009,Vanhatalo+Pietilainen+Vehtari:2010,Jylanki+Vanhatalo+Vehtari:2011,Riihimaki+Vehtari:2012,Joensuu+etal:2012a,Riihimaki+Jylanki+Vehtari:2013,Joensuu+Reichardt+Eriksson+Hall+Vehtari:2014,Riihimaki+Vehtari:2014}

Our intention has been to create a toolbox with which useful,
interpretable results can be produced in a sensible time.
\pkg{GPstuff} provides approximations with varying level of accuracy.
The approximation to be used needs to be chosen so that it
approximates well enough the essential aspects in the model. Thus, the
choice is always data, model, and problem dependent. For example, MCMC
methods are often praised for their asymptotic properties and
seemingly easy implementation.  Algorithms, such as
Metropolis-Hastings, are easy to implement for most models. The
problem, however, is that as a data set grows they may not give
reliable results in a finite time. With GP models this problem is
faced severely. For example, currently problems with over 10~000 data
points would be impractical to analyse with a standard office PC using
MCMC since the convergence rate and mixing of the sample chain would
be too slow. For this reason it is important to have also faster, but
perhaps less accurate, methods.  The choice of the method is then a
compromise between time and accuracy. The inference is the fastest
when using MAP estimate for the parameters and a Gaussian function for
the conditional posterior.  With a Gaussian likelihood, the Gaussian
conditional distribution is exact and the only source of imprecision
is the point estimate for the parameters. If the likelihood is other
than Gaussian, the conditional distribution is an approximation, whose
quality depends on how close to Gaussian the real conditional
posterior is, and how well the mean and covariance are approximated.
The form of the real posterior depends on many things for which reason
the Gaussian approximation has to be assessed independently for every
data.

One of our aims with \pkg{GPstuff} has been to provide an easy way to
detect model misspecifications. This requires the model assessment
tools discussed in the section \ref{sec:model_assessment} and an easy
way to test alternative model structures. The model misfit most often
relates either to the likelihood or GP prior. For example, if the data
contained outliers or observations with clearly higher variance than
what the Gaussian or Poisson observation model predicts, the posterior
of the latent function would be highly compromised. For this reason,
robust alternatives for traditional models, such as Student-$t$ or
negative binomial distribution should be easily testable. Even though
the GP prior is very flexible and only the mean and covariance need to
be fixed, it still contains rather heavy assumptions. For example, GP
associated with the squared exponential covariance function is
indefinitely mean square differentiable. This is a very strong
assumption on the smoothness of the latent function.  In fact it is
rather peculiar how little attention other covariance functions have
gained in machine learning literature. One of the reasons may be that
often machine learning problems take place in high dimensional input
spaces where data are enforced to lie sparsely and for which reason
the possible solutions are smooth. However, the covariance function
sometimes does influence the results \citep[see
e.g.][]{Vanhatalo+Vehtari:2008,Vanhatalo+Pietilainen+Vehtari:2010}.

In statistics, inference in the parameters is a natural concern but in
machine learning literature they are left in less attention. An
indicator of this is the usual approach to maximize the marginal
likelihood which implies uniform prior for the parameters.
\pkg{GPstuff} provides an easy way to define priors explicitly so that
people would really use them \citep[this principle is also in line
with reasoning by][]{Gelman:2006}. We want to stress a few reasons for
explicitly defining a prior. In spatial statistics, it is well known
that the length-scale and magnitude are underidentifiable and the
proportion $\sigma^2/l$ is more important to the predictive
performance than their individual values
\citep{Diggle+Tawn+Moyeed:1998,Zhang:2004,Diggle+Ribeiro:2007}. These
results are shown for Mat{\'e}rn class of covariance functions but
according to our experiments they seem to apply for Wendland's
piecewise polynomials as well. With them, the property can be taken
advantage of since by giving more weight to short length-scales one
favors sparser covariance matrices that are faster in computations.
Other advantage is that priors make the inference problem easier by
narrowing the posterior and making the parameters more identifiable.
This is useful especially for MCMC methods but optimization and other
integration approximations gain from the priors as well. These two
reasons are rather practical. More fundamental reason is that in
Bayesian statistics leaving prior undefined (meaning uniform prior) is
a prior statement as well, and sometimes it may be really awkward. For
example, uniform prior works very badly for the parameters of the
Student-$t$ distribution. Thus, it is better to spend some time
thinking what the prior actually says than blindly use the uniform.

There are also other analytic approximations than the
Laplace approximation or EP proposed in the literature. Most of these
are based on some kind of variational approximation
\citep{Gibbs+MacKay:2000,Csato+Opper:2002,Tipping+Lawrence:2005,Kuss:2006,Opper+Archambeau:2009}.
Laplace approximation and EP were chosen to \pkg{GPstuff} for a few
reasons. They both are, in theory, straightforward to implement for
any factorizable likelihood (although sometimes there are practical
problems with the implementation). Laplace approximation is also fast
and EP is shown to work well in many problems. Here, we have
considered parametric observation models but non-parametric versions
are possible as well. Examples of possible extensions to that
direction are provided by \citet{Snelson+Rasmussen+Ghahramani:2004}
and \citet{Tokdar+Zhu+Ghosh:2010}.

\pkg{GPstuff} is an ongoing project and the package will be updated
whenever new functionalities are ready. The latest version of the
toolbox is available from
\url{http://becs.aalto.fi/en/research/bayes/gpstuff/}. 

\section*{Acknowledgments}

The research leading to \pkg{GPstuff} has been funded by the Academy
of Finland (grant 218248), Finnish Funding Agency for Technology and
Innovation (TEKES), and the Graduate School in Electronics and
Telecommunications and Automation (GETA). The first author also thanks
the Finnish Foundation for Economic and Technology Sciences KAUTE,
Finnish Cultural Foundation, and Emil Aaltonen Foundation for
supporting his post graduate studies during which the code package was
prepared.

The package contains pieces of code written by other people than the
authors. In the Department of Biomedical Engineering and Computational
Science at Aalto University these persons are (in alphabetical order):
Toni Auranen, Pasi Jyl{\"a}nki, Jukka Koskenranta, Enrique Lelo de
Larrea Andrade, Tuomas Nikoskinen, Tomi Peltola, Eero Pennala, Heikki
Peura, Ville Pietil{\"a}inen, Markus Siivola, Arno Solin, Simo
S{\"a}rkk{\"a} and Ernesto Ulloa. People outside Aalto University are
(in alphabetical order): Christopher M. Bishop, Timothy A. Davis,
Matthew D. Hoffman, Kurt Hornik, Dirk-Jan Kroon, Iain Murray, Ian T.
Nabney, Radford M. Neal and Carl E. Rasmussen. We want to thank them all
for sharing their code under a free software license.

\bibliography{GPstuff-nodoi}

\begin{thebibliography}{145}
\providecommand{\natexlab}[1]{#1}
\providecommand{\url}[1]{\texttt{#1}}
\expandafter\ifx\csname urlstyle\endcsname\relax
  \providecommand{\doi}[1]{doi: #1}\else
  \providecommand{\doi}{doi: \begingroup \urlstyle{rm}\Url}\fi

\bibitem[Abramowitz and Stegun(1970)]{Abramowitz+Stegun:1970}
Milton Abramowitz and Irene~A. Stegun.
\newblock \emph{Handbook of mathematical functions}.
\newblock Dover Publications, Inc., 1970.

\bibitem[Ahmad et~al.(2000)Ahmad, Boschi-Pinto, Lopez, Murray, Lozano, and
  Inoue]{Ahmad+all:2000}
Omar~B. Ahmad, Cynthia Boschi-Pinto, Alan~D. Lopez, Christopher~J.L. Murray,
  Rafael Lozano, and Mie Inoue.
\newblock Age standardization of rates: A new {W}{H}{O} standard.
\newblock \emph{GPE Discussion Paper Series}, 31, 2000.

\bibitem[Alvarez et~al.(2010)Alvarez, Luengo, Titsias, and
  Lawrence]{Alvarez+Luengo+Titsias+Lawrence:2010}
M.~A. Alvarez, D.~Luengo, M.~K. Titsias, and N.~D. Lawrence.
\newblock Efficient multioutput {Gaussian} processes through variational
  inducing kernels.
\newblock \emph{JMLR Workshop and conference proceedings}, 9:\penalty0 25--32,
  2010.

\bibitem[Banerjee et~al.(2008)Banerjee, Gelfand, Finley, and
  Sang]{Banerjee+Gelfand+Finley+Sang:2008}
Sudipto Banerjee, Alan~E. Gelfand, Andrew~O. Finley, and Huiyan Sang.
\newblock Gaussian predictive process models for large spatial data sets.
\newblock \emph{Journal of the Royal Statistical Society B}, 70\penalty0
  (4):\penalty0 825--848, 2008.

\bibitem[Berrocal et~al.(2010)Berrocal, Raftery, Gneiting, and
  Steed]{Berrocal+Raftery+Gneiting+Steed:2009}
Veronica~J. Berrocal, Adrian~E. Raftery, Tilmann Gneiting, and Richard~C.
  Steed.
\newblock Probabilistic weather forecasting for winter road maintenance.
\newblock \emph{Journal of the American Statistical Association}, 105\penalty0
  (490):\penalty0 522--537, 2010.

\bibitem[Best et~al.(2005)Best, Richardson, and
  Thomson]{Best+Richardson+Thomson:2005}
Nicky Best, Sylvia Richardson, and Andrew Thomson.
\newblock A comparison of {Bayesian} spatial models for disease mapping.
\newblock \emph{Statistical Methods in Medical Research}, 14:\penalty0 35--59,
  2005.

\bibitem[Boukouvalas et~al.(2012)Boukouvalas, Barillec, and
  Cornford]{Boukouvalas:2012}
Alexis Boukouvalas, Remi Barillec, and Dan Cornford.
\newblock Gaussian process quantile regression using expectation propagation.
\newblock \emph{arXiv preprint arXiv:1206.6391}, 2012.

\bibitem[Broffitt(1988)]{Broffitt:1988}
J.~D. Broffitt.
\newblock Increasing and increasing convex bayesian graduation.
\newblock \emph{Transactions of the Society of Actuaries}, 40\penalty0
  (1):\penalty0 115--148, 1988.

\bibitem[Buhmann(2001)]{Buhmann:2000}
M.~D. Buhmann.
\newblock A new class of radial basis functions with compact support.
\newblock \emph{Mathematics of Computation}, 70\penalty0 (233):\penalty0
  307--318, January 2001.

\bibitem[Burman(1989)]{Burman:1989}
Prabir Burman.
\newblock A comparative study of ordinary cross-validation, $v$-fold
  cross-validation and the repeated learning-testing methods.
\newblock \emph{Biometrika}, 76\penalty0 (3):\penalty0 503--514, 1989.

\bibitem[Christensen et~al.(2006)Christensen, Roberts, and
  Sk\"old]{Christensen+Roberts+Skold:2006}
Ole~F. Christensen, Gareth~O. Roberts, and Martin Sk\"old.
\newblock Robust {Markov} chain {Monte} {Carlo} methods for spatial generalized
  linear mixed models.
\newblock \emph{Journal of Computational and Graphical Statistics},
  15:\penalty0 1--17, 2006.

\bibitem[Conti et~al.(2009)Conti, Gosling, Oakley, and
  O'Hagan]{Conti+Gosling+Oakley+OHagan:2009}
Stefano Conti, John~Paul Gosling, Jeremy Oakley, and Anthony O'Hagan.
\newblock Gaussian process emulation of dynamic computer codes.
\newblock \emph{Biometrika}, 96:\penalty0 663--676, 2009.

\bibitem[Cox(1972)]{Cox:1972}
David~R Cox.
\newblock Regression models and life-tables.
\newblock \emph{Journal of the Royal Statistical Society. Series B
  (Methodological)}, 34\penalty0 (2):\penalty0 187--220, 1972.

\bibitem[Cressie(1993)]{Cressie:1993}
Noel A.~C. Cressie.
\newblock \emph{Statistics for Spatial Data}.
\newblock John Wiley \& Sons, Inc., 1993.

\bibitem[Csat\'o and Opper(2002)]{Csato+Opper:2002}
Lehel Csat\'o and Manfred Opper.
\newblock Sparse online {Gaussian} processes.
\newblock \emph{Neural Computation}, 14\penalty0 (3):\penalty0 641--669, 2002.

\bibitem[Cseke and Heskes(2011)]{Cseke+Heskes:2011}
Botond Cseke and Tom Heskes.
\newblock Approximate marginals in latent {Gaussian} models.
\newblock \emph{Journal of Machine Learning Research}, 12:\penalty0 417--454, 2
  2011.
\newblock ISSN 1532-4435.

\bibitem[Davis(2005)]{Davis:2005}
Timothy~A. Davis.
\newblock Algorithm 849: A concise sparse {Cholesky} factorization package.
\newblock \emph{ACM Trans. Math. Softw.}, 31:\penalty0 587--591, December 2005.

\bibitem[Davis(2006)]{Davis:2006}
Timothy~A. Davis.
\newblock \emph{Direct Methods for Sparse Linear Systems}.
\newblock SIAM, 2006.

\bibitem[Deisenroth et~al.(2009)Deisenroth, Rasmussen, and
  Peters]{Deisenroth+Rasmussen+Peters:2009}
Marc~Peter Deisenroth, Carl~Edward Rasmussen, and Jan Peters.
\newblock Gaussian process dynamic programming.
\newblock \emph{Neurocomputing}, 72\penalty0 (7--9):\penalty0 1508--1524, 2009.

\bibitem[Deisenroth et~al.(2011)Deisenroth, Rasmussen, and
  Fox]{Deisenroth+Rasmussen+Fox:2011}
Marc~Peter Deisenroth, Carl~Edward Rasmussen, and Dieter Fox.
\newblock Learning to control a low-cost manipulator using data-efficient
  reinforcement learning.
\newblock In \emph{In 9th International Conference on Robotics: Science \&
  Systems}, 2011.

\bibitem[Dietterich(1998)]{Dietterich:1998}
Thomas~G. Dietterich.
\newblock Approximate statistical tests for comparing supervised classification
  learning algorithms.
\newblock \emph{Neural Computation}, 10\penalty0 (7):\penalty0 1895--1924,
  1998.

\bibitem[Diggle et~al.(1998)Diggle, Tawn, and Moyeed]{Diggle+Tawn+Moyeed:1998}
P.~J. Diggle, J.~A. Tawn, and R.~A. Moyeed.
\newblock Model-based geostatistics.
\newblock \emph{Journal of the Royal Statistical Society. Series C (Applied
  Statistics)}, 47\penalty0 (3):\penalty0 299--350, 1998.

\bibitem[Diggle and Ribeiro(2007)]{Diggle+Ribeiro:2007}
Peter~J. Diggle and Paulo~J. Ribeiro.
\newblock \emph{Model-based Geostatistics}.
\newblock Springer Science+Business Media, LLC, 2007.

\bibitem[Duane et~al.(1987)Duane, Kennedy, Pendleton, and
  Roweth]{Duane+Kennedy+Pendleton+Roweth:1987}
Simon Duane, A.D. Kennedy, Brian~J. Pendleton, and Duncan Roweth.
\newblock Hybrid {Monte} {Carlo}.
\newblock \emph{Physics Letters B}, 195\penalty0 (2):\penalty0 216--222, 1987.

\bibitem[Elliot et~al.(2001)Elliot, Wakefield, Best, and
  Briggs]{Elliot+Wakefield+Best+Briggs:2001}
P.~Elliot, Jon Wakefield, Nicola Best, and David Briggs, editors.
\newblock \emph{Spatial Epidemiology Methods and Applications}.
\newblock Oxford University Press, 2001.

\bibitem[Finkenst\"{a}dt et~al.(2007)Finkenst\"{a}dt, Held, and
  Isham]{Finkenstadt+Held+Isham:2007}
B\"{a}rbel Finkenst\"{a}dt, Leonhard Held, and Valerie Isham.
\newblock \emph{Statistical Methods for Spatio-Temporal Systems}.
\newblock Chapman \& Hall/CRC, 2007.

\bibitem[Fuentes and Raftery(2005)]{Fuentes+Raftery:2005}
Montserrat Fuentes and Adrian~E. Raftery.
\newblock Model evaluation and spatial interpolation by {Bayesian} combination
  of observations with outputs from numerical models.
\newblock \emph{Biometrics}, 66:\penalty0 36--45, 2005.

\bibitem[Furrer et~al.(2006)Furrer, Genton, and
  Nychka]{Furrer+Genton+Nychka:2006}
Reinhard Furrer, Marc~G. Genton, and Douglas Nychka.
\newblock Covariance tapering for interpolation of large spatial datasets.
\newblock \emph{Journal of Computational and Graphical Statistics}, 15\penalty0
  (3):\penalty0 502--523, September 2006.

\bibitem[Gaspari and Cohn(1999)]{Gaspari+Cohn:1999}
G.~Gaspari and S.E. Cohn.
\newblock Construction of correlation functions in two and three dimensions.
\newblock \emph{Quarterly Journal of the Royal Meteorological Society},
  125\penalty0 (554):\penalty0 723--757, January 1999.

\bibitem[Gelfand et~al.(1992)Gelfand, Dey, and Chang]{Gelfand+Dey+Chang:1992}
Alan~E. Gelfand, D.~K. Dey, and H.~Chang.
\newblock Model determination using predictive distributions with
  implementation via sampling-based methods (with discussion).
\newblock In J.~M. Bernardo, J.~O. Berger, A.~P. Dawid, and A.~F.~M. Smith,
  editors, \emph{Bayesian Statistics 4}, pages 147--167. Oxford University
  Press, 1992.

\bibitem[Gelfand et~al.(2010)Gelfand, Diggle, Fuentes, and
  Guttorp]{Gelfand+Diggle+Fuentes+Guttorp:2010}
Alan~E. Gelfand, Peter~J. Diggle, Montserrat Fuentes, and Peter Guttorp.
\newblock \emph{Handbook of Spatial Statistics}.
\newblock CRC Press, 2010.

\bibitem[Gelman(2006)]{Gelman:2006}
Andrew Gelman.
\newblock Prior distributions for variance parameters in hierarchical models.
\newblock \emph{Bayesian Analysis}, 1\penalty0 (3):\penalty0 515--533, 2006.

\bibitem[Gelman et~al.(2013)Gelman, Carlin, Stern, Dunson, Vehtari, and
  Rubin]{Gelman+etal+BDA3:2013}
Andrew Gelman, John~B. Carlin, Hal~S. Stern, David~B. Dunson, Aki Vehtari, and
  Donald~B. Rubin.
\newblock \emph{Bayesian Data Analysis}.
\newblock Chapman \& Hall/CRC, third edition, 2013.

\bibitem[Gelman et~al.(2014)Gelman, Hwang, and
  Vehtari]{Gelman+Hwang+Vehtari:2014}
Andrew Gelman, Jessica Hwang, and Aki Vehtari.
\newblock Understanding predictive information criteria for {Bayesian} models.
\newblock \emph{Statistics and Computing}, 24\penalty0 (6):\penalty0 997--1016,
  2014.

\bibitem[Geweke(1989)]{Geweke:1989}
John Geweke.
\newblock Bayesian inference in econometric models using {Monte} {Carlo}
  integration.
\newblock \emph{Econometrica}, 57\penalty0 (6):\penalty0 721--741, 1989.

\bibitem[Gibbs and Mackay(2000)]{Gibbs+MacKay:2000}
Mark~N. Gibbs and David J.~C. Mackay.
\newblock Variational {Gaussian} process classifiers.
\newblock \emph{IEEE Transactions on Neural Networks}, 11\penalty0
  (6):\penalty0 1458--1464, 2000.

\bibitem[Gilks et~al.(1996)Gilks, Richardson, and
  Spiegelhalter]{Gilks+Richardson+Spiegelhalter:1996}
W.R. Gilks, S.~Richardson, and D.J. Spiegelhalter.
\newblock \emph{Markov Chain Monte Carlo in Practice}.
\newblock Chapman \& Hall, 1996.

\bibitem[Gneiting(1999)]{Gneiting:1999}
Tilmann Gneiting.
\newblock Correlation functions for atmospheric data analysis.
\newblock \emph{Quarterly Journal of the Royal Meteorological Society},
  125:\penalty0 2449--2464, 1999.

\bibitem[Gneiting(2002)]{Gneiting:2002}
Tilmann Gneiting.
\newblock Compactly supported correlation functions.
\newblock \emph{Journal of Multivariate Analysis}, 83:\penalty0 493--508, 2002.

\bibitem[Goldberg et~al.(1997)Goldberg, Williams, and Bishop]{Goldberg:1997}
Paul~W Goldberg, Christopher~KI Williams, and Christopher~M Bishop.
\newblock Regression with input-dependent noise: A {Gaussian} process
  treatment.
\newblock In Y.~Weiss, B.~Sch{\"o}lkopf, and J.~Platt, editors, \emph{Advances
  in Neural Information Processing Systems}, volume~10, pages 493--499, 1997.

\bibitem[Grewal and Andrews(2001)]{Grewal+Andrews:2001}
Mohinder~S. Grewal and Angus~P. Andrews.
\newblock \emph{Kalman Filtering: Theory and Practice Using Matlab}.
\newblock Wiley Interscience, second edition, 2001.

\bibitem[Hartikainen and S\"arkk\"a(2010)]{Hartikainen+Sarkka:2010}
Jouni Hartikainen and Simo S\"arkk\"a.
\newblock Kalman filtering and smoothing solutions to temporal {G}aussian
  process regression models.
\newblock In \emph{Proceedings of IEEE International Workshop on Machine
  Learning for Signal Processing (MLSP)}, 2010.

\bibitem[Harville(1997)]{Harville:1997}
David~A. Harville.
\newblock \emph{Matrix Algebra From a Statistician's Perspective}.
\newblock Springer-Verlag, 1997.

\bibitem[Held et~al.(2010)Held, Schr{\"o}dle, and Rue]{Held+Schrodle+Rue:2010}
Leonhard Held, Birgit Schr{\"o}dle, and H{\aa}vard Rue.
\newblock Posterior and cross-validatory predictive checks: {A} comparison of
  {MCMC} and {INLA}.
\newblock In Thomas Kneib and Gerhard Tutz, editors, \emph{Statistical
  Modelling and Regression Structures}, pages 91--110. Springer, 2010.

\bibitem[Henao and Winther(2012)]{Henao+Winther:2012}
Ricardo Henao and Ole Winther.
\newblock Predictive active set selection methods for {Gaussian} processes.
\newblock \emph{Neurocomputing}, 80\penalty0 (0):\penalty0 10--18, 2012.

\bibitem[Hoffman and Gelman(2014)]{Hoffman+Gelman:2014}
Matt~D Hoffman and Andrew Gelman.
\newblock The no-{U}-turn sampler: {A}daptively setting path lengths in
  {Hamiltonian} {Monte} {Carlo}.
\newblock \emph{Journal of Machine Learning Research}, 15:\penalty0 1593--1623,
  Apr 2014.

\bibitem[Honkela et~al.(2011)Honkela, Gao, Ropponen, Rattray, and
  Lawrence]{Honkela+Gao+Ropponen+Rattray+Lawrence:2011}
Antti Honkela, Pei Gao, Jonatan Ropponen, Magnus Rattray, and Neil~D. Lawrence.
\newblock tigre: Transcription factor inference through gaussian process
  reconstruction of expression for bioconductor.
\newblock \emph{Bioinformatics}, 27\penalty0 (7):\penalty0 1026--1027, 2011.

\bibitem[Ibrahim et~al.(2001)Ibrahim, Chen, and Sinha]{Ibrahim+Chen+Sinha:2001}
Joseph~G. Ibrahim, Ming-Hui Chen, and Debajyoti Sinha.
\newblock \emph{Bayesian Survival Analysis}.
\newblock Springer, 2001.

\bibitem[Joensuu et~al.(2012)Joensuu, Vehtari, Riihim{\"a}ki, Nishida, Steigen,
  Brabec, Plank, Nilsson, Cirilli, Braconi, Bordoni, Magnusson, Linke,
  Sufliarsky, Massimo, Jonasson, {Paolo Dei Tos}, and
  Rutkowski]{Joensuu+etal:2012a}
Heikki Joensuu, Aki Vehtari, Jaakko Riihim{\"a}ki, Toshirou Nishida, Sonja~E
  Steigen, Peter Brabec, Lukas Plank, Bengt Nilsson, Claudia Cirilli, Chiara
  Braconi, Andrea Bordoni, Magnus~K Magnusson, Zdenek Linke, Jozef Sufliarsky,
  Federico Massimo, Jon~G Jonasson, Angelo {Paolo Dei Tos}, and Piotr
  Rutkowski.
\newblock Risk of gastrointestinal stromal tumour recurrence after surgery: an
  analysis of pooled population-based cohorts.
\newblock \emph{The Lancet Oncology}, 13\penalty0 (3):\penalty0 265--274, 2012.

\bibitem[Joensuu et~al.(2014)Joensuu, Reichardt, Eriksson, Hall, and
  Vehtari]{Joensuu+Reichardt+Eriksson+Hall+Vehtari:2014}
Heikki Joensuu, Peter Reichardt, Mikael Eriksson, Kirsten~Sundby Hall, and Aki
  Vehtari.
\newblock Gastrointestinal stromal tumor: {A} method for optimizing the timing
  of {CT} scans in the follow-up of cancer patients.
\newblock \emph{Radiology}, 271\penalty0 (1):\penalty0 96--106, 2014.

\bibitem[Juntunen et~al.(2012)Juntunen, Vanhatalo, Peltonen, and
  M{\"a}ntyniemi]{Juntunen+Vanhatalo+Peltonen+Mantyniemi:2012}
Teppo Juntunen, Jarno Vanhatalo, Heikki Peltonen, and Samu M{\"a}ntyniemi.
\newblock Bayesian spatial multispecies modelling to assess pelagic fish stocks
  from acoustic- and trawl-survey data.
\newblock \emph{ICES Journal of Marine Science}, 69\penalty0 (1):\penalty0
  95--104, 2012.

\bibitem[Jyl{\"a}nki et~al.(2011)Jyl{\"a}nki, Vanhatalo, and
  Vehtari]{Jylanki+Vanhatalo+Vehtari:2011}
Pasi Jyl{\"a}nki, Jarno Vanhatalo, and Aki Vehtari.
\newblock Robust {Gaussian} process regression with a {Student-$t$} likelihood.
\newblock \emph{Journal of Machine Learning Research}, 12:\penalty0 3227--3257,
  2011.

\bibitem[Kalliom{\"a}ki et~al.(2005)Kalliom{\"a}ki, Vehtari, and
  Lampinen]{Kalliomaki+Vehtari+Lampinen:2005}
Ilkka Kalliom{\"a}ki, Aki Vehtari, and Jouko Lampinen.
\newblock Shape analysis of concrete aggregates for statistical quality
  modeling.
\newblock \emph{Machine Vision and Applications}, 16\penalty0 (3):\penalty0
  197--201, 2005.

\bibitem[Kass and Raftery(1995)]{Kass+Raftery:1995}
Robert~E. Kass and Adrian~E. Raftery.
\newblock Bayes factors.
\newblock \emph{Journal of the American Statistical Association}, 90\penalty0
  (430):\penalty0 773--795, 1995.

\bibitem[Kaufman et~al.(2008)Kaufman, Schervish, and
  Nychka]{Kaufman+Schervish+Nychka:2008}
Cari~G. Kaufman, Mark~J. Schervish, and Douglas~W. Nychka.
\newblock Covariance tapering for likelihood-based estimation in large spatial
  data sets.
\newblock \emph{Journal of the American Statistical Association}, 103\penalty0
  (484):\penalty0 1545--1555, 2008.

\bibitem[Kennedy and O'Hagan(2001)]{Kennedy+OHagan:2001}
Mark~C. Kennedy and Anthony O'Hagan.
\newblock Bayesian calibration of computer models.
\newblock \emph{Journal of the Royal Statistical Society, B}, 63\penalty0
  (3):\penalty0 425--464, 2001.

\bibitem[Kneib(2006)]{Kneib:2006}
Thomas Kneib.
\newblock Mixed model-based inference in geoadditive hazard regression for
  interval censored survival times.
\newblock \emph{Computational Statistics and Data Analysis}, 51\penalty0
  (2):\penalty0 777--792, 2006.

\bibitem[Kot(2001)]{Kot:2001}
M.~Kot.
\newblock \emph{Elements of Mathematical Ecology}.
\newblock Cambridge University Press, 2001.

\bibitem[Kuss(2006)]{Kuss:2006}
Malte Kuss.
\newblock \emph{Gaussian Process Models for Robust Regression, Classification,
  and Reinforcement Learning}.
\newblock PhD thesis, Technische Universit{\"a}t Darmstadt, 2006.

\bibitem[Kuss and Rasmussen(2005)]{Kuss+Rasmussen:2005}
Malte Kuss and Carl~Edward Rasmussen.
\newblock Assessing approximate inference for binary {Gaussian} process
  classification.
\newblock \emph{Journal of Machine Learning Research}, 6:\penalty0 1679--1704,
  October 2005.

\bibitem[Lawrence(2007)]{Lawrence:2007}
Neil Lawrence.
\newblock Learning for larger datasets with the {Gaussian} process latent
  variable model.
\newblock In M.~Meila and X.~Shen, editors, \emph{Proceedings of the Eleventh
  International Workshop on Artificial Intelligence and Statistics}. Omnipress,
  2007.

\bibitem[Lawson(2001)]{Lawson:2001}
Andrew~B. Lawson.
\newblock \emph{Statistical Methods in Spatial Epidemology}.
\newblock John Wiley \& Sons, Ltd, 2001.

\bibitem[Martino et~al.(2011)Martino, Akerkar, and Rue]{Martino:2011}
Sara Martino, Rupali Akerkar, and H{\aa}vard Rue.
\newblock Approximate {Bayesian} inference for survival models.
\newblock \emph{Scandinavian Journal of Statistics}, 38\penalty0 (3):\penalty0
  514--528, 2011.

\bibitem[Matheron(1973)]{Matheron:1973}
G.~Matheron.
\newblock The intrinsic random functions and their applications.
\newblock \emph{Advances in Applied Probability}, 5\penalty0 (3):\penalty0
  439--468, December 1973.

\bibitem[Mathworks(2010)]{MATLAB:2010}
Mathworks.
\newblock \emph{MATLAB Getting Started Quide}.
\newblock The MathWorks, Inc., fifteenth edition, 2010.

\bibitem[Minka(2001)]{Minka:2001}
Thomas Minka.
\newblock \emph{A Family of Algorithms for Approximate {Bayesian} Inference}.
\newblock PhD thesis, Massachusetts Institute of Technology, 2001.

\bibitem[{M\o ller} et~al.(1998){M\o ller}, Syversveen, and
  Waagepetersen]{Moller+Syversveen+Waagepetersen:1998}
Jesper {M\o ller}, Anne~Randi Syversveen, and Rasmus~Plenge Waagepetersen.
\newblock Log {Gaussian} {Cox} processes.
\newblock \emph{Scandinavian Journal of Statistics}, 25:\penalty0 451--482,
  September 1998.

\bibitem[Moala and O'Hagan(2010)]{Moala+OHagan:2010}
Fernando~A Moala and Anthony O'Hagan.
\newblock Elicitation of multivariate prior distributions: a nonparametric
  {Bayesian} approach.
\newblock \emph{Journal of Statistical Planning and Inference}, 140:\penalty0
  1635--1655, 2010.

\bibitem[Moreaux(2008)]{Moreaux:2008}
G.~Moreaux.
\newblock Compactly supported radial covariance functions.
\newblock \emph{Journal of Geodecy}, 82\penalty0 (7):\penalty0 431--443, July
  2008.

\bibitem[Mullahy(1986)]{Mullahy:1986}
J~Mullahy.
\newblock Specification and testing of some modified count data models.
\newblock \emph{Journal of Econometrics}, 33:\penalty0 341--365, 1986.

\bibitem[Murray and Adams(2010)]{Murray+Adams:2010}
Iain Murray and Ryan~Prescott Adams.
\newblock Slice sampling covariance hyperparameters of latent {Gaussian}
  models.
\newblock In J.~Lafferty, C.~K.~I. Williams, R.~Zemel, J.~Shawe-Taylor, and
  A.~Culotta, editors, \emph{Advances in Neural Information Processing Systems
  23}, pages 1732--1740, 2010.

\bibitem[Murray et~al.(2010)Murray, Adams, and
  MacKay]{Murray+Adams+MacKay:2010}
Iain Murray, Ryan~Prescott Adams, and David~J.C. MacKay.
\newblock Elliptical slice sampling.
\newblock \emph{JMLR: Workshop and Conference Proceedings}, 9:\penalty0
  541--548, 2010.

\bibitem[Myllym{\"a}ki et~al.(2014)Myllym{\"a}ki, S{\"a}rkk{\"a}, and
  Vehtari]{Myllymaki+Sarkka+Vehtari:2014}
Mari Myllym{\"a}ki, Aila S{\"a}rkk{\"a}, and Aki Vehtari.
\newblock Hierarchical second-order analysis of replicated spatial point
  patterns with non-spatial covariates.
\newblock \emph{Spatial Statistics}, 8:\penalty0 104--121, 2014.

\bibitem[Nabney(2001)]{Nabney:2001}
Ian~T. Nabney.
\newblock \emph{NETLAB: Algorithms for Pattern Recognition}.
\newblock Springer, 2001.

\bibitem[Neal(1998)]{Neal:1998}
Radford Neal.
\newblock Regression and classification using {Gaussian} process priors.
\newblock In J.~M. Bernardo, J.~O. Berger, A.~P. David, and A.~P.~M. Smith,
  editors, \emph{Bayesian Statistics 6}, pages 475--501. Oxford University
  Press, 1998.

\bibitem[Neal(1996)]{Neal:1996a}
Radford~M. Neal.
\newblock \emph{Bayesian Learning for Neural Networks}.
\newblock Springer, 1996.

\bibitem[Neal(1997)]{Neal:1997}
Radford~M. Neal.
\newblock {Monte} {Carlo} {Implementation} of {Gaussian} {Process} {Models} for
  {Bayesian} {Regression} and {Classification}.
\newblock Technical Report 9702, Dept. of statistics and Dept. of Computer
  Science, University of Toronto, January 1997.

\bibitem[Neal(2003)]{Neal:2003}
Radford~M. Neal.
\newblock Slice sampling.
\newblock \emph{The Annals of Statistics}, 31\penalty0 (3):\penalty0 705--767,
  2003.

\bibitem[Nickisch and Rasmussen(2008)]{Nickisch+Rasmussen:2008}
Hannes Nickisch and Carl~Edward Rasmussen.
\newblock Approximations for binary {Gaussian} process classification.
\newblock \emph{Journal of Machine Learning Research}, 9:\penalty0 2035--2078,
  October 2008.

\bibitem[{Octave community}(2012)]{octave:2012}
{Octave community}.
\newblock {GNU/Octave}, 2012.
\newblock URL \url{www.gnu.org/software/octave/}.

\bibitem[O'Hagan(1978)]{OHagan:1978}
Anthony O'Hagan.
\newblock Curve fitting and optimal design for prediction.
\newblock \emph{Journal of the Royal Statistical Society. Series B.},
  40\penalty0 (1):\penalty0 1--42, 1978.

\bibitem[O'Hagan(1979)]{OHagan:1979}
Anthony O'Hagan.
\newblock On outlier rejection phenomena in {Bayes} inference.
\newblock \emph{Journal of the Royal Statistical Society. Series B.},
  41\penalty0 (3):\penalty0 358--367, 1979.

\bibitem[Opper and Archambeau(2009)]{Opper+Archambeau:2009}
Manfred Opper and C\'edric Archambeau.
\newblock The variational {Gaussian} approximation revisited.
\newblock \emph{Neural Computation}, 21\penalty0 (3):\penalty0 786--792, March
  2009.

\bibitem[Opper and Winther(2000)]{Opper+Winther:2000}
Manfred Opper and Ole Winther.
\newblock Gaussian processes for classification: Mean-field algorithms.
\newblock \emph{Neural Computation}, 12\penalty0 (11):\penalty0 2655--2684,
  2000.

\bibitem[{Qui\~nonero-Candela} and
  Rasmussen(2005)]{Quinonero-Candela+Rasmussen:2005}
Joaquin {Qui\~nonero-Candela} and Carl~Edward Rasmussen.
\newblock A unifying view of sparse approximate {Gaussian} process regression.
\newblock \emph{Journal of Machine Learning Research}, 6\penalty0 (3):\penalty0
  1939--1959, December 2005.

\bibitem[Rantonen et~al.(2012)Rantonen, Vehtari, Karppinen, Luoto,
  Viikari-Juntura, Hupli, Malmivaara, and Taimela]{Rantonen+Vehtari+etal:2012}
Jarmo Rantonen, Aki Vehtari, Jaro Karppinen, Satu Luoto, Eira Viikari-Juntura,
  Markku Hupli, Antti Malmivaara, and Simo Taimela.
\newblock The effectiveness of two active interventions compared to self-care
  advice in employees with non-acute low back symptoms. {A} randomised,
  controlled trial with a 4-year follow-up in the occupational health setting.
\newblock \emph{Occupational and Environmental Medicine}, 69\penalty0
  (1):\penalty0 12--20, 2012.

\bibitem[Rantonen et~al.(2014)Rantonen, Vehtari, Karppinen, Luoto,
  Viikari-Juntura, Hupli, Malmivaara, and Taimela]{Rantonen+Vehtari+etal:2014}
Jarmo Rantonen, Aki Vehtari, Jaro Karppinen, Satu Luoto, Eira Viikari-Juntura,
  Markku Hupli, Antti Malmivaara, and Simo Taimela.
\newblock Face-to-face information in addition to a booklet versus a booklet
  alone for treating mild back pain, a randomized controlled trial.
\newblock \emph{Scandinavian journal of Work Environment and Health},
  40\penalty0 (2):\penalty0 156--166, 2014.

\bibitem[Rantonen et~al.(2011)Rantonen, Luoto, Vehtari, Hupli, Karppinen,
  Malmivaara, and Taimela]{Rantonen+etal:2011}
Jorma Rantonen, Satu Luoto, Aki Vehtari, Markku Hupli, Jaro Karppinen, Antti
  Malmivaara, and Simo Taimela.
\newblock The effectiveness of two active interventions compared to self-care
  advice in employees with non-acute low back symptoms. a randomised,
  controlled trial with a 4-year follow-up in the occupational health setting.
\newblock \emph{Occupational and Environmental Medicine}, 2011.

\bibitem[Rasmussen(1996)]{Rasmussen:1996}
Carl~Edward Rasmussen.
\newblock \emph{Evaluations of {Gaussian} Processes and Other Methods for
  Non-linear Regression}.
\newblock PhD thesis, University of Toronto, 1996.

\bibitem[Rasmussen and Nickisch(2010)]{Rasmussen+Nickisch:2010}
Carl~Edward Rasmussen and Hannes Nickisch.
\newblock Gaussian processes for machine learning ({GPML}) toolbox.
\newblock \emph{Journal of Machine Learning Research}, 11:\penalty0 3011--3015,
  September 2010.

\bibitem[Rasmussen and Williams(2006)]{Rasmussen+Williams:2006}
Carl~Edward Rasmussen and Christopher K.~I. Williams.
\newblock \emph{Gaussian Processes for Machine Learning}.
\newblock The MIT Press, 2006.

\bibitem[Rathbun and Cressie(1994)]{Rathbun+Cressie:1994}
Stephen~L. Rathbun and Noel Cressie.
\newblock Asymptotic properties of estimators for the parameters of spatial
  inhomogeneous poisson point processes.
\newblock \emph{Advances in Applied Probability}, 26\penalty0 (1):\penalty0
  122--154, March 1994.

\bibitem[Richardson(2002)]{Richardson:2002}
Sylvia Richardson.
\newblock Discussion to `bayesian measures of model complexity and fit' by
  spiegelhalter et al.
\newblock \emph{Journal of the Royal Statistical Society. Series B (Statistical
  Methodology)}, 64\penalty0 (4):\penalty0 626--627, 2002.

\bibitem[Richardson(2003)]{Richardson:2003}
Sylvia Richardson.
\newblock Spatial models in epidemiological applications.
\newblock In Peter~J. Green, Nils~Lid Hjort, and Sylvia Richardson, editors,
  \emph{Highly Structured Stochastic Systems}, pages 237--259. Oxford
  University Press, 2003.

\bibitem[Riihim{\"a}ki and Vehtari(2012)]{Riihimaki+Vehtari:2012}
Riihim{\"a}ki and Aki Vehtari.
\newblock Laplace approximation for logistic gp density estimation.
\newblock Technical report, Department of Biomedical Engineering and
  Computational Science, Aalto University, 2012.
\newblock arXiv preprint arXiv:1211.0174.

\bibitem[Riihim{\"a}ki and Vehtari(2014)]{Riihimaki+Vehtari:2014}
Riihim{\"a}ki and Aki Vehtari.
\newblock Laplace approximation for logistic {Gaussian} process density
  estimation and regression.
\newblock \emph{Bayesian analysis}, 9\penalty0 (2):\penalty0 425--448, 2014.

\bibitem[Riihim\"{a}ki and Vehtari(2010)]{Riihimaki+Vehtari:2010}
Jaakko Riihim\"{a}ki and Aki Vehtari.
\newblock Gaussian processes with monotonicity information.
\newblock \emph{Journal of Machine Learning Research: Workshop and Conference
  Proceedings}, 9:\penalty0 645--652, 2010.

\bibitem[Riihim\"{a}ki et~al.(2013)Riihim\"{a}ki, Jyl\"anki, and
  Vehtari]{Riihimaki+Jylanki+Vehtari:2013}
Jaakko Riihim\"{a}ki, Pasi Jyl\"anki, and Aki Vehtari.
\newblock Nested expectation propagation for {Gaussian} process classification
  with a multinomial probit likelihood.
\newblock \emph{Journal of Machine Learning Research}, 14:\penalty0 75--109,
  2013.

\bibitem[Robert and Casella(2004)]{Robert+Casella:2004}
Christian~P. Robert and George Casella.
\newblock \emph{Monte Carlo Statistical Methods}.
\newblock Springer, second edition, 2004.

\bibitem[Rue et~al.(2009)Rue, Martino, and Chopin]{Rue+Martino+Chopin:2009}
H{\aa}vard Rue, Sara Martino, and Nicolas Chopin.
\newblock Approximate {Bayesian} inference for latent {Gaussian} models by
  using integrated nested {Laplace} approximations.
\newblock \emph{Journal of the Royal statistical Society B}, 71\penalty0
  (2):\penalty0 1--35, 2009.

\bibitem[Sanchez and Sanchez(2005)]{Sanchez+Sanchez:2005}
Susan~M. Sanchez and Paul~J. Sanchez.
\newblock Very large fractional factorials and central composite designs.
\newblock \emph{ACM Transactions on Modeling and Computer Simulation},
  15:\penalty0 362--377, 2005.

\bibitem[Sans{\`o} and Schuh(1987)]{Sanso+Schuh:1987}
F.~Sans{\`o} and W.-D. Schuh.
\newblock Finite covariance functions.
\newblock \emph{Journal of Geodesy}, 61\penalty0 (4):\penalty0 331--347, 1987.

\bibitem[S{\"a}rkk{\"a}(2013)]{Sarkka:2013}
Simo S{\"a}rkk{\"a}.
\newblock \emph{Bayesian Filtering and Smoothing}, volume~3 of \emph{Institute
  of Mathematical Statistics Textbooks}.
\newblock Cambridge University Press, 2013.

\bibitem[S\"arkk\"a et~al.(2013)S\"arkk\"a, Solin, and
  Hartikainen]{Sarkka+Solin+Hartikainen:2013}
Simo S\"arkk\"a, Arno Solin, and Jouni Hartikainen.
\newblock Spatiotemporal learning via infinite-dimensional {B}ayesian filtering
  and smoothing.
\newblock \emph{{IEEE} Signal Processing Magazine}, 30\penalty0 (4):\penalty0
  51--61, 2013.

\bibitem[Seeger et~al.(2003)Seeger, Williams, and
  Lawrence]{Seeger+Williams+Lawrence:2003}
Mathias Seeger, Christopher K.~I. Williams, and Neil Lawrence.
\newblock Fast forward selection to speed up sparse {Gaussian} process
  regression.
\newblock In Christopher~M. Bishop and Brendan~J. Frey, editors, \emph{Ninth
  International Workshop on Artificial Intelligence and Statistics}. Society
  for Artificial Intelligence and Statistics, 2003.

\bibitem[Seeger(2005)]{Seeger:2005}
Matthias Seeger.
\newblock Expectation propagation for exponential families.
\newblock Technical report, Max Planck Institute for Biological Cybernetics,
  T\"ubingen, Germany, 2005.

\bibitem[Snelson(2007)]{Snelson:2007b}
Edward Snelson.
\newblock \emph{Flexible and Efficient {Gaussian} Process Models for Machine
  Learning}.
\newblock PhD thesis, University College London, 2007.

\bibitem[Snelson and Ghahramani(2006)]{Snelson+Ghahramani:2006}
Edward Snelson and Zoubin Ghahramani.
\newblock Sparse {Gaussian} process using pseudo-inputs.
\newblock In Y.~Weiss, B.~Sch{\"o}lkopf, and J.~Platt, editors, \emph{Advances
  in Neural Information Processing Systems 18}. The MIT Press, 2006.

\bibitem[Snelson and Ghahramani(2007)]{Snelson+Ghahramani:2007a}
Edward Snelson and Zoubin Ghahramani.
\newblock Local and global sparse {Gaussian} process approximations.
\newblock In Marina Meila and Xiaotong Shen, editors, \emph{Artificial
  Intelligence and Statistics 11}. Omnipress, 2007.

\bibitem[Snelson et~al.(2004)Snelson, Rasmussen, and
  Ghahramani]{Snelson+Rasmussen+Ghahramani:2004}
Edward Snelson, Carl~Edward Rasmussen, and Zoubin Ghahramani.
\newblock Warped {Gaussian} processes.
\newblock In T.~G. Diettrich, S.~Becker, and Z.~Ghahramani, editors,
  \emph{Advances in Neural Information Processing Systems 14}. The MIT Press,
  2004.

\bibitem[Solin and S\"arkk\"a(2014a)]{Solin+Sarkka:2014}
Arno Solin and Simo S\"arkk\"a.
\newblock Explicit link between periodic covariance functions and state space
  models.
\newblock In \emph{Proceedings of the Seventeenth International Conference on
  Artificial Intelligence and Statistics}, volume~33 of \emph{{JMLR W\&CP}},
  pages 904--912, 2014a.

\bibitem[Solin and S{\"a}rkk{\"a}(2014b)]{Solin+Sarkka:2014-MLSP}
Arno Solin and Simo S{\"a}rkk{\"a}.
\newblock Gaussian quadratures for state space approximation of scale mixtures
  of squared exponential covariance functions.
\newblock In \emph{Proceedings of {IEEE} International Workshop on Machine
  Learning for Signal Processing ({MLSP})}, 2014b.

\bibitem[Spiegelhalter et~al.(2002)Spiegelhalter, Best, Carlin, and {van der
  Linde}]{Spiegelhalter+Best+Carlin+Linde:2002}
David~J. Spiegelhalter, Nicola~G. Best, Bradley~P. Carlin, and Angelika {van
  der Linde}.
\newblock Bayesian measures of model complexity and fit.
\newblock \emph{Journal of the Royal Statistical Society B}, 64\penalty0
  (4):\penalty0 583--639, 2002.

\bibitem[Stegle et~al.(2008)Stegle, Fallert, MacKay, and ren
  Brage]{Stegle+Fallert+MacKay+Brage:2008}
Oliver Stegle, Sebastian~V. Fallert, David J.~C. MacKay, and S\o ren Brage.
\newblock Gaussian process robust regression for noisy heart rate data.
\newblock \emph{Biomedical Engineering, IEEE Transactions on}, 55\penalty0
  (9):\penalty0 2143--2151, September 2008.
\newblock ISSN 0018-9294.

\bibitem[Sundararajan and Keerthi(2001)]{Sundararajan+Keerthi:2001a}
S.~Sundararajan and S.~S. Keerthi.
\newblock Predictive approaches for choosing hyperparameters in {Gaussian}
  processes.
\newblock \emph{Neural Computation}, 13\penalty0 (5):\penalty0 1103--1118, May
  2001.

\bibitem[Thompson and Neal(2010)]{Thompson+Neal:2010}
Madeleine Thompson and Radford~M Neal.
\newblock Covariance-adaptive slice sampling.
\newblock \emph{arXiv preprint arXiv:1003.3201}, 2010.

\bibitem[Tierney and Kadane(1986)]{Tierney+Kadane:1986}
Luke Tierney and Joseph~B. Kadane.
\newblock Accurate approximations for posterior moments and marginal densities.
\newblock \emph{Journal of the American Statistical Association}, 81\penalty0
  (393):\penalty0 82--86, 1986.

\bibitem[Tipping and Lawrence(2005)]{Tipping+Lawrence:2005}
Michael~E. Tipping and Neil~D. Lawrence.
\newblock Variational inference for {Student-$t$} models: Robust {Bayesian}
  interpolation and generalised component analysis.
\newblock \emph{Neurocomputing}, 69:\penalty0 123--141, 2005.

\bibitem[Titsias(2009)]{Titsias:2009}
Michalis~K. Titsias.
\newblock Variational learning of inducing variables in sparse {Gaussian}
  processes.
\newblock \emph{JMLR Workshop and Conference Proceedings}, 5:\penalty0
  567--574, 2009.

\bibitem[Tokdar and Ghosh(2007)]{Tokdar+Ghosh:2007}
Surya~T. Tokdar and Jayanta~K. Ghosh.
\newblock Posterior consistency of logistic {Gaussian} process priors in
  density estimation.
\newblock \emph{Journal of Statistical Planning and Inference}, 137:\penalty0
  34--42, 2007.

\bibitem[Tokdar et~al.(2010)Tokdar, Zhu, and Ghosh]{Tokdar+Zhu+Ghosh:2010}
Surya~T. Tokdar, Yu~M. Zhu, and Jayanta~K. Ghosh.
\newblock Bayesian density regression with logistic {Gaussian} process and
  subspace projection.
\newblock \emph{Bayesian Analysis}, 5\penalty0 (2):\penalty0 319--344, 2010.

\bibitem[Vanhatalo and Vehtari(2007)]{Vanhatalo+Vehtari:2007}
Jarno Vanhatalo and Aki Vehtari.
\newblock Sparse log {Gaussian} processes via {MCMC} for spatial epidemiology.
\newblock \emph{JMLR Workshop and Conference Proceedings}, 1:\penalty0 73--89,
  2007.

\bibitem[Vanhatalo and Vehtari(2008)]{Vanhatalo+Vehtari:2008}
Jarno Vanhatalo and Aki Vehtari.
\newblock Modelling local and global phenomena with sparse {Gaussian}
  processes.
\newblock In David~A. McAllester and Petri Myllym{\"a}ki, editors,
  \emph{Proceedings of the 24th Conference on Uncertainty in Artificial
  Intelligence}, pages 571--578, 2008.

\bibitem[Vanhatalo and Vehtari(2010)]{Vanhatalo+Vehtari:2010}
Jarno Vanhatalo and Aki Vehtari.
\newblock Speeding up the binary {Gaussian} process classification.
\newblock In Peter Grünwald and Peter Spirtes, editors, \emph{Proceedings of
  the 26th Conference on Uncertainty in Artificial Intelligence}, pages 1--9,
  2010.

\bibitem[Vanhatalo et~al.(2009)Vanhatalo, Jyl\"{a}nki, and
  Vehtari]{Vanhatalo+Jylanki+Vehtari:2009}
Jarno Vanhatalo, Pasi Jyl\"{a}nki, and Aki Vehtari.
\newblock Gaussian process regression with student-t likelihood.
\newblock In Y.~Bengio, D.~Schuurmans, J.~Lafferty, C.~K.~I. Williams, and
  A.~Culotta, editors, \emph{Advances in Neural Information Processing Systems
  22}, pages 1910--1918. NIPS foundation, 2009.

\bibitem[Vanhatalo et~al.(2010)Vanhatalo, Pietil\"{a}inen, and
  Vehtari]{Vanhatalo+Pietilainen+Vehtari:2010}
Jarno Vanhatalo, Ville Pietil\"{a}inen, and Aki Vehtari.
\newblock Approximate inference for disease mapping with sparse {Gaussian}
  processes.
\newblock \emph{Statistics in Medicine}, 29\penalty0 (15):\penalty0 1580--1607,
  2010.

\bibitem[Vanhatalo et~al.(2013)Vanhatalo, Riihim{\"a}ki, Hartikainen,
  Jyl{\"a}nki, Tolvanen, and Vehtari]{Vanhatalo+gpstuff:2013}
Jarno Vanhatalo, Jaakko Riihim{\"a}ki, Jouni Hartikainen, Pasi Jyl{\"a}nki,
  Ville Tolvanen, and Aki Vehtari.
\newblock {GPstuff}: {Bayesian} modeling with {Gaussian} processes.
\newblock \emph{Journal of Machine Learning Research}, 14:\penalty0 1175--1179,
  2013.

\bibitem[Vehtari(2001)]{Vehtari:2001}
Aki Vehtari.
\newblock \emph{Bayesian Model Assessment and Selection Using Expected
  Utilities}.
\newblock PhD thesis, Helsinki University of Technology, 2001.

\bibitem[Vehtari and Gelman(2015)]{Vehtari+Gelman:2015}
Aki Vehtari and Andrew Gelman.
\newblock Pareto smoothed importance sampling, 2015.

\bibitem[Vehtari and Lampinen(2002)]{Vehtari+Lampinen:2002}
Aki Vehtari and Jouko Lampinen.
\newblock Bayesian model assessment and comparison using cross-validation
  predictive densities.
\newblock \emph{Neural Computation}, 14\penalty0 (10):\penalty0 2439--2468,
  2002.

\bibitem[Vehtari and Ojanen(2012)]{Vehtari+Ojanen:2012}
Aki Vehtari and Janne Ojanen.
\newblock A survey of {Bayesian} predictive methods for model assessment,
  selection and comparison.
\newblock \emph{Statistics Surveys}, 6:\penalty0 142--228, 2012.

\bibitem[Vehtari et~al.(2014)Vehtari, Mononen, Tolvanen, and
  Winther]{Vehtari+etal:2014}
Aki Vehtari, Tommi Mononen, Ville Tolvanen, and Ole Winther.
\newblock {B}ayesian leave-one-out cross-validation approximations for
  {G}aussian latent variable models.
\newblock \emph{arXiv preprint arXiv:1412.7461}, 2014.

\bibitem[Vehtari et~al.(2015)Vehtari, Gelman, and
  Gabry]{Vehtari+Gelman+Gabry:2015b}
Aki Vehtari, Andrew Gelman, and Jonah Gabry.
\newblock Efficient implementation of leave-one-out cross-validation and {WAIC}
  for evaluating fitted {Bayesian} models, 2015.

\bibitem[Watanabe(2009)]{Watanabe:2009}
Sumio Watanabe.
\newblock \emph{Algebraic Geometry and Statistical Learning Theory}.
\newblock Cambridge University Press, 2009.

\bibitem[Watanabe(2010{\natexlab{a}})]{Watanabe:2010a}
Sumio Watanabe.
\newblock Equations of states in singular statistical estimation.
\newblock \emph{Neural Networks}, 23\penalty0 (1):\penalty0 20--34,
  2010{\natexlab{a}}.

\bibitem[Watanabe(2010{\natexlab{b}})]{Watanabe:2010d}
Sumio Watanabe.
\newblock Asymptotic equivalence of {Bayes} cross validation and widely
  applicable information criterion in singular learning theory.
\newblock \emph{Journal of Machine Learning Research}, 11:\penalty0 3571--3594,
  2010{\natexlab{b}}.

\bibitem[Wendland(1995)]{Wendland:1995}
Holger Wendland.
\newblock Piecewise polynomial, positive definite and compactly supported
  radial functions of minimal degree.
\newblock \emph{Advances in Computational Mathematics}, 4\penalty0
  (1):\penalty0 389--396, December 1995.

\bibitem[Wendland(2005)]{Wendland:2005}
Holger Wendland.
\newblock \emph{Scattered Data Approximation}.
\newblock Cambridge University Press, 2005.

\bibitem[Wiener(1949)]{Wiener:1949}
Norbert Wiener.
\newblock \emph{Extrapolation, Interpolation, and Smoothing of Stationary Time
  Series}.
\newblock MIT Press, 1949.

\bibitem[Williams and Rasmussen(1996)]{Williams+Rasmussen:1996}
C.~K.~I. Williams and C.~E. Rasmussen.
\newblock Gaussian processes for regression.
\newblock In D.~S. Touretzky, M.~C. Mozer, and M.~E. Hasselmo, editors,
  \emph{Advances in Neural Information Processing Systems 8}, pages 514--520.
  MIT Press, 1996.

\bibitem[Williams(1996)]{Williams:1996}
Christopher K.~I. Williams.
\newblock Computing with infinite networks.
\newblock In Michael~C. Mozer, Michael~I. Jordan, and Thomas Petsche, editors,
  \emph{Advances in Neural Information Processing Systems}, volume~9. The MIT
  Press, 1996.

\bibitem[Williams(1998)]{Williams:1998}
Christopher K.~I. Williams.
\newblock Computation with infinite neural networks.
\newblock \emph{Neural Computation}, 10\penalty0 (5):\penalty0 1203--1216,
  1998.

\bibitem[Williams and Barber(1998)]{Williams+Barber:1998}
Christopher K.~I. Williams and David Barber.
\newblock Bayesian classification with {Gaussian} processes.
\newblock \emph{IEEE Transactions on Pattern Analysis and Machine
  Intelligence}, 20\penalty0 (12):\penalty0 1342--1351, December 1998.

\bibitem[Wu(1995)]{Wu:1995}
Zongmin Wu.
\newblock Compactly supported positive definite radial functions.
\newblock \emph{Advances in Computational Mathematics}, 4\penalty0
  (1):\penalty0 283--292, 1995.

\bibitem[Zhang(2004)]{Zhang:2004}
Hao Zhang.
\newblock Inconsistent estimation and asymptotically equal interpolations in
  model-based geostatistics.
\newblock \emph{Journal of the American Statistical Association}, 99\penalty0
  (465):\penalty0 250--261, 2004.

\end{thebibliography}

 \appendix

\section{Comparison of features in GPstuff, 
GPML and FBM}\label{appendix:comparison}

Table~\ref{table:comparison} shows comparison of 
features in GPstuff, GPML and FBM.

\begin{table}
  \scriptsize
  \begin{tabular}{p{9.5cm}p{2.3cm}p{2.1cm}p{0.9cm}}
    & GPstuff & GPML & FBM \\
    \textbf{Covariance functions} &  &  & \\
    \hline
    number of elementary functions & 16 & 10 & 4 \\
    sums of elements, masking of inputs  & x & x & x \\
    delta distance & x & & x \\
    products, positive scaling of elements & x & x &  \\    
    \multicolumn{4}{l}{\textbf{Mean functions}} \\
    \hline
    number of elementary functions & 4 & 4 & 0 \\
    sums of elements, masking of inputs & x & x & \\
    products, power, scaling of elements &  & x &  \\
    marginalized parameters  & x &  &  \\
    \multicolumn{4}{l}{\textbf{Single latent likelihood/observation models}}  \\
    \hline
    Gaussian & x & x & x \\
    logistic/logit, erf/probit & x & x & MCMC\\
    Poisson & x & LA/EP/MCMC & MCMC \\
    Gaussian scale mixture & MCMC &  & MCMC\\
    Student-$t$ & x & LA/VB/MCMC & \\
    Laplacian & x & EP/VB/MCMC & \\
    mixture of likelihoods &   &  LA/EP/MCMC  & \\
    sech-squared, uniform for classification &  & x & \\
    derivative observations & for sexp covf only &  & \\
    \hangindent=0.3cm binomial, negative binomial, zero-trunc. negative binomial, 
    log-Gaussian Cox process; Weibull, log-Gaussian and log-logistic with censoring & x &  & \\
    quantile regression & MCMC/EP & & \\
    \multicolumn{4}{l}{\textbf{Multilatent likelihood/observation models}} \\
    \hline
    \hangindent=0.3cm multinomial, Cox proportional hazard model, density estimation,
    density regression, input dependent noise, input dependent overdispersion in Weibull,
    zero-inflated negative binomial  & MCMC/LA &  &  \\
    multinomial logit (softmax) & MCMC/LA & & MCMC \\
    multinomial probit & EP & & MCMC \\
    monotonicity & EP & & \\
    \multicolumn{4}{l}{\textbf{Priors for parameters ($\vartheta$)}}  \\
    \hline
    several priors, hierarchical priors & x & & x \\
    \multicolumn{4}{l}{\textbf{Sparse models}} \\
    \hline
    FITC & x & exact/EP/LA & \\
    CS, FIC, CS+FIC, PIC, VAR, DTC, SOR & x &  & \\
    PASS-GP & LA/EP & & \\
    \multicolumn{4}{l}{\textbf{Latent inference}} \\
    \hline
    exact (Gaussian only) & x & x & x  \\
    scaled Metropolis, HMC & x &  & x \\
    LA, EP, elliptical slice sampling   & x & x & \\
    variational Bayes (VB) & & x & \\
    scaled HMC (with inverse of prior cov.) &  & x & \\
    scaled HMC (whitening with approximate posterior covariance) & x  &  & \\
    parallel EP, Robust-EP  & x & & \\
    marginal corrections (cm2 and fact) & x & & \\
    state space inference (1D for some covariance functions) & x & & \\
    \multicolumn{4}{l}{\textbf{Hyperparameter inference}} \\
    \hline
    type II ML  & x & x & x \\
    type II MAP, Metropolis, HMC  & x & & x \\
    LOO-CV for Gaussian & x & x & \\
    least squares LOO-CV for non-Gaussian & & some likelihoods &\\
    LA/EP LOO-CV for non-Gaussian, k-fold CV & x & & \\
    \hangindent=0.3cm NUTS, slice sampling (SLS), surrogate SLS,
    shrinking-rank SLS, covariance-matching SLS, grid, CCD, importance sampling  & x & & \\
    \multicolumn{4}{l}{\textbf{Model assessment}} \\
    \hline
    marginal likelihood & MAP,ML & ML & \\
    LOO-CV for fixed hyperparameters & x & x & \\
    LOO-CV for integrated hyperparameters, k-fold CV, WAIC, DIC & x &  &  \\
    average predictive comparison & x &  & 
  \end{tabular}
  \normalsize
  \caption{The comparison of features in GPstuff (v4.5), GPML (v3.2) and FBM (2004-11-10)
    toolboxes. In case of model blocks the notation x means that it
    can be inferred with any inference method (EP, LA (Laplace), MCMC and
    in case of GPML also with VB). In case of sparse approximations,
    inference methods and model assessment methods x means
    that the method is available for all model blocks.\label{table:comparison}}
\end{table}

\section{Covariance functions}\label{covariance_functions}

In this section we summarize all the covariance functions in the
GPstuff package.

\subsection*{Squared exponential covariance function (\code{gpcf\_sexp})}

Probably the most widely-used covariance function is the squared
exponential (SE) 
\begin{equation}\label{sexp_ARD}
k(\x_i,\x_j)=\sigma_{\textrm{sexp}}^2\exp\left(-\frac{1}{2}\sum_{k=1}^d
  \frac{(x_{i,k}-x_{j,k})^2}{l_k^2} \right).
\end{equation}
The length scale $l_k$ governs the correlation scale in input
dimension $k$ and the magnitude $\sigma_{\mathrm{sexp}}^2$ the overall
variability of the process. A squared exponential covariance function
leads to very smooth GPs that are infinitely times mean square
differentiable.

\subsection*{Exponential covariance function (\code{gpcf\_exp})}

Exponential covariance function is defined as
\begin{equation}
k(\x_i,\x_j)=\sigma_{\textrm{exp}}^2\exp\left(-\sqrt{\sum_{k=1}^d
  \frac{(x_{i,k}-x_{j,k})^2}{l_k^2}} \right).
\end{equation}
The parameters $l_k$ and $\sigma_{\textrm{exp}}^2$ have similar
role as with the SE covariance function. The exponential covariance
function leads to very rough GPs that are not mean square
differentiable.

\subsection*{Mat\'ern class of covariance functions (\code{gpcf\_maternXX})}

The Mat\'ern class of covariance functions is given by
\begin{equation}\label{matern}
  k_{\nu}(\x_i,\x_j)=\sigma_{\textrm{m}}^2
  \frac{2^{1-\nu}}{\Gamma(\nu)}\left(\sqrt{2\nu}r\right)^{\nu}
  K_{\nu}\left(\sqrt{2\nu}r\right),
\end{equation}
where $r = \left(\sum_{k=1}^d
  \frac{(x_{i,k}-x_{j,k})^2}{l_k^2}\right)^{1/2}$. The parameter $\nu$
governs the smoothness of the process, and $K_{\nu}$ is a modified
Bessel function \citep[sec.  9.6]{Abramowitz+Stegun:1970}. 
The Mat\'ern covariance functions can be represent in a simpler form
when $\nu$ is a half integer. The Mat\'ern covariance
functions with $\nu=3/2$ (\code{gpcf\_matern32}) and $\nu=5/2$ (\code{gpcf\_matern52}) are:
\begin{align}\label{matern3/2_ARD}
k_{\nu=3/2}(\x_i,\x_j)& =\sigma_{\mathrm{m}}^2\left(1+\sqrt{3}r\right)
\exp\left(-\sqrt{3}r\right) \\
k_{\nu=5/2}(\x_i,\x_j) & =\sigma_{\textrm{m}}^2\left(1+\sqrt{5}r
  + \frac{5r^2}{3}\right)\exp\left(-\sqrt{5}r\right).\label{matern5/2_ARD}
\end{align}

\subsection*{Neural network covariance function (\code{gpcf\_neuralnetwork})}

A neural network with suitable transfer function and prior
distribution converges to a GP as the number of hidden units in the
network approaches to infinity
\citep{Neal:1996a,Williams:1996,Williams:1998,Rasmussen+Williams:2006}. A
nonstationary neural network covariance function is 

\begin{equation}\label{cf_neuralnetwork}
k(\x_i,\x_j)=\frac{2}{\pi}\sin^{-1}\left(\frac{2\mathbf{\tilde{x}}_i^{\text{T}}\Sigma \mathbf{\tilde{x}}_j}{(1+2\mathbf{\tilde{x}}_i^{\text{T}}\Sigma
      \mathbf{\tilde{x}}_i)(1+2\mathbf{\tilde{x}}_j^{\text{T}}\Sigma
      \mathbf{\tilde{x}}_j)}\right),
\end{equation}
where $\mathbf{\tilde{x}}=(1,x_1,\ldots,x_d)^{\text{T}}$ is an input
vector augmented with 1.
$\Sigma=\text{diag}(\sigma_0^2,\sigma_1^2,\ldots,\sigma_d^2)$ is a
diagonal weight prior, where $\sigma_0^2$ is a variance for bias
parameter controlling the functions offset from the origin. The
variances for weight parameters are
$\sigma_1^2,\ldots,\sigma_d^2$, and with small values for weights, the
neural network covariance function produces smooth and rigid looking
functions. The larger values for the weight variances produces more
flexible solutions.

\subsection*{Constant covariance function (\code{gpcf\_constant})}

Perhaps the simplest covariance function is the constant covariance
function
\begin{equation}\label{cf_constant}
  k(\x_i,\x_j) = \sigma^2
\end{equation}
with variance parameter $\sigma^2$. This function can be used to
implement the constant term in the dot-product covariance function
\citep{Rasmussen+Williams:2006} reviewed below.

\subsection*{Linear  covariance function  (\code{gpcf\_linear})}

The linear covariance function is
\begin{equation}\label{cf_linear}
  k(\x_i,\x_j)=\mathbf{x}_i^{\text{T}}\Sigma \mathbf{x}_j
\end{equation}
where the diagonal matrix
$\Sigma=\text{diag}(\sigma_1^2,\ldots,\sigma_D^2)$ contains the
prior variances of the linear model coefficients. Combining this
with the constant function above we can form covariance function
\citet{Rasmussen+Williams:2006}, which calls a dot-product
covariance function
\begin{equation}\label{cf_linear2}
  k(\x_i,\x_j)= \sigma^2 + \mathbf{x}_i^{\text{T}}\Sigma \mathbf{x}_j.
\end{equation}

\subsection*{Michelis-Menten covariance function  (\code{gpcf\_linearMichelismenten})}

The Michelis-Menten functional form (equivalently Type-II functional
response or Monod form, \citep{Kot:2001}) is given by
\begin{equation}
h(x) = bx / (a + x)
\end{equation}
where $b$ is asymptote of the function and $a$ the half-saturation 
point defining at which point the function is half of the asymptotic 
level. By giving a zero mean Gaussian prior for asymptote, $b \sim N(0, \sigma^2)$
the prior for $h(x)$ is $h(x) \sim N(0, H(x)H(x)^{\text{T}}\sigma^2)$ where 
$H(x) = [x_1/(a + x_1, ... , x_n/(a + x_n]^{\text{T}}$. Hence, the 
covariance function related to the Michelis-Menten mean function in case of $D$
 inputs is 
\begin{equation}\label{cf_linearMichelismenten}
k(\x_i,\x_j)= H(\x_i)\Sigma H(\x_j)^{\text{T}},
\end{equation}
where the diagonal matrix
$\Sigma=\text{diag}(\sigma_1^2,\ldots,\sigma_D^2)$ contains the
prior variances of the $b_d$ terms. 

\subsection*{Logistic mean covariance function  (\code{gpcf\_linearLogistic})}

The logistic functional form is given by
\begin{equation}
h(x) = w (\mathrm{logit}^{-1}(ax + b) - 0.5) 
\end{equation}
where $w$ is the weight of the mean function, $b$ is the intercept of the linear part 
and $a$ regression coefficient of linear part. By giving a zero mean Gaussian prior 
for weight, $w \sim N(0, \sigma^2)$
the prior for $h(x)$ is $h(x) \sim N(0, H(x)H(x)^{\text{T}}\sigma^2)$ where 
$H(x) = [h(x_1), ... , h(x_n)]^{\text{T}}$. Hence, the covariance function related 
to the logistic mean function in case of $D$ inputs is 
\begin{equation}\label{cf_linearlinearLogistic}
k(\x_i,\x_j)= H(\x_i)\Sigma H(\x_j)^{\text{T}},
\end{equation}
where the diagonal matrix
$\Sigma=\text{diag}(\sigma_1^2,\ldots,\sigma_D^2)$ contains the
prior variances of the $b_d$ terms. 

\subsection*{Piecewise polynomial functions (\code{gpcf\_ppcsX})}

The piecewise polynomial functions are the only compactly supported
covariance functions (see section \ref{cha:sparse-GP}) in
GPstuff.  There are four of them with the following forms
\begin{align}
k_{pp0}(\x_i,\x_j) =& \sigma^2(1-r)_+^{j} \\
k_{pp1}(\x_i,\x_j) =& \sigma^2(1-r)_+^{j+1} \left( (j+1)r + 1 \right) \\
k_{pp2}(\x_i,\x_j) =& \frac{\sigma^2}{3}(1-r)_+^{j+2}((j^2+4j+3)r^2+(3j+6)r+3) \\
k_{pp3}(\x_i,\x_j) =& \frac{\sigma^2}{15}(1-r)_+^{j+3} ( (j^3 + 9j^2 +23j + 15)r^3 +\nonumber\\
&(6j^2 + 36j + 45)r^2 + (15j+45)r + 15 )
\end{align}
where $j = \lfloor d/2 \rfloor + q + 1$. These functions correspond to
processes that are $2q$ times mean square differentiable at the zero
and positive definite up to the dimension $d$ \citep{Wendland:2005}.
The covariance functions are named \code{gpcf\_ppcs0},
\code{gpcf\_ppcs1}, \code{gpcf\_ppcs2}, and \code{gpcf\_ppcs3}.

\subsection*{Rational quadratic covariance function (\code{gpcf\_rq})}

The rational quadratic (RQ) covariance function \citep{Rasmussen+Williams:2006}
\begin{equation}
k_{\text{RQ}}(\x_i,\x_j) = \left(1+\frac{1}{2 \alpha} \sum_{k=1}^d
  \frac{(x_{i,k}-x_{j,k})^2}{l_k^2} \right)^{-\alpha}
%
\end{equation}
can be seen as a scale mixture of squared exponential covariance
functions with different length-scales. The smaller the parameter
$\alpha > 0$ is the more diffusive the length-scales of the mixing
components are. The parameter $l_k > 0$ characterizes the typical
length-scale of the individual components in input dimension $k$.

\subsection*{Periodic covariance function (\code{gpcf\_periodic})}

Many real world systems exhibit periodic phenomena, which can be
modelled with a periodic covariance function.  One possible
construction \citep{Rasmussen+Williams:2006} is
\begin{equation}
%
k(\x_i,\x_j) = \exp \left( - \sum_{k=1}^d \frac{2 \sin^2(\pi (x_{i,k}-x_{j,k})
    / \gamma)}{l_k^2} \right),
\end{equation}
where the parameter $\gamma$ controls the inverse length of the
periodicity and $l_k$ the smoothness of the process in dimension $k$.

\subsection*{Product covariance function (\code{gpcf\_product})}

A product of two or more covariance functions, $k_1(\x,\x') \cdot
k_2(\x,\x')...$, is a valid covariance function as well.  Combining
covariance functions in a product form can be done with
\code{gpcf\_prod}, for which the user can freely specify the
covariance functions to be multiplied with each other from the
collection of covariance functions implemented in GPstuff.

\subsection*{Categorical covariance function (\code{gpcf\_cat})}

Categorical covariance function \code{gpcf\_cat} returns
correlation 1 if input values are equal and 0 otherwise.
\begin{equation}
  k(\x_i,\x_j) =
  \begin{cases}
    1 & \text{if }\x_i-\x_j=0 \\
    0 & \text{otherwise}
  \end{cases}
\end{equation}
Categorical covariance function can be combined with other
covariance functions using \code{gpcf\_prod}, for example, to
produce hierarchical models.


\section{Observation models}

Here, we summarize all the observation models in GPstuff. Most
of them are implemented in files \code{lik\_*} which reminds that
at the inference step they are considered likelihood functions.

\subsection*{Gaussian (\code{lik\_gaussian})}

The i.i.d Gaussian noise with variance $\sigma^2$ is
\begin{equation}
\y|\f,\sigma^2 \sim N(\f,\sigma^2\mb{I}).
\end{equation}

\subsection*{Student-$t$ (\code{lik\_t}, \code{lik\_gaussiansmt})}

The Student-$t$ observation model (implemented in \code{lik\_t}) is
\begin{equation}
 \y | \f, \nu, \sigma_t  \sim \prod_{i=1}^n \frac{\Gamma((\nu+1)/2)}{\Gamma(\nu/2)\sqrt{\nu\pi}\sigma_t}\left(1 +
  \frac{(y_i-f_i)^2}{\nu\sigma_t^2} \right)^{-(\nu+1)/2},
\end{equation}
where $\nu$ is the degrees of freedom and $\sigma$ the scale
parameter. The scale
mixture version of the Student-$t$ distribution is implemented in
\code{lik\_gaussiansmt} and it is parametrized as
\begin{align}
y_i | f_i,\alpha, U_i & \sim N(f_i, \alpha U_i)\\
U_i & \sim \text{Inv-}\chi^2(\nu, \tau^2), 
\end{align}
where each observation has its own noise variance $\alpha U_i$
\citep{Neal:1997,Gelman+etal+BDA3:2013}. 
\subsection*{Logit (\code{lik\_logit})}

The logit transformation gives the probability for $y_i$ of being
$1$ or $-1$ as
\begin{equation}
p_{\text{logit}}(y_i|f_i) = \frac{1}{1 + \exp(-y_i f_i)}.
\end{equation}

\subsection*{Probit (\code{lik\_probit})}

The probit transformation gives the probability for $y_i$ of being $1$
or $-1$ as
\begin{equation}
p_{\text{probit}}(y_i|f_i) = \Phi(y_if(\mb{x}_i)) =
\int_{-\infty}^{y_if(\mb{x}_i)} N(z|0,1) d z.
\end{equation}

\subsection*{Poisson (\code{lik\_poisson})}

The Poisson observation model with expected number of cases $\mb{e}$ is
\begin{equation}
 \y|\f,\mb{e}  \sim \prod_{i=1}^{n} \Poisson(y_i|\exp(f_i)e_i).
\end{equation}

\subsection*{Negative-Binomial (\code{lik\_negbin})}

The negative-binomial is a robustified version of the Poisson
distribution. It is parametrized
\begin{equation}
 \y |\f,\mb{e}, r  \sim \prod_{i=1}^n
 \frac{\Gamma(r+y_i)}{y_i!\Gamma(r)}
\left(\frac{r}{r+\mu_i}\right)^r \left(\frac{\mu_i}{r+\mu_i}\right)^{y_i}
\end{equation}
where $\mu_i = \mb{e}\exp(f_i)$, $r$ is the dispersion parameter
governing the variance, $e_i$ is the expected number of cases and
$y_i$ is positive integer telling the observed count.

\subsection*{Binomial (\code{lik\_binomial})}

The binomial observation model with the probability of success $p_i =
\exp(f_i)/ (1+\exp(f_i))$ is
\begin{equation}
\y|\f,\mathbf{z} \sim \prod_{i=1}^N \frac{z_i!}{y_i!(z_i-y_i)!} p_i^{y_i}(1-p_i)^{(z_i-y_i)}.
\end{equation}
Here, $z_i$ denotes the number of trials and $y_i$ is the number of
successes.

\subsection*{Weibull (\code{lik\_weibull})}

The Weibull likelihood is defined as
\begin{equation}
  L = \prod_{i=1}^n r^{1-z_i} \exp \left( (1-z_i)f(\x_i)+(1-z_i)(r-1)\log(y_i)-\exp(f(\x_i))y_i^r \right),
\end{equation}
where $\mathbf{z}$ is a vector of censoring indicators with $z_i = 0$ for
uncensored event and $z_i = 1$ for right censored event for
observation $i$ and $r$ is the shape parameter. Here we present only the likelihood function because we don't have observation model for the censoring.

\subsection*{Log-Gaussian (\code{lik\_loggaussian})}

The Log-Gaussian likelihood is defined as
\begin{eqnarray}
  L =  && \prod_{i=1}^n (2\pi \sigma^2)^{-(1-z_i)/2}y_i^{1-z_i} \exp \left(-\frac{1}{2\sigma^2}(1-z_i)(\log (y_i) - f(\x_i))^2\right) \\ \nonumber
  &\times & \left(1 - \Phi \left(\frac{\log(y_i) - f(\x_i)}{\sigma}\right)\right)^{z_i} 
\end{eqnarray}
where $\sigma$ is the scale parameter.

\subsection*{Log-logistic (\code{lik\_loglogistic})}
The log-logistic likelihood is defined as
\begin{equation}
 L = \prod_{i=1}^n \left( \frac{ry^{r-1}}{\exp(f(\x_i))} \right)^{1-z_i} \left( 1 + \left(\frac{y}{\exp(f(\x_i))}\right)^r \right)^{z_i-2}
\end{equation}

\subsection*{Cox proportional hazard model (\code{lik\_coxph})}

The likelihood contribution for the possibly right censored $i$th
observation $(y_i,\delta_i)$ is assumed to be
\begin{equation}
l_i=h_i(y_i)^{(1-\delta_i)} \exp \left(
  -\int_0^{y_i}h_i(t)dt \right).
\end{equation}
Using the piecewise log-constant assumption for the hazard rate
function, the contribution of the observation $i$ for the likelihood
results in
\begin{equation}
l_i=[\lambda_k \exp(\eta_i)]^{(1-\delta_i)}\exp \left( -[(y_i-s_{k-1})\lambda_k
  + \sum_{g=1}^{k-1}(s_g-s_{g-1})\lambda_g ]\exp(\eta_i) \right),
\end{equation}
where $y_i\in(s_{k-1},s_k]$

\subsection*{Input-dependent noise (\code{lik\_inputdependentnoise})}

The input-dependent noise observation model is defined as
\begin{align}
  \y|\f^{(1)}, \f^{(2)}, \sigma^2 \sim \prod_{i=1}^n N(y_i | f_i^{(1)}, \sigma^2 \exp(f_i^{(2)})),
\end{align}
with latent function $f^{(1)}$ defining the mean and $f^{(2)}$
defining the variance.

\subsection*{Input-dependent Weibull (\code{lik\_inputdependentweibull})}

The input-dependent Weibull observation model is defined as
\begin{eqnarray}
\y |\f^{(1)}, \f^{(2)},\mathbf{z}  \sim && \prod_{i=1}^n \exp(f_i^{(2)})^{1-z_i} \exp \Big{(} (1-z_i)f^{(1)}_i \\ \nonumber
 &+&(1-z_i)(\exp(f_i^{(2)})-1)\log(y_i)\exp(f^{(1)}_i)y_i^{\exp(f_i^{(2)})} \Big{)}, \nonumber
\end{eqnarray}
where $\mathbf{z}$ is a vector of censoring indicators with $z_i = 0$ for
uncensored event and $z_i = 1$ for right censored event for
observation $i$. 

\subsection*{Zero-inflated Negative-Binomial (\code{lik\_zinegbin})}
The Zero-Inflated Negative-Binomial observation model is defined as 
\begin{equation}
 \y |\f,\mb{e}, r  \sim \prod_{i=1}^n
 \frac{\Gamma(r+y_i)}{y_i!\Gamma(r)}
\left(\frac{r}{r+\mu_i}\right)^r \left(\frac{\mu_i}{r+\mu_i}\right)^{y_i},
\end{equation}
where $\mu_i = e_i\exp(f(\x_i))$ and $r$ is the dispersion parameter
governing the variance.


\section{Priors}

This appendix lists all the priors implemented in the
GPstuff package.

\subsection*{Gaussian prior (\code{prior\_gaussian})}

The Gaussian distribution is parametrized as
\begin{equation}
p(\theta)=\frac{1}{\sqrt{2\pi\sigma^2}}\exp\left(-\frac{1}{2\sigma^2}(\theta-\mu)^2\right)
\end{equation}
where $\mu$ is a location parameter and $\sigma^2$ is a scale
parameter.

\subsection*{Log-Gaussian prior (\code{prior\_loggaussian})}

The log-Gaussian distribution is parametrized as
\begin{equation}
p(\theta)=\frac{1}{\theta\sqrt{2\pi\sigma^2}}\exp\left(-\frac{1}{2\sigma^2}(\log(\theta)-\mu)^2\right)
\end{equation}
where $\mu$ is a location parameter and $\sigma^2$ is a scale
parameter.

\subsection*{Laplace prior (\code{prior\_laplace})}

The Laplace distribution is parametrized as
\begin{equation}
p(\theta)=\frac{1}{2\sigma}\exp\left(-\frac{|\theta-\mu|}{\sigma}\right)
\end{equation}
where $\mu$ is a location parameter and $\sigma>0$ is a scale
parameter.

\subsection*{Student-$t$ prior (\code{prior\_t})}

The Student-$t$ distribution is parametrized as
\begin{equation}
p(\theta)=\frac{\Gamma((\nu+1)/2)}{\Gamma(\nu/2)\sqrt{\nu\pi\sigma^2}}\left(1+\frac{(\theta-\mu)^2}{\nu\sigma^2}\right)^{-(\nu+1)/2}
\end{equation}
where $\mu$ is a location parameter, $\sigma^2$ is a scale
parameter and $\nu>0$ is the degrees of freedom.

\subsection*{Square root Student-$t$ prior (\code{prior\_sqrtt})}

The square root Student-$t$ distribution is parametrized as
\begin{equation}
p(\theta^{1/2})=\frac{\Gamma((\nu+1)/2)}{\Gamma(\nu/2)\sqrt{\nu\pi\sigma^2}}\left(1+\frac{(\theta-\mu)^2}{\nu\sigma^2}\right)^{-(\nu+1)/2}
\end{equation}
where $\mu$ is a location parameter, $\sigma^2$ is a scale
parameter and $\nu>0$ is the degrees of freedom.

\subsection*{Scaled inverse-$\chi^2$ prior (\code{prior\_sinvchi2})}

The scaled inverse-$\chi^2$ distribution is parametrized as
\begin{equation}
p(\theta)=\frac{(\nu/2)^{\nu/2}}{\Gamma(\nu/2)}(s^2)^{\nu/2}\theta^{-(\nu/2+1)}e^{-\nu s^2/(2\theta)}
\end{equation}
where $s^2$ is a scale parameter and $\nu>0$ is the degrees of freedom
parameter.

\subsection*{Gamma prior (\code{prior\_gamma})}

The gamma distribution is parametrized as
\begin{equation}
p(\theta)=\frac{\beta^{\alpha}}{\Gamma(\alpha)}\theta^{\alpha-1}e^{-\beta\theta}
\end{equation}
where $\alpha>0$ is a shape parameter and $\beta>0$ is an inverse
scale parameter.

\subsection*{Inverse-gamma prior (\code{prior\_invgamma})}

The inverse-gamma distribution is parametrized as
\begin{equation}
p(\theta)=\frac{\beta^{\alpha}}{\Gamma(\alpha)}\theta^{-(\alpha+1)}e^{-\beta/\theta}
\end{equation}
where $\alpha>0$ is a shape parameter and $\beta>0$ is a scale
parameter.

\subsection*{Uniform prior (\code{prior\_unif})}

The uniform prior is parametrized as
\begin{equation}
p(\theta)\propto 1.
\end{equation}

\subsection*{Square root uniform prior (\code{prior\_sqrtunif})}

The square root uniform prior is parametrized as
\begin{equation}
p(\theta^{1/2})\propto 1.
\end{equation}

\subsection*{Log-uniform prior (\code{prior\_logunif})}

The log-uniform prior is parametrized as
\begin{equation}
p(\log(\theta))\propto 1.
\end{equation}

\subsection*{Log-log-uniform prior (\code{prior\_loglogunif})}

The log-log-uniform prior is parametrized as
\begin{equation}
p(\log(\log(\theta)))\propto 1.
\end{equation} 

\section{Transformation of hyperparameters}
\label{sec_log_transformation}

The inference on the parameters of covariance functions is
conducted mainly transformed space. Most of ten used transformation
is log-transformation, which has the advantage that the parameter
space $(0,\infty)$ is transformed into $(-\infty,\infty)$. The
change of parametrization has to be taken into account in the
evaluation of the probability densities of the model. If parameter
$\theta$ with probability density $p_{\theta}(\theta)$ is
transformed into the parameter $w = f(\theta)$ with equal number of
components, the probability density of $w$ is given by

\begin{equation}\label{Transform.of.variables}
p_w(w) = |J|p_{\theta}(f^{-1}(w)),
\end{equation}
where $J$ is the Jacobian of the transformation $\theta =
f^{-1}(w)$. The parameter transformations are discussed shortly,
for example, in \citet{Gelman+etal+BDA3:2013}[p. 21].

Due to the log transformation $w=\log(\theta)$ transformation the
probability densities $p_{\theta}(\theta)$ are changed to the
densities
\begin{equation}\label{Transform.from.th.to.log}
p_w(w) = |J|p_{\theta}(\exp(w)) = |J|p_{\theta}(\theta),
\end{equation}
where the Jacobian is $J = \frac{\partial \exp(w)}{\partial w} =
\exp(w) = \theta$. Now, given Gaussian observation model (see
Section \ref{sec:gauss-observ-model}) the posterior of $w$ can be
written as
\begin{equation}
p_w(w|\mathcal{D}) \propto
p(\mb{y}|\mb{X},\mathbf{\theta})p(\mathbf{\theta}|\mathbf{\gamma})\theta,
\end{equation}
which leads to energy function
\begin{align}
E(w) &= -\log p(\mb{y}|\mb{X},\mathbf{\theta}) - \log
p(\mathbf{\theta}|\mathbf{\gamma}) - \log(|\theta|). \nonumber \\ 
 & = E(\theta) - \log(\theta)\nonumber,
\end{align}
where the absolute value signs are not shown explicitly around $\theta$
because it is strictly positive. Thus, the log transformation just adds
term $-\log \theta$ in the energy function.

The inference on  $w$ requires also the gradients of an energy
function $E(w)$. These can be obtained easily with the chain rule
\begin{align}
\frac{\partial E(w)}{\partial w} &= \frac{\partial}{\partial
  \theta}\left[E(\theta)-\log(|J|)\right]\frac{\partial \theta}{\partial w}
\nonumber \\
 & = \left[\frac{\partial E(\theta)}{\partial \theta}-\frac{\partial
   \log(|J|)}{\partial \theta}\right]\frac{\partial \theta}{\partial
 w} \nonumber \\
 & = \left[\frac{\partial E(\theta)}{\partial
     \theta}-\frac{1}{|J|}\frac{\partial |J|}{\partial \theta}\right]J.
\end{align}
Here we have used the fact that the last term, derivative of $\theta$
with respect to $w$, is the same as the Jacobian $J = \frac{\partial
  \theta}{\partial w} = \frac{\partial f^{-1}}{\partial w}$.  Now in
the case of log transformation the Jacobian can be replaced by
$\theta$ and the gradient is gotten an easy expression

\begin{align}
\frac{\partial E(w)}{\partial w} 
 & = \frac{\partial E(\theta)}{\partial \theta}\theta-1.
\end{align}

\section{Developer appendix}\label{developer_appendix}

This section provides additional description of \code{lik},
\code{gpcf} and \code{prior} functions, and argument interface for
optimisation and sampling functions.

\subsection{\code{lik} functions}

New likelihoods can be added by copying and modifying one of the
existing \code{lik} functions. We use \code{lik\_negbin} as an
example of log-concave likelihood and \code{lik\_t} as an example
of non-log-concave likelihood. Note that adding a new
non-log-concave likelihood is more challenging as the posterior of
the latent values may be multimodal, which makes it more difficult
to implement stable Laplace and EP methods
\citep[see, e.g.,][]{Vanhatalo+Jylanki+Vehtari:2009,Jylanki+Vanhatalo+Vehtari:2011}.

\subsubsection{\code{lik\_negbin}}

inputParser is used to make argument checks and set some default
values. If the new likelihood does not have parameters, see for example,
\code{lik\_poisson}.
\begin{verbatim}
  ip=inputParser;
  ip.FunctionName = 'LIK_NEGBIN';
  ip.addOptional('lik', [], @isstruct);
  ip.addParamValue('disper',10, @(x) isscalar(x) && x>0);
  ip.addParamValue('disper_prior',prior_logunif(), @(x) isstruct(x) || isempty(x));
  ip.parse(varargin{:});
  lik=ip.Results.lik;
...
\end{verbatim}

Function handles to the subfunctions provide object-oriented
behaviour (at time when GPstuff project was started, real classes
in Matlab were too inefficient and Octave still lacks proper support).
If some of the subfunctions have not been implemented,
corresponding function handles should not be defined.
\begin{verbatim}
  if init
    % Set the function handles to the subfunctions
    lik.fh.pak = @lik_negbin_pak;
    lik.fh.unpak = @lik_negbin_unpak;
    lik.fh.lp = @lik_negbin_lp;
    lik.fh.lpg = @lik_negbin_lpg;
    lik.fh.ll = @lik_negbin_ll;
    lik.fh.llg = @lik_negbin_llg;    
    lik.fh.llg2 = @lik_negbin_llg2;
    lik.fh.llg3 = @lik_negbin_llg3;
    lik.fh.tiltedMoments = @lik_negbin_tiltedMoments;
    lik.fh.siteDeriv = @lik_negbin_siteDeriv;
    lik.fh.predy = @lik_negbin_predy;
    lik.fh.predprcty = @lik_negbin_predprcty;
    lik.fh.invlink = @lik_negbin_invlink;
    lik.fh.recappend = @lik_negbin_recappend;
  end

end
\end{verbatim}

\code{lik\_negbin\_pak} and \code{lik\_negbin\_unpak} functions are
used to support generic optimization and sampling functions, which
assume that the parameters are presented as a vector. Parameters which
have empty prior ([]) are ignored. These subfunctions are
mandatory (even if there are no parameters).

\code{lik\_negbin\_lp} and \code{lik\_negbin\_lpg} compute the log
prior density of the parameters and its gradient with respect to
parameters. Parameters which have empty prior ([]) are ignored.
These subfunctions are needed if there are likelihood parameters.

\code{lik\_negbin\_ll} and \code{lik\_negbin\_llg} compute the log
likelihood and its gradient with respect to parameters and latents.
Parameters which have empty prior ([]) are ignored. These
subfunctions are mandatory.

\code{lik\_negbin\_llg2} computes second gradients of the log
likelihood. This subfunction is needed when using Laplace
approximation or EP for inference with non-Gaussian likelihoods.

\code{lik\_negbin\_llg3} computes third gradients of the log
likelihood. This subfunction is needed when using Laplace
approximation for inference with non-Gaussian likelihoods.

\code{lik\_negbin\_tiltedMoments} returns the marginal moments for
the EP algorithm. This subfunction is needed when using EP for
inference with non-Gaussian likelihoods.

\code{lik\_negbin\_siteDeriv} evaluates the expectation of the
gradient of the log likelihood term with respect to the likelihood
parameters for EP. This subfunction is needed when using EP for
inference with non-Gaussian likelihoods and there are likelihood
parameters.

\code{lik\_negbin\_predy} returns the predictive mean, variance and
density of $y$. This subfunction is needed when computing posterior
predictive distributions for future observations.

\code{lik\_negbin\_predprcty} returns the percentiles of predictive
density of $y$. This subfunction is needed when using function
\code{gp\_predprcty}.

\code{lik\_negbin\_init\_negbin\_norm} is a private function for
\code{lik\_negbin}. It returns function handle to a function
evaluating Negative-Binomial * Gaussian which is used for
evaluating (likelihood * cavity) or (likelihood * posterior). This
subfunction is needed by subfunctions
\code{lik\_negbin\_tiltedMoments}, \code{lik\_negbin\_siteDeriv}
and \code{lik\_negbin\_predy}. Note that this is not needed for
those likelihoods for which integral over the likelihood times
Gaussian is computed using other than quadrature integration.

\code{lik\_negbin\_invlink} returns values of inverse link
function. This subfunction is needed when using function
\code{gp\_predprctmu}.

\code{lik\_negbin\_recappend} This subfunction is needed when using
MCMC sampling (\code{gp\_mc}).

\subsubsection{\code{lik\_t}}

\code{lik\_t} includes some additional subfunctions useful for
non-log-concave likelihoods.

\code{lik\_t\_tiltedMoments2} returns the marginal moments for the
EP algorithm. This subfunction is needed when using robust-EP for
inference with non-Gaussian likelihoods.

\code{lik\_t\_siteDeriv2} evaluates the expectation of the gradient
of the log likelihood term with respect to the likelihood
parameters for EP. This subfunction is needed when using robust-EP
for inference with non-Gaussian likelihoods and there are
likelihood parameters.

\code{lik\_t\_optimizef} function to optimize the latent variables
with EM algorithm. This subfunction is needed when using likelihood
specific optimization method for mode finding in the Laplace
algorithm.

\subsection{\code{gpcf} functions}

New covariance functions can be added by copying and modifying one of the
existing \code{gpcf} functions. We use \code{gpcf\_sexp} as an
example.

\subsubsection{\code{gpcf\_sexp}}

inputParser is used to make argument checks and set some default
values. If the new covariance function does not have parameters,
corresponding lines can be removed (see, e.g.,
\code{lik\_cat}).
\begin{verbatim}
  ip=inputParser;
  ip.FunctionName = 'GPCF_SEXP';
  ip.addOptional('gpcf', [], @isstruct);
  ip.addParamValue('magnSigma2',0.1, @(x) isscalar(x) && x>0);
  ip.addParamValue('lengthScale',1, @(x) isvector(x) && all(x>0));
  ip.addParamValue('metric',[], @isstruct);
  ip.addParamValue('magnSigma2_prior', prior_logunif(), ...
                   @(x) isstruct(x) || isempty(x));
  ip.addParamValue('lengthScale_prior',prior_t(), ...
                   @(x) isstruct(x) || isempty(x));
  ip.addParamValue('selectedVariables',[], @(x) isempty(x) || ...
                   (isvector(x) && all(x>0)));
  ip.parse(varargin{:});
  gpcf=ip.Results.gpcf;
...
\end{verbatim}

Optional 'metric' can be used to replace default simple euclidean
metric. Currently it is possible to use delta distance or group
covariates to have common lengthScale parameters. Other potential
metrics could be, for example, Manhattan or distance matrix based
metrics. The metric function is called only if it has been
explicitly set. This adds some additional code branches to the
code, but avoids extra overhead in computation when using simple
euclidean metric case.

Function handles to the subfunctions provide object-oriented
behaviour (at time when GPstuff project was started, real classes
in Matlab were too inefficient and Octave still lacks proper support).
If some of the subfunctions have not been implemented,
corresponding function handles should not be defined.
\begin{verbatim}
  if init
    % Set the function handles to the subfunctions
    gpcf.fh.pak = @gpcf_sexp_pak;
    gpcf.fh.unpak = @gpcf_sexp_unpak;
    gpcf.fh.lp = @gpcf_sexp_lp;
    gpcf.fh.lpg= @gpcf_sexp_lpg;
    gpcf.fh.cfg = @gpcf_sexp_cfg;
    gpcf.fh.cfdg = @gpcf_sexp_cfdg;
    gpcf.fh.cfdg2 = @gpcf_sexp_cfdg2;
    gpcf.fh.ginput = @gpcf_sexp_ginput;
    gpcf.fh.ginput2 = @gpcf_sexp_ginput2;
    gpcf.fh.ginput3 = @gpcf_sexp_ginput3;
    gpcf.fh.ginput4 = @gpcf_sexp_ginput4;
    gpcf.fh.cov = @gpcf_sexp_cov;
    gpcf.fh.trcov  = @gpcf_sexp_trcov;
    gpcf.fh.trvar  = @gpcf_sexp_trvar;
    gpcf.fh.recappend = @gpcf_sexp_recappend;
  end
\end{verbatim}

\code{gpcf\_sexp\_pak} and \code{gpcf\_sexp\_unpak} functions are
used to support generic optimization and sampling functions, which
assume that parameters are presented as a vector. Parameters which
have empty prior ([]) are ignored. These subfunctions are
mandatory (even if there are no parameters).

\code{gpcf\_sexp\_lp} and \code{gpcf\_sexp\_lpg} compute the log
prior density of the parameters and its gradient with respect to
parameters. Parameters which have empty prior ([]) are ignored.
These subfunctions are mandatory.

\code{gpcf\_sexp\_cov} evaluates covariance matrix between two
input vectors. This is a mandatory subfunction used for example in
prediction and energy computations.

\code{gpcf\_sexp\_trcov} evaluates training covariance matrix of
inputs. This is a mandatory subfunction used for example in
prediction and energy computations. If available,
\code{gpcf\_sexp\_trcov} uses faster C-implementation \code{trcov}
for covariance computation (see linuxCsource/trcov.c). \code{trcov}
includes C-code implementation for all currently available
covariance functions. If \code{trcov} is not available,
\code{gpcf\_sexp\_trcov} uses M-code computation.

\code{gpcf\_sexp\_trvar} evaluates training variance vector. This
is a mandatory subfunction used for example in prediction and
energy computations.

\code{gpcf\_sexp\_cfg} evaluates gradient of covariance function
with respect to the parameters. \code{gpcf\_sexp\_cfg} has four
different calling syntaxes. First one is a mandatory used in
gradient computations. Second and third are needed when using
sparse approximations (e.g. FIC). Fourth one is needed when using
memory save option in \code{gp\_set} (without memory save option, all
matrix derivatives with respect to all covariance parameters are
computed at once, which may take lot of memory if $n$ is large and
there are many parameters).

\code{gpcf\_sexp\_cfdg} evaluates gradient of covariance function,
of which has been taken partial derivative with respect to x, with
respect to parameters. This subfunction is needed when using
derivative observations.

\code{gpcf\_sexp\_cfdg2} evaluates gradient of covariance function,
of which has been taken partial derivative with respect to both
input variables x, with respect to parameters. This subfunction is
needed when using derivative observations.

\code{gpcf\_sexp\_ginput} evaluates gradient of covariance function with
respect to x. This subfunction is needed when computing gradients
with respect to inducing inputs in sparse approximations.

\code{gpcf\_sexp\_ginput2} evaluates gradient of covariance
function with respect to both input variables x and x2. This
subfunction is needed when computing gradients with respect to both
input variables x and x2 (in same dimension). This subfunction is
needed when using derivative observations.

\code{gpcf\_sexp\_ginput3} evaluates gradient of covariance
function with respect to both input variables x and x2. This
subfunction is needed when computing gradients with respect to both
input variables x and x2 (in different dimensions). This
subfunction is needed when using derivative observations.

\code{gpcf\_sexp\_ginput4} evaluates gradient of covariance
function with respect to x. Simplified and faster version of
\code{sexp\_ginput}, returns full matrices. This subfunction is needed when
using derivative observations.

\code{gpcf\_sexp\_recappend} is needed when using MCMC sampling
(\code{gp\_mc}).

\subsection{\code{prior} functions}

New priors can be added by copying and modifying one of the
existing \code{prior} functions. We use \code{prior\_t} as an
example.

inputParser is used to make argument checks and set some default
values. If the new prior does not have parameters,
corresponding lines can be removed (see, e.g., \code{prior\_unif}).

\begin{verbatim}
  ip=inputParser;
  ip.FunctionName = 'PRIOR_T';
  ip.addOptional('p', [], @isstruct);
  ip.addParamValue('mu',0, @(x) isscalar(x));
  ip.addParamValue('mu_prior',[], @(x) isstruct(x) || isempty(x));
  ip.addParamValue('s2',1, @(x) isscalar(x) && x>0);
  ip.addParamValue('s2_prior',[], @(x) isstruct(x) || isempty(x));
  ip.addParamValue('nu',4, @(x) isscalar(x) && x>0);
  ip.addParamValue('nu_prior',[], @(x) isstruct(x) || isempty(x));
  ip.parse(varargin{:});
  p=ip.Results.p;
...
\end{verbatim}

Function handles to the subfunctions provide object-oriented
behaviour (at time when GPstuff project was started, real classes
in Matlab were too inefficient and Octave still lacks proper support).
If some of the subfunctions have not been implemented,
corresponding function handles should not be defined.
\begin{verbatim}
  if init
    % set functions
    p.fh.pak = @prior_t_pak;
    p.fh.unpak = @prior_t_unpak;
    p.fh.lp = @prior_t_lp;
    p.fh.lpg = @prior_t_lpg;
    p.fh.recappend = @prior_t_recappend;
  end
\end{verbatim}

\code{prior\_t\_pak} and \code{prior\_t\_unpak} functions are
used to support generic optimization and sampling functions, which
assume that parameters are presented as a vector. Parameters which
have empty prior ([]) are ignored. These subfunctions are
mandatory (even if there are no parameters).

\code{prior\_t\_lp} and \code{prior\_t\_lpg} compute the log
prior density of the parameters and its gradient with respect to
parameters. Parameters which have empty prior ([]) are ignored.
These subfunctions are mandatory.

\code{prior\_t\_recappend} This subfunction is needed when using
MCMC sampling (\code{gp\_mc}).

\subsection{Optimisation functions}

Adding new optimisation functions is quite easy as covariance
function and likelihood parameters and inducing inputs can be
optimised using any optimisation function using same syntax as, for
example, \code{fminunc} by Mathworks or \code{fminscg} and
\code{fminlbfgs} included in GPstuff.

Syntax is 
\begin{verbatim}
  X = FMINX(FUN, X0, OPTIONS)
\end{verbatim}
where FUN accepts input X and returns a scalar function value F and
its scalar or vector gradient G evaluated at X, X0 is initial value
and OPTIONS is option structure.

Then new optimisation algorithm can be used with code
\begin{verbatim}
gp=gp_optim(gp,x,y,'opt',opt,'optimf',@fminx);
\end{verbatim}

\subsection{Other inference methods}

Adding new Monte Carlo sampling functions is more complicated.
\code{gp\_mc} function is used to allow the use of different
sampling methods for covariance function parameters, likelihood
parameters and latent values with one function call. Thus adding
new sampling method requires modification of \code{gp\_mc}. Note
also the subfunctions \code{gpmc\_e} and \code{gpmc\_g}, which
return energy (negative log posterior density) and its gradient
while handling \code{infer\_params} option properly. See \code{gp\_mc}
for further information.

Adding new analytic approximation, for example, variational Bayes
approach, would require major effort implementing many inference
related functions. To start see functions with name starting
\code{gpla\_} or \code{gpep\_} and \code{gp\_set}.

\end{document}